\ifdefined\pdfoutput\pdfoutput=1\fi
\documentclass{article}

\PassOptionsToPackage{numbers,sort&compress}{natbib}
\usepackage[eandd, final]{neurips_2026}

\usepackage[utf8]{inputenc}
\usepackage[T1]{fontenc}
\usepackage[hidelinks]{hyperref}
\usepackage{url}
\usepackage{booktabs}
\usepackage{amsfonts}
\usepackage{amsmath}
\usepackage{amssymb}
\usepackage{nicefrac}
\usepackage{microtype}
\usepackage[table]{xcolor}
\usepackage{graphicx}
\usepackage{rotating}
\usepackage{placeins}
\usepackage{wrapfig}
\usepackage{tabularx}
\usepackage{threeparttable}
\newcolumntype{Y}{>{\raggedright\arraybackslash}X}

\definecolor{rankA}{HTML}{B36C81}
\definecolor{rankB}{HTML}{CD96A4}
\definecolor{rankC}{HTML}{E0B9C1}
\definecolor{rankD}{HTML}{F1DDE2}

\makeatletter
\def\stripzero#1{\@stripzero@aux#1\@nil}
\def\@stripzero@aux0.#1\@nil{.#1}
\makeatother
\newcommand{\ci}[3]{$#1^{+\stripzero{#2}}_{-\stripzero{#3}}$}
\newcommand{\ciB}[3]{$\mathbf{#1}^{+\stripzero{#2}}_{-\stripzero{#3}}$}

\title{Data-Scarce, Multi-Granular, Multimodal: Benchmark Design for Biosynthetic Gene Clusters}

\author{%
  Hanlin Xiao$^{1,2}$ \quad Eriko Takano$^{1,3,4}$ \quad Mauricio A.~\'Alvarez$^{2}$ \quad Rainer Breitling$^{5,*}$ \\[3pt]
  $^{1}$Manchester Institute of Biotechnology; $^{2}$Department of Computer Science \\
  $^{3}$Department of Chemistry; $^{4}$Division of Evolution, Infection and Genomics \\
  $^{5}$Division of Molecular and Cellular Function \\
  The University of Manchester, Manchester, United Kingdom \\
  $^{*}$Corresponding author: \texttt{rainer.breitling@manchester.ac.uk}
}

\begin{document}

\maketitle

\begin{abstract}
\looseness=-1
Biosynthetic gene clusters (BGCs) are co-located genes that encode the biosynthesis of specialized metabolites, an important source of antibiotics, anticancer agents, and agricultural bioactives. Recent advances in deep learning, particularly self-supervised foundation models, have spurred growing interest in BGC sequence modeling, but evaluation has lagged behind: existing benchmarks often rely on coarse classification over small curated datasets, providing limited discrimination among models and little insight into downstream utility. We propose a benchmark-design framework in which an informative benchmark should be \emph{discriminative}, separating models with statistical confidence, and \emph{applicable}, spanning scientifically meaningful tasks under different levels of generalization. We instantiate this framework as \textbf{BGC-Bench}\footnote{Code and benchmark data are available at \url{https://github.com/Nugkta/BGC-Bench} and \url{https://huggingface.co/datasets/Nugkta/bgc-bench}.}, which derives five tasks from MIBiG, ranging from local A-domain substrate prediction to broad BGC--molecule retrieval, evaluates them under genotype- and phenotype-shift splits, and includes an active-learning setting for label-scarce discovery. BGCs further provide a shared testbed across granularities and modalities: BGC-Bench compares twelve encoders on nucleotide, amino-acid, or Pfam-domain sequences, including frequency-based baselines for each input type, and combines BGC and chemical encoders for cross-modal retrieval. No single representation dominates: the best granularity depends on the biological scale of the task, and pretrained encoders do not consistently outperform other representations across tasks and splits. Active-learning experiments show that most tasks remain data-limited, encoder rankings change with labeling budget, and acquisition strategies yield modest gains. These results show that BGC model evaluation requires task-diverse, shift-aware benchmarks, and the framework can be generalized to benchmark design beyond BGCs.
\end{abstract}

\section{Introduction}
\label{sec:intro}
Biosynthetic gene clusters (BGCs) are physically co-localized, often co-regulated groups of genes that encode the biosynthesis of specialized metabolites~\citep{Medema2015MIBiG}. They occur most commonly in microbial genomes, and also in fungi and plants~\citep{Polturak2021PlantBGCs}. The metabolites they produce constitute one of the richest sources of bioactive compounds, ranging from antibiotics and antifungals to anticancer agents and crop-protection compounds~\citep{Pham2019MicrobialNaturalProducts, Zhang2023Agrochemicals}. Natural products and their derivatives accounted for 32\% of small-molecule drugs first approved worldwide from 1981 to 2019~\citep{Newman2020NaturalProducts}, making BGCs a high-value target for both biological and computational study.

The ordered structure of BGCs makes them natural targets for sequence modeling, a field that has advanced rapidly with foundation models for natural language~\citep{Min2023NLPPretrainedLMs}, protein~\citep{Hayes2025ESM3}, and genomic sequences~\citep{Nguyen2024Evo, DallaTorre2025NucleotideTransformer, brixi2026evo2}. BGCs differ from these settings in several ways. A cluster can be represented at multiple biological scales: as a sequence of nucleotides, amino acids, genes, or Pfam domains~\citep{mistry2021pfam} (where each domain represents a functional subunit of a protein), with mappings between scales defined by biological processes and computational annotation. They are also intrinsically cross-modal, since each sequence encodes the biosynthetic pathway to a chemical product, and further corresponds to the mass-spectrometric measurement pattern of that product~\citep{Schorn2021PairedOmics, VanDerHooft2020LinkingOmics,Leao2022NPOmix}. Furthermore, they are data-scarce: experimental characterization of BGC products is complicated and time-consuming; an estimated 80--90\% of BGCs in prolific bacterial secondary-metabolite producers are silent or only weakly expressed under standard laboratory conditions~\citep{Mao2018SilentBGCs,Liu2021StreptomycesSilentBGCs}, while the latest curated dataset contains only about three thousand experimentally characterized BGCs~\citep{mibig}.

\looseness=-1
Recent advances in BGC sequence modeling, ranging from adaptations of general protein-sequence foundation models~\citep{bgcprophet,Malygina2024NaturalPPLuM} to dedicated encoders pretrained on BGC corpora~\citep{bigcarp,kawano2026roberta}, have not been matched by comparable progress in model evaluation. Common practice reports coarse class prediction on a curated dataset of experimentally validated BGCs~\citep{deepbgc,bigcarp,mibig}, or finer classification schemes derived from structure-based natural-product classifiers~\citep{npclassifier,porokhin2025bgcat}. Coarse class prediction provides an established reference, but its taxonomy mixes biosynthetic logic and product chemistry, and domain composition already predicts many classes well. BGCat adds temporal and gene cluster family (GCF) holdouts; broader comparisons across genetic novelty, chemical novelty, and downstream tasks remain fragmented. A shared evaluation suite is needed to distinguish representations beyond coarse classification and assess their utility for natural-product discovery and experimental design.

We introduce \textbf{BGC-Bench}, a benchmark that builds on MIBiG~\citep{mibig}, the community-curated reference database of experimentally characterized BGCs and their associated natural products, by adding biologically and chemically grounded evaluation splits and derived annotations. We first formalize a domain-agnostic benchmark-design framework (Figure~\ref{fig:benchframe}), in which an informative benchmark should be \emph{discriminative}, separating models with statistical confidence, and \emph{applicable}, evaluating scientifically meaningful tasks across levels of generalization. BGC-Bench is structured around two evaluation axes: \emph{generalization}, assessed through genotype- and phenotype-shift splits, and \emph{functionality}, evaluated through five tasks spanning biological scales from single adenylation (A) domains to whole-cluster BGC--molecule retrieval. Because a BGC can be read at several granularities, BGC-Bench evaluates twelve encoders that read the same clusters as nucleotide, amino-acid, or Pfam-domain sequences~\citep{xiao2025cross}, including simple frequency-based baselines based on domain counts or sequence k-mer frequencies, and pairs them with chemical encoders for cross-modal retrieval. We further evaluate models in an active-learning setting, reflecting a central motivation for biological models: prioritizing wet-lab characterization under limited experimental budgets~\citep{Hie2020Uncertainty,Biswas2021LowN,Jiang2024EVOLVEpro}.

BGC-Bench discriminates among existing models and shows that no single encoder dominates. The best representation granularity depends on the biological scale of the task: Pfam-domain encoders lead on cluster-level tasks, whereas amino-acid encoders lead on A-domain substrate prediction. Pretrained encoders do not consistently outperform other representations across tasks and splits; simple frequency-based baselines rank first in six of fifteen task--split settings, and larger models are not reliably better. On cross-modal retrieval, a frequency-based Pfam baseline paired with substructure fingerprints, both free of pretraining, outperforms learned BGC encoders, suggesting that coarse-grained BGC composition already captures a strong signal for BGC--molecule matching and pointing to a gap in how current sequence representations align with chemical space. Active learning shows that most tasks remain data-limited and that encoder rankings depend on the labeling budget, with pretrained encoders leading at small budgets when paired with their selected acquisition strategy, while acquisition strategies yield modest gains over random sampling. These findings support the view that in data-scarce scientific domains, meaningful evaluation depends not only on developing stronger models but also on rigorous, mechanistically grounded benchmark design that extracts more value from existing datasets. They also motivate the development of general-purpose models that integrate multiple modalities and biological granularities for BGCs and biology more broadly~\citep{xiao2025cross,Cui2025MultimodalFoundationModels}.

Our main contributions are:

\begin{itemize}
    \item \textbf{BGC-Bench}, a curated, extensible evaluation suite for BGC representations, spanning five biologically meaningful tasks, genotype- and phenotype-shift splits, and active-learning settings, with labels and evaluation protocols derived using established bio- and cheminformatics tools. We release the evaluation pipeline, with which future users can easily add new tasks and encoders.

    \item A \textbf{systematic multi-granularity and multimodal evaluation} of twelve BGC encoders across nucleotide, amino-acid, and Pfam-domain representations, including frequency-based baselines, paired with seven chemical encoders for cross-modal retrieval. Our findings reveal how performance varies with task scale, generalization regime, and labeling budget, and show that pretrained models do not consistently outperform simpler frequency-based baselines.

    \item A \textbf{generalizable benchmark-design framework} for data-scarce scientific ML, providing a taxonomic guide for constructing benchmarks that are discriminative and reflect downstream utility, which we also use to map existing benchmarks in other domains and expose evaluation gaps.
\end{itemize}

\section{Related Work}
\label{sec:related}

\paragraph{BGC modeling.} Computational BGC analysis has evolved from rule- and HMM-based genome-mining systems such as ClusterFinder and antiSMASH~\citep{cimermancic2014clusterfinder,antismash8} to self-supervised foundation models~\citep{bigcarp,kawano2026roberta}. The BGC models evaluated in this work are summarized by model family in Table~\ref{tab:model-inventory}. Recent work has also studied BGC--natural-product retrieval~\citep{liu2025unified,proscope}, as well as linking BGCs to their mass-spectrometry-detected products~\citep{Schorn2021PairedOmics,Louwen2023NPLinker}. DeepBGC~\citep{deepbgc} uses random cross-validation and BiGCARP~\citep{bigcarp} a zero-shot MIBiG evaluation with coarse class labels. BGCat~\citep{porokhin2025bgcat} predicts finer NPClassifier labels and evaluates temporal and GCF-held-out splits, while \citet{liu2025unified} evaluate class-held-out and temporal retrieval. BGC-Bench brings classification, regression, retrieval, substrate prediction, and active learning into one shared evaluation across genotype and phenotype shifts.

\paragraph{Benchmarks for biological and chemical representation learning.} Protein benchmarks such as TAPE~\citep{rao2019tape}, PEER~\citep{xu2022peer}, FLIP~\citep{dallago2021flip}, and ProteinGym~\citep{notin2023proteingym} assess sequence representations and fitness prediction under biologically meaningful splits; molecular benchmarks such as MoleculeNet~\citep{wu2018moleculenet} and Lo-Hi~\citep{steshin2023lohi} emphasize drug-discovery-oriented evaluation with scaffold-based and similarity-thresholded splits, respectively; MassSpecGym~\citep{bushuiev2024massspecgym} standardizes MS/MS-to-molecule evaluation; and a recent DNA foundation-model benchmark shows that pooling choices, task definitions, and specialized baselines can materially change conclusions~\citep{feng2025dnafm}. BGC-Bench follows the evaluation principles of these adjacent efforts, makes them explicit as a framework, and adapts them to the data scarcity of BGCs.

\begin{table}[!h]
  \caption{\textbf{Model inventory.} The twelve BGC encoders, grouped by input granularity; each group starts with its parameter-free frequency-based baselines. For BGC-MAC, BGCat, BGC-Prophet, and LC-PLM, residue representations are averaged within each CDS, then across CDSs. (CDS: coding sequence; MLM: masked language model; LM: language model.)}
  \label{tab:model-inventory}
  \centering
  \footnotesize
  \setlength{\tabcolsep}{4pt}
  \newcommand{\mref}[1]{\mbox{\citep{#1}}}
  \newcommand{\modelgroup}[1]{%
    \addlinespace[5pt]
    \multicolumn{4}{@{}l}{\textit{#1}}\\[-1pt]
  }
  \begin{tabularx}{\textwidth}{@{}
    >{\raggedright\arraybackslash\hangindent=0.8em\hangafter=1}p{127pt}
    >{\raggedleft\arraybackslash}p{19pt}
    >{\raggedleft\arraybackslash}p{27pt}
    Y@{}}
    \toprule
    \textbf{Model} & \textbf{Dim.} & \textbf{Params} & \textbf{Representation details} \\
    \midrule
    \modelgroup{Pfam-domain encoders}
    Bag-of-Pfam                      & 3{,}932 & --           & Pfam-domain count vector, $L_2$ normalized \\
    Random-Pfam                      & 256     & --           & Random per-Pfam-token vectors, normalized after pooling \\
    BiGCARP \mref{bigcarp}           & 256     & $\sim$34\,M  & ByteNet MLM on antiSMASH-DB~\citep{blin2019antismashdb2} BGCs \\
    Kawano \mref{kawano2026roberta}  & 1{,}024 & $\sim$110\,M & RoBERTa MLM on antiSMASH-DB bacterial BGCs \\
    DeepBGC \mref{deepbgc}           & 256     & $\sim$1.6\,M & Frozen Pfam2vec + BiLSTM; BGC-detection pretraining \\
    \modelgroup{Amino-acid encoders}
    Amino-acid 3-mer                 & 8{,}000 & --           & 3-mer frequencies within each CDS, pooled over the cluster \\
    BGC-MAC \mref{lu2026bgcmac}      & 1{,}280 & $\sim$650\,M & ESM-2~\citep{lin2023esm2} (650\,M) protein MLM \\
    BGCat \mref{porokhin2025bgcat}   & 1{,}152 & $\sim$600\,M & ESM~C~\citep{candido2026esmc} (600\,M) protein MLM \\
    BGC-Prophet \mref{bgcprophet}    & 320     & $\sim$8\,M   & ESM-2~\citep{lin2023esm2} (8\,M) protein MLM \\
    LC-PLM \mref{wang2025lcplm}      & 1{,}536 & $\sim$750\,M & BiMamba-S protein MLM on UniRef50 \\
    \modelgroup{Nucleotide encoders}
    Nucleotide 6-mer                 & 2{,}080 & --           & Strand-collapsed 6-mer frequencies of the cluster DNA \\
    Evo 2 7B \mref{brixi2026evo2}    & 4{,}096 & 7\,B         & Autoregressive StripedHyena 2 DNA LM on OpenGenome2; block-28 states averaged over the cluster DNA (48\,kb windows) \\
    \bottomrule
  \end{tabularx}
\end{table}

\section{BGC-Bench}
\label{sec:bm}

\subsection{Model inventory}
\label{sec:bm:models}
A BGC can be read as a sequence of nucleotides, amino acids, genes, or Pfam domains, which makes it an unusual arena in which encoders of different granularities can be compared on the same samples. BGC-Bench evaluates twelve encoders at three of these granularities
(Table~\ref{tab:model-inventory}). Five read the Pfam-domain sequence,
five read the amino-acid sequence of each gene, and two read the nucleotide sequence. The encoders range from parameter-free frequency-based baselines to pretrained models of 1.6\,M to 7\,B parameters, trained either on Pfam-domain sequences of BGCs and genomes or on
general protein and DNA corpora. Each
granularity has its own frequency-based baseline, so every pretrained encoder can be compared with the composition of its own input. Backbones, pretraining corpora, and pooling rules are given in Appendix~\ref{app:encoders}.

\subsection{Design framework}
\label{sec:bm:framework}
\begin{wrapfigure}{r}{0.65\textwidth}
  \centering
  \includegraphics[width=\linewidth]{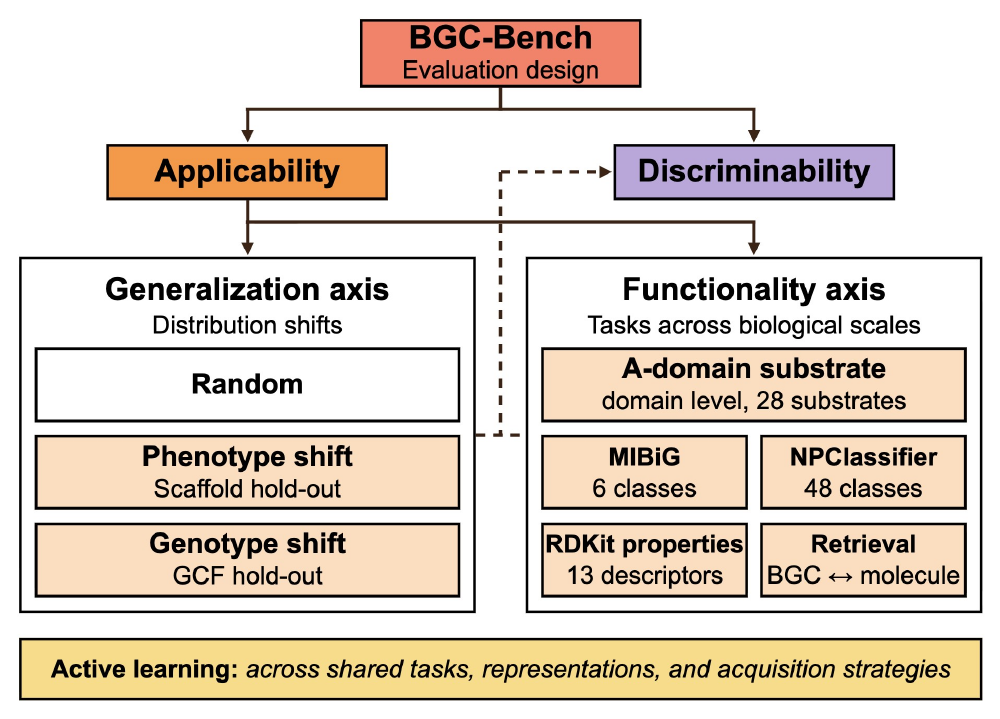}
  \caption{\textbf{BGC-Bench evaluation design framework.} Applicability is characterized along a \textit{generalization} axis (random, phenotype-shift and genotype-shift splits) and a \textit{functionality} axis (tasks from local A-domain substrate prediction to broad BGC--molecule retrieval); performance across these settings is then used to assess \textit{discriminability}, and active learning adds a limited-labeling-budget setting.}
  \label{fig:benchframe}
\end{wrapfigure}
Figure~\ref{fig:benchframe} summarizes our design framework. We propose that an informative benchmark should satisfy two criteria: \emph{discriminative} and \emph{applicable}. A discriminative benchmark should reliably reveal differences in model capabilities, rather than having rankings dominated by task saturation, metric insensitivity, or statistical noise. Still, some scientifically meaningful tasks may offer limited separation among current models, yet remain informative by exposing shared limitations and highlighting opportunities for future progress. An applicable benchmark, meanwhile, should evaluate capabilities that correspond to scientifically meaningful problems.

Applicability can be further decomposed into two aspects. The first is \emph{functionality}: the benchmark should span tasks serving distinct scientific purposes, matching the breadth of use expected of a general-purpose foundation model. The second is \emph{generalization}: although foundation models are expected to acquire general knowledge of the underlying domain, their capabilities are usually bounded by their training distribution, and models generalize to varying extents. It is therefore essential to evaluate performance across generalization levels, ranging from in-distribution to different settings of out-of-distribution prediction.

Finally, we incorporate active learning as an additional evaluation setting. Annotated data remain scarce in many scientific domains, including BGC research. Active learning therefore re-evaluates every task, on its random split, under an iterative labeling protocol, probing a learning dynamic that static accuracy cannot capture. It offers two complementary views: it measures how effectively a model's representations can guide the acquisition of new labels, and it indicates how model performance may scale as more annotated data become available.

Following this framework, we construct BGC-Bench from a compact but carefully curated dataset. The framework itself is domain-agnostic (Section~\ref{sec:discussion} and Table~\ref{tab:bgc-framework}). The following sections describe how each component is instantiated in BGC-Bench.

\subsection{Generalization: genotype vs phenotype}
\label{sec:bm:splits}
We organize the generalization aspect around the classical genotype--phenotype dichotomy in biology. For BGCs, the \textbf{genotype} is the genetic content of the cluster: the ordered set of genes that specify a biosynthetic route. The \textbf{phenotype} is the chemical outcome of that route: the molecular structure of the natural product it yields. Genotype and phenotype represent distinct but mechanistically linked levels of biological variation: similar biosynthetic architectures can yield substantially different products, while distinct architectures can converge on similar chemistry. Generalization along one axis therefore does not always imply generalization along the other; we further examine the relationship between these two forms of distribution shift in Appendix~\ref{app:gp-relationship}.

We instantiate the two shifts with established annotation tools applied to the paired BGC--product data. For \textbf{genotype shift}, we cluster MIBiG BGCs into gene cluster families (GCFs) with BiG-SCAPE~\citep{navarro2020bigscape,draisma2026bigscape2} (v2.0.3), the standard tool for grouping BGCs into families. BiG-SCAPE scores each pair of BGCs by a distance that combines three Pfam-domain measures (the overlap of their domain types, the overlap of their adjacent domain pairs, and the sequence similarity of shared domains), and partitions the resulting network into families by affinity propagation. We then assign a random 80\% of GCFs to training and hold out the remaining 20\% for testing. For \textbf{phenotype shift}, we compute the Bemis--Murcko scaffold~\citep{bemis1996murcko} of each product with RDKit~\citep{rdkit}, which keeps the ring systems and their linkers and removes side chains. BGCs whose first products share the same scaffold form one group, and we hold out 20\% of the scaffold groups in the same way. We chose these tools because they are widely used in BGC and natural-product research; the clustering threshold and a sensitivity analysis of both choices are given in Appendix~\ref{app:split-sensitivity}. For retrieval, we additionally hold out one MIBiG class at a time (class holdout; Appendix~\ref{app:splits}).

\subsection{Functionality: from sequence to molecule}
\label{sec:bm:tasks}
To assess the practical utility of BGC representations, we curate a suite of downstream tasks spanning biological scales from single A domains to broad BGC--molecule retrieval. We introduce each task and its motivation below, with detailed experimental specifications, including dataset sizes, preprocessing, hyperparameters, and evaluation metrics, provided in Appendix~\ref{app:tasks}.

\paragraph{MIBiG coarse class prediction.}
We evaluate multi-label classification over the six biosynthetic classes defined in MIBiG 4.0: PKS, NRPS, Ribosomal, Terpene, Saccharide, and Other~\citep{mibig}. These curated labels provide broad information about the biosynthetic pathways encoded by BGCs, helping identify their underlying biosynthetic machinery and prioritize clusters for further characterization. We include this task as an established reference, as MIBiG class prediction has been widely adopted in previous BGC model evaluations~\citep{deepbgc,bigcarp,lu2026bgcmac}. However, the taxonomy has several limitations. First, its categories are not defined at a uniform level: PKS and NRPS identify specific biosynthetic enzyme systems, whereas Saccharide encompasses a broader category of carbohydrate-related biosynthesis. Second, the taxonomy is non-exhaustive, with a heterogeneous Other category that can coexist with other biosynthetic labels. These coarse labels also provide limited information about the chemical structures and properties of the resulting products. We therefore complement this established task with more product-specific evaluations.

\paragraph{NPClassifier class prediction.} To evaluate whether BGC representations capture information about their chemical products beyond broad biosynthetic mechanisms, we introduce natural-product classification using NPClassifier~\citep{npclassifier}. Unlike the six MIBiG-derived labels, which primarily characterize biosynthetic machinery, NPClassifier assigns structure-informed labels directly from product SMILES. It organizes natural products into a hierarchical taxonomy of 7 pathways, 70 superclasses, and 672 classes. For example, nystatin and spiramycin are both PKS-derived macrolides, yet NPClassifier places them in distinct classes (polyene macrolides and tylosin-type macrolides), associated with antifungal and antibacterial activities, respectively. Such distinctions could help prioritize BGCs for targeted investigation. This task therefore evaluates whether BGC representations capture product-specific chemical distinctions beyond broad biosynthetic categories, providing a more practical measure of their utility for natural-product discovery.

Because a BGC can produce several products, each BGC is labeled with every NPClassifier class of its products, a multi-label task over the $48$ classes carried by at least ten BGCs ($1{,}221$ BGCs; Appendix~\ref{app:task-npc}). We report macro-AUROC weighted by each class's test positives, so that the many classes with only a few test BGCs do not dominate the score. We choose NPClassifier because its taxonomy was built specifically for natural products and is organized by biosynthetic pathway, which makes it widely relevant to natural-product discovery. More general chemical classifiers could be used in principle; we test the sensitivity of our results to this choice by relabeling the same products with ClassyFire~\citep{classyfire}, a general-purpose chemical classifier, in Appendix~\ref{app:label-provenance}.

\paragraph{Chemical property prediction.}
While NPClassifier characterizes products through a hierarchical natural-product taxonomy, chemical property prediction evaluates continuous properties derived directly from molecular structure. We use RDKit~\citep{rdkit} to compute molecular descriptors from product SMILES, a string-based representation encoding the atoms, bonds, and connectivity of a molecule~\citep{weininger1988smiles}. From over $200$ available descriptors, we manually select $13$ representative properties, covering ADMET-related descriptors commonly used in drug-likeness assessment~\citep{lipinski1997,veber2002,bickerton2012qed}, physicochemical properties included in the GuacaMol molecular-design benchmark~\citep{brown2019guacamol}, and properties characteristic of natural products~\citep{feher2003property}. A probe trained on each encoder's frozen embeddings predicts these $13$ descriptors, and performance is measured by the mean Spearman correlation across descriptors. To assess sensitivity to descriptor selection, we additionally evaluate a comprehensive set of $66$ descriptors, obtained by removing descriptors that are constant or rarely non-zero across BGC products, followed by those exhibiting high pairwise correlations. Results of this sensitivity analysis and detailed descriptor definitions are provided in Appendix~\ref{app:tchem-descriptors}.

\paragraph{BGC--molecule retrieval.}
BGC--molecule retrieval evaluates whether a model can identify the specific chemical product of a BGC within a pool of candidate molecules. We formulate this as a multimodal alignment task, analogous to image--text pairing, in which BGC and molecular representations are mapped into a shared embedding space. We evaluate pairings of BGC encoders with molecular representations ranging from ECFP4 fingerprints, which encode local atom environments~\citep{rogers2010ecfp}, to MoLFormer embeddings from a chemical language model pretrained on SMILES strings~\citep{ross2022molformer}. For each pairing, we train alignment heads using a CLIP-style contrastive objective over paired BGC--molecule examples~\citep{radford2021clip} (Appendix~\ref{app:task-retrieval}). We report class-macro R@$10$, which measures how often the correct product appears among the top ten retrieved candidates, averaged across classes. A heatmap summarizes retrieval performance across BGC--chemical encoder pairings (Figure~\ref{fig:t3-chem-matrix}).

\paragraph{A-domain substrate prediction.}
While the other tasks evaluate whole BGCs or their products, adenylation (A) domain substrate prediction examines the sequence--chemistry relationship at a finer, domain-level scale. In nonribosomal peptide synthetases (NRPSs), A domains select and activate specific building blocks incorporated into the peptide product. Using substrate annotations from MIBiG, we match each annotated A domain to its corresponding Pfam AMP-binding (PF00501) hit, yielding 2,504 domains from 529 BGCs (Appendix~\ref{app:adomain}). Each encoder represents the domain at its own sequence granularity: amino-acid encoders use the amino-acid sequence of the Pfam hit; nucleotide encoders use the corresponding coding sequence reconstructed from the parent gene; BiGCARP and Kawano use the contextual state at the AMP-binding token, and Bag-of-Pfam uses the parent cluster's domain composition. To prevent information leakage, we split the data by BGC and report top-1 accuracy across 28 substrate classes, each with at least ten single-substrate examples.

\paragraph{Active learning as an additional evaluation layer.}
In addition, we evaluate the encoders under active learning, which tests how well a representation prioritizes experimental targets and projects how performance scales with more labeled data. We cover all five tasks on their random splits with every encoder available for each task, and compare random acquisition with two strategies: TypiClust~\citep{hacohen2022typiclust}, designed for the scarce-label regime, which queries the most typical point of each under-covered cluster; and BADGE~\citep{ash2020badge}, a standard hybrid that combines uncertainty and diversity through gradient embeddings (not applicable to chemical property regression). For each task, a fixed $20\%$ of the training split is held out for validation and the remaining $80\%$ forms the acquisition pool. Each run starts from $16$ random labels and grows by a factor of $\sqrt{2}$ until the pool is exhausted; $20$ seeds give $3{,}240$ learning curves. We summarize each curve by $\mathrm{AULC}_{\mathrm{low}}$, the area under the test-score curve over $16$--$256$ labels on a $\log_2$ axis, normalized by the range width, and use it for all statistical comparisons. Appendix~\ref{app:al-savings} reports \emph{label savings}, adapted from the labeling efficiency of \citet{beck2021effective}: relative to Bag-of-Pfam with random acquisition, the labels this baseline needs to reach a given score divided by the labels an encoder needs under random acquisition, combined by a geometric mean over three reference scores. Full specifications are given in Appendix~\ref{app:al}.

\begin{table}[!t]
  \caption{\textbf{Main results.} Performance of twelve BGC encoders on five tasks across their generalization splits. R, S, G, and C denote the random, scaffold-holdout, GCF-holdout, and class-holdout splits; CV denotes five-fold cross-validation for A-domain substrate prediction. For retrieval, molecules are represented by frozen embeddings from MoLFormer, a pretrained chemical foundation model. Separability is the fraction of encoder pairs whose exact CIs do not overlap.}
  \label{tab:t3-main}
  \centering
  \footnotesize
  \setlength{\tabcolsep}{1.95pt}
  \renewcommand{\arraystretch}{1.0}
  \begin{tabular}{l*{15}{c}}
  \toprule
  Model & \multicolumn{3}{c}{Coarse classes} & \multicolumn{3}{c}{Fine classes} & \multicolumn{3}{c}{Chem.\ properties} & \multicolumn{4}{c}{BGC--molecule retrieval} & \multicolumn{2}{c}{A-domain substrate} \\
   & \multicolumn{3}{c}{\tiny macro-AUROC} & \multicolumn{3}{c}{\tiny weighted macro-AUROC} & \multicolumn{3}{c}{\tiny mean Spearman $\rho$} & \multicolumn{4}{c}{\tiny class-macro R@10} & \multicolumn{2}{c}{\tiny top-1 accuracy} \\
  \cmidrule(lr){2-4}\cmidrule(lr){5-7}\cmidrule(lr){8-10}\cmidrule(lr){11-14}\cmidrule(lr){15-16}
   & R & S & G & R & S & G & R & S & G & R & S & G & C & CV & G \\
  \midrule
  \multicolumn{16}{l}{\emph{Pfam-domain encoders}} \\
  Bag-of-Pfam & .929 & .877 & .884 & \cellcolor{rankC}.880 & .825 & .753 & .354 & .281 & .283 & \cellcolor{rankA}\textbf{.624} & \cellcolor{rankA}\textbf{.546} & \cellcolor{rankA}\textbf{.495} & \cellcolor{rankA}\textbf{.275} & .196 & .151 \\[-2.5pt]
   & {\tiny$\pm$.020} & {\tiny$\pm$.042} & {\tiny$\pm$.029} & \cellcolor{rankC}{\tiny$\pm$.031} & {\tiny$\pm$.028} & {\tiny$\pm$.041} & {\tiny$\pm$.057} & {\tiny$\pm$.057} & {\tiny$\pm$.054} & \cellcolor{rankA}{\tiny$\pm$.062} & \cellcolor{rankA}{\tiny$\pm$.109} & \cellcolor{rankA}{\tiny$\pm$.081} & \cellcolor{rankA}{\tiny$\pm$.039} & {\tiny$\pm$.018} & {\tiny$\pm$.043} \\
  Random-Pfam & .936 & .909 & .896 & \cellcolor{rankC}.880 & \cellcolor{rankC}.851 & .786 & \cellcolor{rankC}.523 & \cellcolor{rankC}.467 & \cellcolor{rankC}.417 & \cellcolor{rankD}.542 & \cellcolor{rankC}.504 & \cellcolor{rankB}.448 & \cellcolor{rankD}.209 & -- & -- \\[-2.5pt]
   & {\tiny$\pm$.017} & {\tiny$\pm$.026} & {\tiny$\pm$.022} & \cellcolor{rankC}{\tiny$\pm$.027} & \cellcolor{rankC}{\tiny$\pm$.026} & {\tiny$\pm$.033} & \cellcolor{rankC}{\tiny$\pm$.049} & \cellcolor{rankC}{\tiny$\pm$.053} & \cellcolor{rankC}{\tiny$\pm$.061} & \cellcolor{rankD}{\tiny$\pm$.068} & \cellcolor{rankC}{\tiny$\pm$.119} & \cellcolor{rankB}{\tiny$\pm$.105} & \cellcolor{rankD}{\tiny$\pm$.031} &  &  \\
  BiGCARP & \cellcolor{rankB}.952 & \cellcolor{rankB}.935 & \cellcolor{rankD}.909 & \cellcolor{rankC}.880 & \cellcolor{rankA}\textbf{.872} & \cellcolor{rankA}\textbf{.834} & \cellcolor{rankB}.574 & \cellcolor{rankB}.519 & \cellcolor{rankA}\textbf{.509} & \cellcolor{rankB}.559 & .466 & \cellcolor{rankC}.435 & \cellcolor{rankB}.222 & .233 & .106 \\[-2.5pt]
   & \cellcolor{rankB}{\tiny$\pm$.015} & \cellcolor{rankB}{\tiny$\pm$.022} & \cellcolor{rankD}{\tiny$\pm$.024} & \cellcolor{rankC}{\tiny$\pm$.027} & \cellcolor{rankA}{\tiny$\pm$.023} & \cellcolor{rankA}{\tiny$\pm$.031} & \cellcolor{rankB}{\tiny$\pm$.045} & \cellcolor{rankB}{\tiny$\pm$.051} & \cellcolor{rankA}{\tiny$\pm$.056} & \cellcolor{rankB}{\tiny$\pm$.073} & {\tiny$\pm$.121} & \cellcolor{rankC}{\tiny$\pm$.087} & \cellcolor{rankB}{\tiny$\pm$.031} & {\tiny$\pm$.021} & {\tiny$\pm$.038} \\
  Kawano & .934 & \cellcolor{rankC}.926 & \cellcolor{rankC}.911 & .873 & .844 & \cellcolor{rankD}.810 & .455 & .345 & .337 & .532 & .382 & .387 & .187 & .234 & .137 \\[-2.5pt]
   & {\tiny$\pm$.017} & \cellcolor{rankC}{\tiny$\pm$.028} & \cellcolor{rankC}{\tiny$\pm$.020} & {\tiny$\pm$.029} & {\tiny$\pm$.024} & \cellcolor{rankD}{\tiny$\pm$.031} & {\tiny$\pm$.051} & {\tiny$\pm$.054} & {\tiny$\pm$.059} & {\tiny$\pm$.075} & {\tiny$\pm$.107} & {\tiny$\pm$.092} & {\tiny$\pm$.032} & {\tiny$\pm$.020} & {\tiny$\pm$.038} \\
  DeepBGC & \cellcolor{rankA}\textbf{.953} & \cellcolor{rankA}\textbf{.945} & \cellcolor{rankA}\textbf{.922} & \cellcolor{rankA}\textbf{.893} & \cellcolor{rankB}.863 & \cellcolor{rankB}.830 & \cellcolor{rankA}\textbf{.591} & \cellcolor{rankA}\textbf{.558} & \cellcolor{rankB}.482 & .440 & .397 & .407 & .163 & -- & -- \\[-2.5pt]
   & \cellcolor{rankA}{\tiny$\pm$.015} & \cellcolor{rankA}{\tiny$\pm$.023} & \cellcolor{rankA}{\tiny$\pm$.020} & \cellcolor{rankA}{\tiny$\pm$.026} & \cellcolor{rankB}{\tiny$\pm$.022} & \cellcolor{rankB}{\tiny$\pm$.029} & \cellcolor{rankA}{\tiny$\pm$.043} & \cellcolor{rankA}{\tiny$\pm$.045} & \cellcolor{rankB}{\tiny$\pm$.054} & {\tiny$\pm$.082} & {\tiny$\pm$.102} & {\tiny$\pm$.072} & {\tiny$\pm$.026} &  &  \\
  \midrule
  \multicolumn{16}{l}{\emph{Amino-acid encoders}} \\
  Amino-acid 3-mer & .920 & .913 & .863 & \cellcolor{rankB}.887 & \cellcolor{rankD}.850 & \cellcolor{rankC}.822 & \cellcolor{rankD}.506 & \cellcolor{rankD}.428 & .332 & \cellcolor{rankC}.549 & .418 & .318 & .198 & \cellcolor{rankA}\textbf{.781} & \cellcolor{rankA}\textbf{.747} \\[-2.5pt]
   & {\tiny$\pm$.021} & {\tiny$\pm$.029} & {\tiny$\pm$.036} & \cellcolor{rankB}{\tiny$\pm$.027} & \cellcolor{rankD}{\tiny$\pm$.029} & \cellcolor{rankC}{\tiny$\pm$.037} & \cellcolor{rankD}{\tiny$\pm$.055} & \cellcolor{rankD}{\tiny$\pm$.050} & {\tiny$\pm$.062} & \cellcolor{rankC}{\tiny$\pm$.085} & {\tiny$\pm$.115} & {\tiny$\pm$.090} & {\tiny$\pm$.036} & \cellcolor{rankA}{\tiny$\pm$.018} & \cellcolor{rankA}{\tiny$\pm$.051} \\
  BGC-MAC & \cellcolor{rankC}.948 & .920 & \cellcolor{rankB}.917 & .853 & .812 & .763 & .466 & .395 & .355 & .507 & \cellcolor{rankB}.507 & \cellcolor{rankD}.421 & \cellcolor{rankC}.220 & \cellcolor{rankB}.759 & \cellcolor{rankC}.736 \\[-2.5pt]
   & \cellcolor{rankC}{\tiny$\pm$.014} & {\tiny$\pm$.021} & \cellcolor{rankB}{\tiny$\pm$.019} & {\tiny$\pm$.039} & {\tiny$\pm$.030} & {\tiny$\pm$.039} & {\tiny$\pm$.057} & {\tiny$\pm$.051} & {\tiny$\pm$.062} & {\tiny$\pm$.078} & \cellcolor{rankB}{\tiny$\pm$.108} & \cellcolor{rankD}{\tiny$\pm$.079} & \cellcolor{rankC}{\tiny$\pm$.033} & \cellcolor{rankB}{\tiny$\pm$.019} & \cellcolor{rankC}{\tiny$\pm$.055} \\
  BGCat & .930 & .913 & .891 & .847 & .789 & .750 & .423 & .399 & .366 & .469 & \cellcolor{rankD}.485 & .373 & .201 & \cellcolor{rankC}.758 & \cellcolor{rankD}.733 \\[-2.5pt]
   & {\tiny$\pm$.019} & {\tiny$\pm$.020} & {\tiny$\pm$.027} & {\tiny$\pm$.039} & {\tiny$\pm$.034} & {\tiny$\pm$.039} & {\tiny$\pm$.055} & {\tiny$\pm$.052} & {\tiny$\pm$.058} & {\tiny$\pm$.077} & \cellcolor{rankD}{\tiny$\pm$.056} & {\tiny$\pm$.074} & {\tiny$\pm$.037} & \cellcolor{rankC}{\tiny$\pm$.020} & \cellcolor{rankD}{\tiny$\pm$.051} \\
  BGC-Prophet & .939 & .913 & .902 & .854 & .833 & .763 & .500 & .419 & \cellcolor{rankD}.412 & .461 & .431 & .406 & .189 & .656 & .610 \\[-2.5pt]
   & {\tiny$\pm$.016} & {\tiny$\pm$.020} & {\tiny$\pm$.023} & {\tiny$\pm$.031} & {\tiny$\pm$.027} & {\tiny$\pm$.037} & {\tiny$\pm$.046} & {\tiny$\pm$.055} & \cellcolor{rankD}{\tiny$\pm$.056} & {\tiny$\pm$.078} & {\tiny$\pm$.124} & {\tiny$\pm$.100} & {\tiny$\pm$.026} & {\tiny$\pm$.020} & {\tiny$\pm$.057} \\
  LC-PLM & \cellcolor{rankD}.940 & \cellcolor{rankD}.921 & .905 & .843 & .810 & .701 & .407 & .304 & .265 & .250 & .250 & .184 & .137 & \cellcolor{rankD}.745 & \cellcolor{rankB}.740 \\[-2.5pt]
   & \cellcolor{rankD}{\tiny$\pm$.015} & \cellcolor{rankD}{\tiny$\pm$.021} & {\tiny$\pm$.025} & {\tiny$\pm$.036} & {\tiny$\pm$.026} & {\tiny$\pm$.048} & {\tiny$\pm$.051} & {\tiny$\pm$.053} & {\tiny$\pm$.058} & {\tiny$\pm$.083} & {\tiny$\pm$.100} & {\tiny$\pm$.067} & {\tiny$\pm$.034} & \cellcolor{rankD}{\tiny$\pm$.018} & \cellcolor{rankB}{\tiny$\pm$.055} \\
  \midrule
  \multicolumn{16}{l}{\emph{Nucleotide encoders}} \\
  Nucleotide 6-mer & .808 & .773 & .701 & .834 & .778 & .740 & .153 & .150 & .033 & .299 & .163 & .150 & .151 & .597 & .466 \\[-2.5pt]
   & {\tiny$\pm$.030} & {\tiny$\pm$.034} & {\tiny$\pm$.038} & {\tiny$\pm$.029} & {\tiny$\pm$.030} & {\tiny$\pm$.036} & {\tiny$\pm$.057} & {\tiny$\pm$.046} & {\tiny$\pm$.065} & {\tiny$\pm$.066} & {\tiny$\pm$.124} & {\tiny$\pm$.059} & {\tiny$\pm$.034} & {\tiny$\pm$.022} & {\tiny$\pm$.058} \\
  Evo 2 7B & .896 & .857 & .858 & .801 & .750 & .751 & .442 & .409 & .355 & .183 & .128 & .161 & .116 & .562 & .527 \\[-2.5pt]
   & {\tiny$\pm$.022} & {\tiny$\pm$.029} & {\tiny$\pm$.033} & {\tiny$\pm$.034} & {\tiny$\pm$.033} & {\tiny$\pm$.037} & {\tiny$\pm$.050} & {\tiny$\pm$.050} & {\tiny$\pm$.057} & {\tiny$\pm$.063} & {\tiny$\pm$.044} & {\tiny$\pm$.070} & {\tiny$\pm$.033} & {\tiny$\pm$.023} & {\tiny$\pm$.058} \\
  \midrule
  Separability & .27 & .33 & .24 & .11 & .29 & .30 & .45 & .56 & .44 & .45 & .33 & .41 & .36 & .80 & .76 \\
  \bottomrule
  \end{tabular}
\end{table}

\subsection{Experimental setup}
\label{sec:setup}

We adopt five protocol choices for consistent evaluation across encoders. First, BGC-level embeddings are mean-pooled over the full BGC, individual CDSs, or nucleotide windows for Pfam-domain, amino-acid, and nucleotide encoders, respectively, reflecting their different context-length constraints. For A-domain substrate prediction, embeddings are instead constructed at the A-domain level where possible, while Bag-of-Pfam uses the composition of the parent BGC. Second, all pretrained encoders are kept frozen, because fine-tuning through region-pooled representations is complicated and admits many different approaches; continued-pretraining results for Pfam-domain encoders are reported in Appendix~\ref{app:adaptation}. Third, embeddings are used at their native dimensionalities (Table~\ref{tab:model-inventory}), with results after PCA reduction to 256 dimensions reported in Appendix~\ref{app:pca}. Fourth, all tasks use linear readouts, avoiding per-encoder tuning of downstream architectures. We additionally evaluate two-layer MLP probes to test sensitivity to readout capacity (Appendix~\ref{app:probe-mlp}). Finally, we report 95\% percentile bootstrap confidence intervals from 500 resamples of the held-out examples (Appendix~\ref{app:variance}).

\section{Results}
\label{sec:results}

\paragraph{A benchmark of applicability and discriminability.} The main benchmark results across tasks, splits, and encoders are presented in Table~\ref{tab:t3-main}. In terms of applicability, we introduce five practical tasks spanning different biological scales, from local A-domain substrate prediction to broad BGC--molecule retrieval (Section~\ref{sec:bm:tasks}). We further investigate model generalization through genotype- and phenotype-shift splits, constructed using GCF and chemical scaffold clustering, respectively (Section~\ref{sec:bm:splits}). These tasks and splits provide a discriminative benchmark, revealing substantial performance differences among the twelve encoders. To quantify this discriminability, we adopt the separability metric of \citet{li2025arenahard}, defined as the fraction of encoder pairs whose 95\% confidence intervals do not overlap, providing a conservative and interpretable measure of distinguishable performance. Across the fifteen task--split combinations, separability ranges from 0.11 to 0.80, with 39\% of all encoder pairs (371 of 948) distinguishable under this criterion.

\paragraph{Models perform differently across scales and tasks.} No single encoder leads across all tasks. On BGC-level tasks, including class prediction, chemical property prediction, and BGC--molecule retrieval, the strongest methods use Pfam-domain representations, whereas A-domain substrate prediction favors amino-acid-level representations. This suggests that the optimal representation granularity depends on the biological scale of the task. Pretrained encoders also do not consistently outperform other representations across tasks and splits, and simple frequency-based baselines rank first in six of the fifteen task--split settings. Larger models are not reliably better, with Evo~2 7B, the largest encoder, ranking among the weakest on class prediction and retrieval. This agrees with a previous evaluation showing that the benefits of foundation models are task-dependent~\citep{feng2025dnafm}.

\begin{wrapfigure}{r}{0.475\textwidth}
  \centering
  \includegraphics{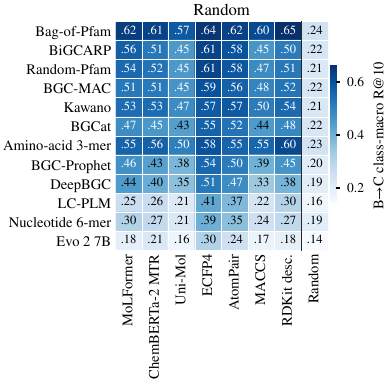}
  \caption{\textbf{BGC--chemical encoder retrieval, random split.} Class-macro R@10 for the twelve
    BGC encoders (rows) paired with seven chemical encoders and a random-vector control (columns),
    mean of three seeds.}
  \label{fig:t3-chem-matrix}
\end{wrapfigure}

\paragraph{Retrieval depends on both BGC and chemical encoders.} Figure~\ref{fig:t3-chem-matrix} pairs the twelve BGC encoders with seven chemical encoders and a random-vector control on the random split; Figure~\ref{fig:t3-chem-matrix-full} shows all four splits. Both sides contribute to retrieval performance, but the choice of BGC encoder has a much larger effect, explaining 78--92\% of the variance across the matrix on each split compared with 2--12\% for the chemical encoder. Interestingly, parameter-free representations perform best on both sides: Bag-of-Pfam is the strongest BGC encoder with every chemical encoder on every split, and descriptor-based chemical encoders perform better on average than the learned MoLFormer, ChemBERTa-2~\citep{ahmad2022chemberta2}, and Uni-Mol~\citep{zhou2023unimol} on most splits. Bag-of-Pfam paired with RDKit descriptors is the best-performing combination on all four splits. The same retrieval framework can naturally extend to other product modalities, such as mass spectra, providing additional targets for evaluating BGC representations.

\begin{figure}[!t]
  \centering
  \includegraphics{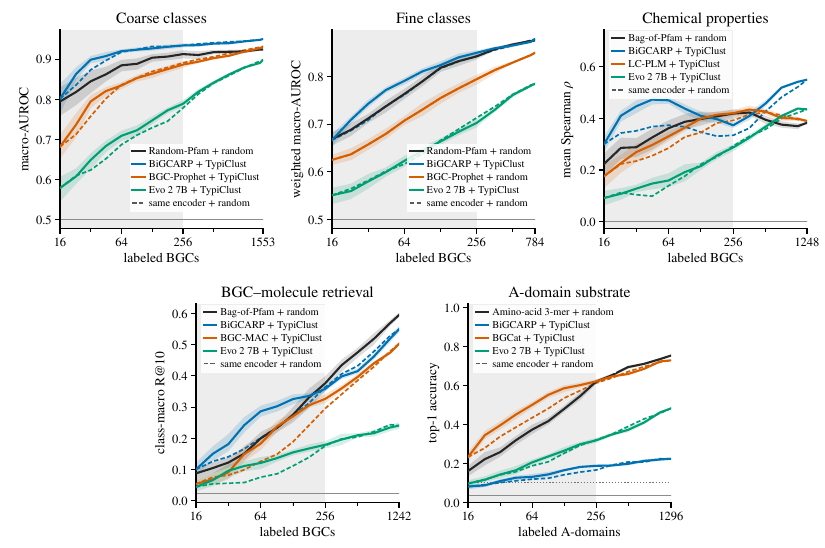}
  \caption{\textbf{Active-learning curves.} Performance against the number of labeled training examples on five tasks. Each panel shows the frequency-based baseline with the highest validation $\mathrm{AULC}_{\mathrm{low}}$ under random acquisition, and BiGCARP, Evo~2, and the best protein language model; dashed lines show the same encoder with random acquisition. Bands show 95\% bootstrap confidence intervals, and gray shading marks the $\mathrm{AULC}_{\mathrm{low}}$ range ($n \le 256$). Thin gray lines mark chance level, and the dotted line in the A-domain panel marks the accuracy of always predicting the majority substrate.}
  \label{fig:al-main}
\end{figure}

\paragraph{Active learning highlights data scarcity and acquisition strategy differences.} The active-learning curves in Figure~\ref{fig:al-main} reveal three main patterns. First, most tasks remain data-limited: except for coarse class prediction, which largely saturates, performance mostly continues to improve up to the full pool. Second, encoders already differ substantially at low budgets, with gaps of 0.13--0.31 at 64 labels under random acquisition, and their ranking depends on the available budget. With each encoder's validation-selected acquisition strategy, pretrained encoders lead all five tasks at 32 labels, whereas Bag-of-Pfam leads retrieval with the full label pool. Third, acquisition strategies provide modest gains (Appendix~\ref{app:al-gain}). TypiClust significantly improves $\mathrm{AULC}_{\mathrm{low}}$ over random acquisition on four of the five tasks, with the largest gain on retrieval (0.040 averaged across encoders), while BADGE significantly improves only coarse class prediction.

\section{Discussion}
\label{sec:discussion}

\paragraph{A general framework for diagnostic evaluation.}
BGC-Bench provides a concrete instantiation of the benchmark framework introduced in Figure~\ref{fig:benchframe}, covering both discriminability and applicability by evaluating models across multiple tasks and under different levels of generalization. More broadly, the same framework can be applied beyond BGCs. Table~\ref{tab:bgc-framework} maps existing benchmarks and tasks from other domains onto the same set of evaluation dimensions. This reframing offers two benefits: first, it enables systematic diagnosis of individual benchmarks by making their design choices explicit along common axes; second, it enables comparison across benchmarks, revealing recurring blind spots and biases that are hard to identify when each benchmark is considered in isolation. For example, MS/MS benchmarks emphasize generalization to unseen molecular structures, while generalization to novel spectra or acquisition conditions remains less systematically evaluated.

\begin{table}[!t]
\caption{\textbf{Representative biochemical benchmarks under the framework.}
Existing benchmarks are organized by domain and decomposed along genotype-side and phenotype-side generalization axes, with genotype and phenotype used in a broad sense: the structure of an entity and its measured function or readout. For each domain, we summarize split definitions, associated functionalities, and representative benchmark datasets.}
\label{tab:bgc-framework}
\centering
\footnotesize
\setlength{\tabcolsep}{4pt}
\newcommand{\bmitem}[1]{\par\noindent\hangindent=0.8em\hangafter=1 #1}
\begin{tabularx}{\textwidth}{@{}
  >{\raggedright\arraybackslash\hsize=0.80\hsize}X
  >{\raggedright\arraybackslash\hsize=0.92\hsize}X
  >{\raggedright\arraybackslash\hsize=0.86\hsize}X
  >{\raggedright\arraybackslash\hsize=1.18\hsize}X
  >{\raggedright\arraybackslash\hsize=1.24\hsize}X@{}}
\toprule
 & \multicolumn{2}{c}{\textbf{Generalization splits}} & & \\
\cmidrule(lr){2-3}
\textbf{Domain} & \textbf{Genotype side} & \textbf{Phenotype side} & \textbf{Functionality} & \textbf{Benchmarks} \\
\midrule
Small-molecule drug discovery
  & Scaffold, similarity-thresholded
  & Assay, target family
  & Property prediction, ADMET prediction, compound--target retrieval, hit identification, lead optimization
  & \bmitem{MoleculeNet~\citep{wu2018moleculenet}}\bmitem{Lo-Hi~\citep{steshin2023lohi}} \\
\addlinespace[4pt]
Protein sequence modeling
  & Sequence identity, family/superfamily, mutation distance
  & Function class, substrate/binding, condition
  & Function annotation, fitness prediction, binding retrieval, protein engineering
  & \bmitem{TAPE~\citep{rao2019tape}}\bmitem{FLIP~\citep{dallago2021flip}}\bmitem{PEER~\citep{xu2022peer}}\bmitem{ProteinGym~\citep{notin2023proteingym}} \\
\addlinespace[4pt]
MS/MS metabolomics
  & Scaffold, MCES distance, structure similarity
  & Spectral similarity, instrument, adduct, collision energy
  & Formula prediction, molecule retrieval, de novo structure generation
  & \bmitem{MassSpecGym~\citep{bushuiev2024massspecgym}} \\
\bottomrule
\end{tabularx}
\end{table}

\paragraph{A shared testbed across granularities and modalities.}
BGC-Bench provides a unique testbed for evaluating models across biological granularities and modalities on the same BGCs, while demonstrating how informative and diverse evaluations can be constructed in data-scarce settings. Nucleotide, amino-acid, gene, and Pfam-domain sequences describe the same underlying biology, yet models at each of these levels have largely been developed and evaluated independently. A BGC sits at an intermediate scale, spanning many genes but far shorter than a genome, so encoders from every level can be applied to it and compared on tasks ranging from single A domains to whole clusters. Despite the scarcity of annotated data, varying how the data are split and which aspect of each BGC is labeled yields a diverse set of tasks from a single curated corpus. As models in each domain continue to develop, more encoders, particularly long-context amino-acid and nucleotide models, can be added to the benchmark directly, and collecting additional data modalities would further extend its scope.

\paragraph{Limitations.}
BGC-Bench is built on MIBiG, which remains relatively small despite being the principal curated resource for BGCs, limiting statistical power and potentially biasing evaluation toward well-characterized taxa and biosynthetic classes. Several labels and splits are derived using computational tools, including NPClassifier, RDKit, and BiG-SCAPE, and therefore inherit their underlying assumptions, although we assess sensitivity to the choice of chemical classifier and descriptor set. Finally, BGC-Bench covers only twelve encoders, does not yet include experimentally important modalities such as MS/MS spectra, and focuses on representation quality via classification, regression, and retrieval. Genome-level tasks (BGC detection, boundary delineation) and generative tasks (product structure generation, de novo BGC design) remain outside its current scope.

\section{Conclusion}
\label{sec:conclusion}

BGC-Bench provides a curated benchmark for evaluating representations of biosynthetic gene clusters under the constraints of limited annotated data. By deriving multiple tasks from a common dataset and constructing genotype- and phenotype-shift splits with established computational tools, it assesses both the applicability of models across tasks and their generalization under distribution shift, while discriminating among encoders with statistical confidence. The benchmark spans multiple biological granularities, from nucleotide and protein sequences to Pfam-domain representations, as well as multiple modalities through paired BGC and chemical representations, and further evaluates models under active learning with limited labeling budgets. Across these settings, model performance varies substantially with task scale and generalization regime, and pretrained encoders do not consistently outperform simple frequency-based baselines. More broadly, BGC-Bench instantiates a general benchmark-design framework that can be applied beyond BGCs to systematically characterize what benchmarks measure, how they test generalization, and where important evaluation gaps remain.

\clearpage
\begin{ack}
This work was supported by the Engineering and Physical Sciences Research Council [EP/Y030826/1]. The authors acknowledge the use of resources provided by the Isambard-AI National AI Research Resource (AIRR)~\citep{mcintoshsmith2024isambardai}. Isambard-AI is operated by the University of Bristol and is funded by the UK Government's Department for Business, Innovation, Science and Trade (DBIST) via UK Research and Innovation; and the Science and Technology Facilities Council [ST/AIRR/I-A-I/1023]. The authors declare no competing interests.
\end{ack}

\medskip
{\small
\bibliographystyle{unsrtnat}
\bibliography{ref}

\begin{thebibliography}{77}
\providecommand{\natexlab}[1]{#1}
\providecommand{\url}[1]{\texttt{#1}}
\expandafter\ifx\csname urlstyle\endcsname\relax
  \providecommand{\doi}[1]{doi: #1}\else
  \providecommand{\doi}{doi: \begingroup \urlstyle{rm}\Url}\fi

\bibitem[Medema et~al.(2015)Medema, Kottmann, Yilmaz, Cummings, Biggins, Blin,
  de~Bruijn, Chooi, Claesen, Coates, Cruz-Morales, Duddela, D{\"u}sterhus,
  Edwards, Fewer, Garg, Geiger, Gomez-Escribano, Greule, Hadjithomas, Haines,
  Helfrich, Hillwig, Ishida, Jones, Jones, Jungmann, Kegler, Kim, K{\"o}tter,
  Krug, Masschelein, Melnik, Mantovani, Monroe, Moore, Moss, N{\"u}tzmann, Pan,
  Pati, Petras, Reen, Rosconi, Rui, Tian, Tobias, Tsunematsu, Wiemann, Wyckoff,
  Yan, Yim, Yu, Xie, Aigle, Apel, Balibar, Balskus, Barona-G{\'o}mez,
  Bechthold, Bode, Borriss, Brady, Brakhage, Caffrey, Cheng, Clardy, Cox,
  De~Mot, Donadio, Donia, van~der Donk, Dorrestein, Doyle, Driessen,
  Ehling-Schulz, Entian, Fischbach, Gerwick, Gerwick, Gross, Gust, Hertweck,
  H{\"o}fte, Jensen, Ju, Katz, Kaysser, Klassen, Keller, Kormanec, Kuipers,
  Kuzuyama, Kyrpides, Kwon, Lautru, Lavigne, Lee, Linquan, Liu, Liu,
  Luzhetskyy, Mahmud, Mast, M{\'e}ndez, Mets{\"a}-Ketel{\"a}, Micklefield,
  Mitchell, Moore, Moreira, M{\"u}ller, Neilan, Nett, Nielsen, O'Gara, Oikawa,
  Osbourn, Osburne, Ostash, Payne, Pernodet, Petricek, Piel, Ploux,
  Raaijmakers, Salas, Schmitt, Scott, Seipke, Shen, Sherman, Sivonen, Smanski,
  Sosio, Stegmann, S{\"u}ssmuth, Tahlan, Thomas, Tang, Truman, Viaud, Walton,
  Walsh, Weber, van Wezel, Wilkinson, Willey, Wohlleben, Wright, Ziemert,
  Zhang, Zotchev, Breitling, Takano, and Gl{\"o}ckner]{Medema2015MIBiG}
Marnix~H. Medema, Renzo Kottmann, Pelin Yilmaz, Matthew Cummings, John~B.
  Biggins, Kai Blin, Irene de~Bruijn, Yit~Heng Chooi, Jan Claesen, R.~Cameron
  Coates, Pablo Cruz-Morales, Srikanth Duddela, Stephanie D{\"u}sterhus,
  Daniel~J. Edwards, David~P. Fewer, Neha Garg, Christoph Geiger, Juan~Pablo
  Gomez-Escribano, Anja Greule, Michalis Hadjithomas, Anthony~S. Haines, Eric
  J.~N. Helfrich, Matthew~L. Hillwig, Keishi Ishida, Adam~C. Jones, Carla~S.
  Jones, Katrin Jungmann, Carsten Kegler, Hyun~Uk Kim, Peter K{\"o}tter, Daniel
  Krug, Joleen Masschelein, Alexey~V. Melnik, Simone~M. Mantovani, Emily~A.
  Monroe, Marcus Moore, Nathan Moss, Hans-Wilhelm N{\"u}tzmann, Guohui Pan,
  Amrita Pati, Daniel Petras, F.~Jerry Reen, Federico Rosconi, Zhe Rui, Zhenhua
  Tian, Nicholas~J. Tobias, Yuta Tsunematsu, Philipp Wiemann, Elizabeth
  Wyckoff, Xiaohui Yan, Grace Yim, Fengan Yu, Yunchang Xie, Bertrand Aigle,
  Alexander~K. Apel, Carl~J. Balibar, Emily~P. Balskus, Francisco
  Barona-G{\'o}mez, Andreas Bechthold, Helge~B. Bode, Rainer Borriss, Sean~F.
  Brady, Axel~A. Brakhage, Patrick Caffrey, Yi-Qiang Cheng, Jon Clardy,
  Russell~J. Cox, Ren{\'e} De~Mot, Stefano Donadio, Mohamed~S. Donia,
  Wilfred~A. van~der Donk, Pieter~C. Dorrestein, Sean Doyle, Arnold J.~M.
  Driessen, Monika Ehling-Schulz, Karl-Dieter Entian, Michael~A. Fischbach,
  Lena Gerwick, William~H. Gerwick, Harald Gross, Bertolt Gust, Christian
  Hertweck, Monica H{\"o}fte, Susan~E. Jensen, Jianhua Ju, Leonard Katz,
  Leonard Kaysser, Jonathan~L. Klassen, Nancy~P. Keller, Jan Kormanec, Oscar~P.
  Kuipers, Tomohisa Kuzuyama, Nikos~C. Kyrpides, Hyung-Jin Kwon, Sylvie Lautru,
  Rob Lavigne, Chia~Y. Lee, Bai Linquan, Xinyu Liu, Wen Liu, Andriy Luzhetskyy,
  Taifo Mahmud, Yvonne Mast, Carmen M{\'e}ndez, Mikko Mets{\"a}-Ketel{\"a},
  Jason Micklefield, Douglas~A. Mitchell, Bradley~S. Moore, Leonilde~M.
  Moreira, Rolf M{\"u}ller, Brett~A. Neilan, Markus Nett, Jens Nielsen, Fergal
  O'Gara, Hideaki Oikawa, Anne Osbourn, Marcia~S. Osburne, Bohdan Ostash,
  Shelley~M. Payne, Jean-Luc Pernodet, Miroslav Petricek, J{\"o}rn Piel,
  Olivier Ploux, Jos~M. Raaijmakers, Jos{\'e}~A. Salas, Esther~K. Schmitt,
  Barry Scott, Ryan~F. Seipke, Ben Shen, David~H. Sherman, Kaarina Sivonen,
  Michael~J. Smanski, Margherita Sosio, Evi Stegmann, Roderich~D. S{\"u}ssmuth,
  Kapil Tahlan, Christopher~M. Thomas, Yi~Tang, Andrew~W. Truman, Muriel Viaud,
  Jonathan~D. Walton, Christopher~T. Walsh, Tilmann Weber, Gilles~P. van Wezel,
  Barrie Wilkinson, Joanne~M. Willey, Wolfgang Wohlleben, Gerard~D. Wright,
  Nadine Ziemert, Changsheng Zhang, Sergey~B. Zotchev, Rainer Breitling, Eriko
  Takano, and Frank~Oliver Gl{\"o}ckner.
\newblock Minimum information about a biosynthetic gene cluster.
\newblock \emph{Nature Chemical Biology}, 11\penalty0 (9):\penalty0 625--631,
  2015.
\newblock \doi{10.1038/nchembio.1890}.

\bibitem[Polturak and Osbourn(2021)]{Polturak2021PlantBGCs}
Guy Polturak and Anne Osbourn.
\newblock The emerging role of biosynthetic gene clusters in plant defense and
  plant interactions.
\newblock \emph{PLOS Pathogens}, 17\penalty0 (7):\penalty0 e1009698, 2021.
\newblock \doi{10.1371/journal.ppat.1009698}.

\bibitem[Pham et~al.(2019)Pham, Yilma, Feliz, Majid, Maffetone, Walker, Kim,
  Cho, Reynolds, Song, Park, and Yoon]{Pham2019MicrobialNaturalProducts}
Janette~V. Pham, Mariamawit~A. Yilma, Adriana Feliz, Murtadha~T. Majid,
  Nicholas Maffetone, Jorge~R. Walker, Eunji Kim, Hyo~Je Cho, Jared~M.
  Reynolds, Myoung~Chong Song, Sung~Ryeol Park, and Yeo~Joon Yoon.
\newblock A review of the microbial production of bioactive natural products
  and biologics.
\newblock \emph{Frontiers in Microbiology}, 10:\penalty0 1404, 2019.
\newblock \doi{10.3389/fmicb.2019.01404}.

\bibitem[Zhang et~al.(2023)Zhang, Duan, Jin, Ali, Han, Zhang, Zhang, Zhang, and
  Gu]{Zhang2023Agrochemicals}
Pei Zhang, Chun-Bao Duan, Bing Jin, Abdallah~S. Ali, Xinya Han, Haifeng Zhang,
  Ming-Zhi Zhang, Wei-Hua Zhang, and Yu-Cheng Gu.
\newblock Recent advances in the natural products-based lead discovery for new
  agrochemicals.
\newblock \emph{Advanced Agrochem}, 2\penalty0 (4):\penalty0 324--339, 2023.
\newblock \doi{10.1016/j.aac.2023.09.004}.

\bibitem[Newman and Cragg(2020)]{Newman2020NaturalProducts}
David~J. Newman and Gordon~M. Cragg.
\newblock Natural products as sources of new drugs over the nearly four decades
  from 01/1981 to 09/2019.
\newblock \emph{Journal of Natural Products}, 83\penalty0 (3):\penalty0
  770--803, 2020.
\newblock \doi{10.1021/acs.jnatprod.9b01285}.

\bibitem[Min et~al.(2024)Min, Ross, Sulem, Veyseh, Nguyen, Sainz, Agirre,
  Heintz, and Roth]{Min2023NLPPretrainedLMs}
Bonan Min, Hayley Ross, Elior Sulem, Amir Pouran~Ben Veyseh, Thien~Huu Nguyen,
  Oscar Sainz, Eneko Agirre, Ilana Heintz, and Dan Roth.
\newblock Recent advances in natural language processing via large pre-trained
  language models: A survey.
\newblock \emph{ACM Computing Surveys}, 56\penalty0 (2):\penalty0 1--40, 2024.
\newblock \doi{10.1145/3605943}.

\bibitem[Hayes et~al.(2025)Hayes, Rao, Akin, Sofroniew, Oktay, Lin, Verkuil,
  Tran, Deaton, Wiggert, Badkundri, Shafkat, Gong, Derry, Molina, Thomas, Khan,
  Mishra, Kim, Bartie, Nemeth, Hsu, Sercu, Candido, and Rives]{Hayes2025ESM3}
Thomas Hayes, Roshan Rao, Halil Akin, Nicholas~J. Sofroniew, Deniz Oktay,
  Zeming Lin, Robert Verkuil, Vincent~Q. Tran, Jonathan Deaton, Marius Wiggert,
  Rohil Badkundri, Irhum Shafkat, Jun Gong, Alexander Derry, Raul~S. Molina,
  Neil Thomas, Yousuf~A. Khan, Chetan Mishra, Carolyn Kim, Liam~J. Bartie,
  Matthew Nemeth, Patrick~D. Hsu, Tom Sercu, Salvatore Candido, and Alexander
  Rives.
\newblock Simulating 500 million years of evolution with a language model.
\newblock \emph{Science}, 387\penalty0 (6736):\penalty0 850--858, 2025.
\newblock \doi{10.1126/science.ads0018}.

\bibitem[Nguyen et~al.(2024)Nguyen, Poli, Durrant, Kang, Katrekar, Li, Bartie,
  Thomas, King, Brixi, Sullivan, Ng, Lewis, Lou, Ermon, Baccus,
  Hernandez-Boussard, R{\'e}, Hsu, and Hie]{Nguyen2024Evo}
Eric Nguyen, Michael Poli, Matthew~G. Durrant, Brian Kang, Dhruva Katrekar,
  David~B. Li, Liam~J. Bartie, Armin~W. Thomas, Samuel~H. King, Garyk Brixi,
  Jeremy Sullivan, Madelena~Y. Ng, Ashley Lewis, Aaron Lou, Stefano Ermon,
  Stephen~A. Baccus, Tina Hernandez-Boussard, Christopher R{\'e}, Patrick~D.
  Hsu, and Brian~L. Hie.
\newblock Sequence modeling and design from molecular to genome scale with
  {Evo}.
\newblock \emph{Science}, 386\penalty0 (6723):\penalty0 eado9336, 2024.
\newblock \doi{10.1126/science.ado9336}.

\bibitem[Dalla-Torre et~al.(2025)Dalla-Torre, Gonzalez, Mendoza-Revilla,
  Lopez~Carranza, Grzywaczewski, Oteri, Dallago, Trop, de~Almeida, Sirelkhatim,
  Richard, Skwark, Beguir, Lopez, and
  Pierrot]{DallaTorre2025NucleotideTransformer}
Hugo Dalla-Torre, Liam Gonzalez, Javier Mendoza-Revilla, Nicolas
  Lopez~Carranza, Adam~Henryk Grzywaczewski, Francesco Oteri, Christian
  Dallago, Evan Trop, Bernardo~P. de~Almeida, Hassan Sirelkhatim, Guillaume
  Richard, Marcin Skwark, Karim Beguir, Marie Lopez, and Thomas Pierrot.
\newblock {Nucleotide Transformer}: Building and evaluating robust foundation
  models for human genomics.
\newblock \emph{Nature Methods}, 22\penalty0 (2):\penalty0 287--297, 2025.
\newblock \doi{10.1038/s41592-024-02523-z}.

\bibitem[Brixi et~al.(2026)Brixi, Durrant, Ku, Naghipourfar, Poli, Sun,
  Brockman, Chang, Fanton, Gonzalez, King, Li, Merchant, Nguyen, Ricci-Tam,
  Romero, Schmok, Taghibakhshi, Vorontsov, Yang, Deng, Gorton, Nguyen, Wang,
  Pearce, Simon, Adams, Amador, Ashley, Baccus, Dai, Dillmann, Ermon, Guo,
  Herschl, Ilango, Janik, Lu, Mehta, Mofrad, Ng, Pannu, R{\'e}, St.~John,
  Sullivan, Tey, Viggiano, Zhu, Zynda, Balsam, Collison, Costa,
  Hernandez-Boussard, Ho, Liu, McGrath, Powell, Pinglay, Burke, Goodarzi, Hsu,
  and Hie]{brixi2026evo2}
Garyk Brixi, Matthew~G. Durrant, Jerome Ku, Mohsen Naghipourfar, Michael Poli,
  Gwanggyu Sun, Greg Brockman, Daniel Chang, Alison Fanton, Gabriel~A.
  Gonzalez, Samuel~H. King, David~B. Li, Aditi~T. Merchant, Eric Nguyen, Chiara
  Ricci-Tam, David~W. Romero, Jonathan~C. Schmok, Ali Taghibakhshi, Anton
  Vorontsov, Brandon Yang, Myra Deng, Liv Gorton, Nam Nguyen, Nicholas~K. Wang,
  Michael~T. Pearce, Elana Simon, Etowah Adams, Zachary~J. Amador, Euan~A.
  Ashley, Stephen~A. Baccus, Haoyu Dai, Steven Dillmann, Stefano Ermon, Daniel
  Guo, Michael~H. Herschl, Rajesh Ilango, Ken Janik, Amy~X. Lu, Reshma Mehta,
  Mohammad R.~K. Mofrad, Madelena~Y. Ng, Jaspreet Pannu, Christopher R{\'e},
  John St.~John, Jeremy Sullivan, Joseph Tey, Ben Viggiano, Kevin Zhu, Greg
  Zynda, Daniel Balsam, Patrick Collison, Anthony~B. Costa, Tina
  Hernandez-Boussard, Eric Ho, Ming-Yu Liu, Thomas McGrath, Kimberly Powell,
  Sudarshan Pinglay, Dave~P. Burke, Hani Goodarzi, Patrick~D. Hsu, and Brian~L.
  Hie.
\newblock Genome modelling and design across all domains of life with {Evo 2}.
\newblock \emph{Nature}, 652\penalty0 (8112):\penalty0 1349--1361, 2026.
\newblock \doi{10.1038/s41586-026-10176-5}.

\bibitem[Mistry et~al.(2021)Mistry, Chuguransky, Williams, Qureshi, Salazar,
  Sonnhammer, Tosatto, Paladin, Raj, Richardson, Finn, and
  Bateman]{mistry2021pfam}
Jaina Mistry, Sara Chuguransky, Lowri Williams, Matloob Qureshi, Gustavo~A.
  Salazar, Erik L.~L. Sonnhammer, Silvio C.~E. Tosatto, Lisanna Paladin, Shriya
  Raj, Lorna~J. Richardson, Robert~D. Finn, and Alex Bateman.
\newblock {Pfam}: The protein families database in 2021.
\newblock \emph{Nucleic Acids Research}, 49\penalty0 (D1):\penalty0 D412--D419,
  2021.
\newblock \doi{10.1093/nar/gkaa913}.

\bibitem[Schorn et~al.(2021)Schorn, Verhoeven, Ridder, Huber, Acharya, Aksenov,
  Aleti, Moghaddam, Aron, Aziz, Bauermeister, Bauman, Baunach, Beemelmanns,
  Beman, Berlanga-Clavero, Blacutt, Bode, Boullie, Brejnrod, Bugni, Calteau,
  Cao, Carri{\'o}n, Castelo-Branco, Chanana, Chase, Chevrette, Costa-Lotufo,
  Crawford, Currie, Cuypers, Dang, de~Rond, Demko, Dittmann, Du, Drozd,
  Dujardin, Dutton, Edlund, Fewer, Garg, Gauglitz, Gentry, Gerwick, Glukhov,
  Gross, Gugger, Guill{\'e}n~Matus, Helfrich, Hempel, Hur, Iorio, Jensen, Kang,
  Kaysser, Kelleher, Kim, Kim, Koester, K{\"o}nig, Leao, Lee, Lee, Li, Little,
  Maloney, M{\"a}nnle, {Martin H.}, McAvoy, Metcalf, Mohimani, Molina-Santiago,
  Moore, Mullowney, Muskat, Nothias, O'Neill, Parkinson, Petras, Piel, Pierce,
  Pires, Reher, Romero, Roper, Rust, Saad, Saenz, Sanchez, S{\o}rensen, Sosio,
  S{\"u}ssmuth, Sweeney, Tahlan, Thomson, Tobias, Trindade-Silva, van Wezel,
  Wang, Weldon, Zhang, Ziemert, Duncan, Cr{\"u}semann, Rogers, Dorrestein,
  Medema, and van~der Hooft]{Schorn2021PairedOmics}
Michelle~A. Schorn, Stefan Verhoeven, Lars Ridder, Florian Huber, Deepa~D.
  Acharya, Alexander~A. Aksenov, Gajender Aleti, Jamshid~Amiri Moghaddam,
  Allegra~T. Aron, Saefuddin Aziz, Anelize Bauermeister, Katherine~D. Bauman,
  Martin Baunach, Christine Beemelmanns, J.~Michael Beman, Mar{\'\i}a~Victoria
  Berlanga-Clavero, Alex~A. Blacutt, Helge~B. Bode, Anne Boullie, Asker
  Brejnrod, Tim~S. Bugni, Alexandra Calteau, Liu Cao, V{\'\i}ctor~J.
  Carri{\'o}n, Raquel Castelo-Branco, Shaurya Chanana, Alexander~B. Chase,
  Marc~G. Chevrette, Leticia~V. Costa-Lotufo, Jason~M. Crawford, Cameron~R.
  Currie, Bart Cuypers, Tam Dang, Tristan de~Rond, Alyssa~M. Demko, Elke
  Dittmann, Chao Du, Christopher Drozd, Jean-Claude Dujardin, Rachel~J. Dutton,
  Anna Edlund, David~P. Fewer, Neha Garg, Julia~M. Gauglitz, Emily~C. Gentry,
  Lena Gerwick, Evgenia Glukhov, Harald Gross, Muriel Gugger, Dulce~G.
  Guill{\'e}n~Matus, Eric J.~N. Helfrich, Benjamin-Florian Hempel, Jae-Seoun
  Hur, Marianna Iorio, Paul~R. Jensen, Kyo~Bin Kang, Leonard Kaysser, Neil~L.
  Kelleher, Chung~Sub Kim, Ki~Hyun Kim, Irina Koester, Gabriele~M. K{\"o}nig,
  Tiago Leao, Seoung~Rak Lee, Yi-Yuan Lee, Xuanji Li, Jessica~C. Little,
  Katherine~N. Maloney, Daniel M{\"a}nnle, Christian {Martin H.}, Andrew~C.
  McAvoy, Willam~W. Metcalf, Hosein Mohimani, Carlos Molina-Santiago,
  Bradley~S. Moore, Michael~W. Mullowney, Mitchell Muskat, Louis-F{\'e}lix
  Nothias, Ellis~C. O'Neill, Elizabeth~I. Parkinson, Daniel Petras, J{\"o}rn
  Piel, Emily~C. Pierce, Karine Pires, Raphael Reher, Diego Romero, M.~Caroline
  Roper, Michael Rust, Hamada Saad, Carmen Saenz, Laura~M. Sanchez,
  S{\o}ren~Johannes S{\o}rensen, Margherita Sosio, Roderich~D. S{\"u}ssmuth,
  Douglas Sweeney, Kapil Tahlan, Regan~J. Thomson, Nicholas~J. Tobias, Amaro~E.
  Trindade-Silva, Gilles~P. van Wezel, Mingxun Wang, Kelly~C. Weldon, Fan
  Zhang, Nadine Ziemert, Katherine~R. Duncan, Max Cr{\"u}semann, Simon Rogers,
  Pieter~C. Dorrestein, Marnix~H. Medema, and Justin J.~J. van~der Hooft.
\newblock A community resource for paired genomic and metabolomic data mining.
\newblock \emph{Nature Chemical Biology}, 17\penalty0 (4):\penalty0 363--368,
  2021.
\newblock \doi{10.1038/s41589-020-00724-z}.

\bibitem[van~der Hooft et~al.(2020)van~der Hooft, Mohimani, Bauermeister,
  Dorrestein, Duncan, and Medema]{VanDerHooft2020LinkingOmics}
Justin J.~J. van~der Hooft, Hosein Mohimani, Anelize Bauermeister, Pieter~C.
  Dorrestein, Katherine~R. Duncan, and Marnix~H. Medema.
\newblock Linking genomics and metabolomics to chart specialized metabolic
  diversity.
\newblock \emph{Chemical Society Reviews}, 49\penalty0 (11):\penalty0
  3297--3314, 2020.
\newblock \doi{10.1039/D0CS00162G}.

\bibitem[Le{\~a}o et~al.(2022)Le{\~a}o, Wang, da~Silva, Gurevich, Bauermeister,
  Gomes, Brejnrod, Glukhov, Aron, Louwen, Kim, Reher, Fiore, van~der Hooft,
  Gerwick, Gerwick, Bandeira, and Dorrestein]{Leao2022NPOmix}
Tiago~F. Le{\~a}o, Mingxun Wang, Ricardo da~Silva, Alexey Gurevich, Anelize
  Bauermeister, Paulo Wender~P. Gomes, Asker Brejnrod, Evgenia Glukhov,
  Allegra~T. Aron, Joris J.~R. Louwen, Hyun~Woo Kim, Raphael Reher, Marli~F.
  Fiore, Justin J.~J. van~der Hooft, Lena Gerwick, William~H. Gerwick, Nuno
  Bandeira, and Pieter~C. Dorrestein.
\newblock {NPOmix}: A machine learning classifier to connect mass spectrometry
  fragmentation data to biosynthetic gene clusters.
\newblock \emph{PNAS Nexus}, 1\penalty0 (5):\penalty0 pgac257, 2022.
\newblock \doi{10.1093/pnasnexus/pgac257}.

\bibitem[Mao et~al.(2018)Mao, Okada, Wu, Xu, and
  Seyedsayamdost]{Mao2018SilentBGCs}
Dainan Mao, Bethany~K. Okada, Yihan Wu, Fei Xu, and Mohammad~R. Seyedsayamdost.
\newblock Recent advances in activating silent biosynthetic gene clusters in
  bacteria.
\newblock \emph{Current Opinion in Microbiology}, 45:\penalty0 156--163, 2018.
\newblock \doi{10.1016/j.mib.2018.05.001}.

\bibitem[Liu et~al.(2021)Liu, Zhao, Huang, and
  Luo]{Liu2021StreptomycesSilentBGCs}
Zhenyu Liu, Yatong Zhao, Chaoqun Huang, and Yunzi Luo.
\newblock Recent advances in silent gene cluster activation in
  \textit{Streptomyces}.
\newblock \emph{Frontiers in Bioengineering and Biotechnology}, 9:\penalty0
  632230, 2021.
\newblock \doi{10.3389/fbioe.2021.632230}.

\bibitem[Zdouc et~al.(2025)Zdouc, Blin, Louwen, Navarro, Loureiro, Bader,
  Bailey, Barra, Booth, Bozh{\"u}y{\"u}k, Cediel-Becerra, Charlop-Powers,
  Chevrette, Chooi, D'Agostino, de~Rond, Del~Pup, Duncan, Gu, Hanif, Helfrich,
  Jenner, Katsuyama, Korenskaia, Krug, Libis, Lund, Mantri, Morgan, Owen, Phan,
  Philmus, Reitz, Robinson, Singh, Teufel, Tong, Tugizimana, Ulanova, Winter,
  Aguilar, Akiyama, Al-Salihi, Alanjary, Alberti, Aleti, Alharthi, Rojo,
  Arishi, Augustijn, Avalon, Avelar-Rivas, Axt, Barbieri, Barbosa,
  Barboza~Segato, Barrett, Baunach, Beemelmanns, Beqaj, Berger,
  Bernaldo-Ag{\"u}ero, Bettenb{\"u}hl, Bielinski, Biermann, Borges, Borriss,
  Breitenbach, Bretscher, Brigham, Buedenbender, Bulcock, Cano-Prieto, Capela,
  Carrion, Carter, Castelo-Branco, Castro-Falc{\'o}n, Chagas,
  Charria-Gir{\'o}n, Chaudhri, Chaudhry, Choi, Choi, Choupannejad, Chromy,
  Donahey, Collemare, Connolly, Creamer, Cr{\"u}semann, Cruz, Cumsille,
  Dallery, Damas-Ramos, Damiani, de~Kruijff, Mart{\'i}n, Sala, Dillen, Doering,
  Dommaraju, Durusu, Egbert, Ellerhorst, Faussurier, Fetter, Feuermann, Fewer,
  Foldi, Frediansyah, Garza, Gavriilidou, Gentile, Gerke, Gerstmans,
  Gomez-Escribano, Gonz{\'a}lez-Salazar, Grayson, Greco, Gomez, Guerra, Flores,
  Gurevich, Guti{\'e}rrez-Garc{\'i}a, Hart, Haslinger, He, Hebra, Hemmann,
  Hindra, H{\"o}ing, Holland, Holme, Horch, Hrab, Hu, Huynh, Hwang, Iacovelli,
  Iftime, Iorio, Jayachandran, Jeong, Jing, Jung, Kakumu, Kalkreuter, Kang,
  Kang, Kim, Kim, Kim, Kim, Klapper, Koetsier, Kollten, Kov{\'a}cs, Kriukova,
  Kubach, Kunjapur, Kushnareva, Kust, Lamber, Larralde, Larsen, Launay, Le,
  Lebeer, Lee, Lee, Lev, Li, Li, Licona-Cassani, Lien, Liu, Lopez,
  Machushynets, Macias, Mahmud, Maleckis, Martinez-Martinez, Mast, Maximo,
  McBride, McLellan, Bhatt, Melkonian, Merrild, Mets{\"a}-Ketel{\"a}, Mitchell,
  M{\"u}ller, Nguyen, Nguyen, Niedermeyer, O'Hare, Ossowicki, Ostash, Otani,
  Padva, Paliyal, Pan, Panghal, Parade, Park, Parra, Rubio, Pham, Pidot, Piel,
  Pourmohsenin, Rakhmanov, Ramesh, Rasmussen, Rego, Reher, Rice, Rigolet,
  Romero-Otero, Rosas-Becerra, Rosiles, Rutz, Ryu, Sahadeo, Saldanha, Salvi,
  S{\'a}nchez-Carvajal, Santos-Medellin, Sbaraini, Schoellhorn, Schumm, Sehnal,
  Selem, Shah, Shishido, Sieber, Silviani, Singh, Singh, Sokolova,
  Sonnenschein, Sosio, Sowa, Steffen, Stegmann, Streiff, Str{\"u}der, Surup,
  Svenningsen, Sweeney, Szenei, Tagirdzhanov, Tan, Tarnowski, Terlouw, Rey,
  Thome, Torres~Ortega, T{\o}rring, Trindade, Truman, Tvilum, Udwary, Ulbricht,
  Vader, van Wezel, Walmsley, Warnasinghe, Weddeling, Weir, Williams, Williams,
  Witte, Rocca, Yamada, Yang, Yang, Yu, Zhou, Ziemert, Zimmer, Zimmermann,
  Zimmermann, van~der Hooft, Linington, Weber, and Medema]{mibig}
Mitja~M. Zdouc, Kai Blin, Nico L.~L. Louwen, Jorge Navarro, Catarina Loureiro,
  Chantal~D. Bader, Constance~B. Bailey, Lena Barra, Thomas~J. Booth, Kenan
  A.~J. Bozh{\"u}y{\"u}k, Jos{\'e} D.~D. Cediel-Becerra, Zachary
  Charlop-Powers, Marc~G. Chevrette, Yit~Heng Chooi, Paul~M. D'Agostino,
  Tristan de~Rond, Elena Del~Pup, Katherine~R. Duncan, Wenjia Gu, Novriyandi
  Hanif, Eric J.~N. Helfrich, Matthew Jenner, Yohei Katsuyama, Aleksandra
  Korenskaia, Daniel Krug, Vincent Libis, George~A. Lund, Shrikant Mantri,
  Kalindi~D. Morgan, Charlotte Owen, Chin-Soon Phan, Benjamin Philmus,
  Zachary~L. Reitz, Serina~L. Robinson, Kumar~Saurabh Singh, Robin Teufel,
  Yaojun Tong, Fidele Tugizimana, Dana Ulanova, Jaclyn~M. Winter, C{\'e}sar
  Aguilar, Daniel~Y. Akiyama, Suhad A.~A. Al-Salihi, Mohammad Alanjary,
  Fabrizio Alberti, Gajender Aleti, Shumukh~A. Alharthi, Mariela Y.~Arias Rojo,
  Amr~A. Arishi, Hannah~E. Augustijn, Nicole~E. Avalon, J.~Abraham
  Avelar-Rivas, Kyle~K. Axt, Hellen~B. Barbieri, Julio Cesar~J. Barbosa,
  Lucas~Gabriel Barboza~Segato, Susanna~E. Barrett, Martin Baunach, Christine
  Beemelmanns, Dardan Beqaj, Tim Berger, Jordan Bernaldo-Ag{\"u}ero, Sandra~M.
  Bettenb{\"u}hl, Vincent~A. Bielinski, Friederike Biermann, Ricardo~M. Borges,
  Rainer Borriss, Milena Breitenbach, Kevin~M. Bretscher, Michael~W. Brigham,
  Larissa Buedenbender, Brodie~W. Bulcock, Carolina Cano-Prieto, Jo{\~a}o
  Capela, Victor~J. Carrion, Riley~S. Carter, Raquel Castelo-Branco, Gabriel
  Castro-Falc{\'o}n, Fernanda~O. Chagas, Esteban Charria-Gir{\'o}n,
  Ayesha~Ahmed Chaudhri, Vasvi Chaudhry, Hyukjae Choi, Yukyung Choi, Roya
  Choupannejad, Jakub Chromy, Melinda S.~Chue Donahey, J{\'e}r{\^o}me
  Collemare, Jack~A. Connolly, Kaitlin~E. Creamer, Max Cr{\"u}semann,
  Andres~Arredondo Cruz, Andres Cumsille, Jean-Felix Dallery, Luis~Caleb
  Damas-Ramos, Tito Damiani, Martinus de~Kruijff, Bel{\'e}n~Delgado Mart{\'i}n,
  Gerardo~Della Sala, Jelle Dillen, Drew~T. Doering, Shravan~R. Dommaraju,
  Suhan Durusu, Susan Egbert, Mark Ellerhorst, Baptiste Faussurier, Artem
  Fetter, Marc Feuermann, David~P. Fewer, Jonathan Foldi, Andri Frediansyah,
  Erin~A. Garza, Athina Gavriilidou, Andrea Gentile, Jennifer Gerke, Hans
  Gerstmans, Juan~Pablo Gomez-Escribano, Luz~A. Gonz{\'a}lez-Salazar,
  Natalie~E. Grayson, Claudio Greco, Juan E.~Gris Gomez, Sebastian Guerra,
  Shaday~Guerrero Flores, Alexey Gurevich, Karina Guti{\'e}rrez-Garc{\'i}a,
  Lauren Hart, Kristina Haslinger, Beibei He, Teo Hebra, Jethro~L. Hemmann,
  Hindra Hindra, Lars H{\"o}ing, Darren~C. Holland, Jonathan~E. Holme, Therese
  Horch, Pavlo Hrab, Jie Hu, Thanh-Hau Huynh, Ji-Yeon Hwang, Riccardo
  Iacovelli, Dumitrita Iftime, Marianna Iorio, Sidharth Jayachandran, Eunah
  Jeong, Jiayi Jing, Jung~J. Jung, Yuya Kakumu, Edward Kalkreuter, Kyo~Bin
  Kang, Sangwook Kang, Wonyong Kim, Geum~Jin Kim, Hyunwoo Kim, Hyun~Uk Kim,
  Martin Klapper, Robert~A. Koetsier, Cassandra Kollten, {\'A}kos~T.
  Kov{\'a}cs, Yelyzaveta Kriukova, Noel Kubach, Aditya~M. Kunjapur,
  Aleksandra~K. Kushnareva, Andreja Kust, Jessica Lamber, Martin Larralde,
  Niels~J. Larsen, Adrien~P. Launay, Ngoc-Thao-Hien Le, Sarah Lebeer, Byung~Tae
  Lee, Kyungha Lee, Katherine~L. Lev, Shu-Ming Li, Yong-Xin Li, Cuauht{\'e}moc
  Licona-Cassani, Annette Lien, Jing Liu, Julius Adam~V. Lopez, Nataliia~V.
  Machushynets, Marla~I. Macias, Taifo Mahmud, Matiss Maleckis,
  A{\~n}adir~Maharai Martinez-Martinez, Yvonne Mast, Marina~F. Maximo,
  Christina~M. McBride, Rose~M. McLellan, Khyati~Mehta Bhatt, Chrats Melkonian,
  Aske Merrild, Mikko Mets{\"a}-Ketel{\"a}, Douglas~A. Mitchell, Alison~V.
  M{\"u}ller, Giang-Son Nguyen, Hera~T. Nguyen, Timo H.~J. Niedermeyer,
  Julia~H. O'Hare, Adam Ossowicki, Bohdan~O. Ostash, Hiroshi Otani, Leo Padva,
  Sunaina Paliyal, Xinya Pan, Mohit Panghal, Dana~S. Parade, Jiyoon Park,
  Jonathan Parra, Marcos~Pedraza Rubio, Huong~T. Pham, Sacha~J. Pidot, J{\"o}rn
  Piel, Bita Pourmohsenin, Malik Rakhmanov, Sangeetha Ramesh, Michelle~H.
  Rasmussen, Adriana Rego, Raphael Reher, Andrew~J. Rice, Augustin Rigolet,
  Adriana Romero-Otero, Luis~Rodrigo Rosas-Becerra, Pablo~Y. Rosiles, Adriano
  Rutz, Byeol Ryu, Libby-Ann Sahadeo, Murrel Saldanha, Luca Salvi, Eduardo
  S{\'a}nchez-Carvajal, Christian Santos-Medellin, Nicolau Sbaraini, Sydney~M.
  Schoellhorn, Clemens Schumm, Ludek Sehnal, Nelly Selem, Anjali~D. Shah,
  Tania~K. Shishido, Simon Sieber, Velina Silviani, Garima Singh, Hemant Singh,
  Nika Sokolova, Eva~C. Sonnenschein, Margherita Sosio, Sven~T. Sowa, Karin
  Steffen, Evi Stegmann, Alena~B. Streiff, Alena Str{\"u}der, Frank Surup,
  Tiziana Svenningsen, Douglas Sweeney, Judit Szenei, Azat Tagirdzhanov, Bin
  Tan, Matthew~J. Tarnowski, Barbara~R. Terlouw, Thomas Rey, Nicola~U. Thome,
  Laura~Rosina Torres~Ortega, Thomas T{\o}rring, Marla Trindade, Andrew~W.
  Truman, Marie Tvilum, Daniel~W. Udwary, Christoph Ulbricht, Lisa Vader,
  Gilles~P. van Wezel, Max Walmsley, Randika Warnasinghe, Heiner~G. Weddeling,
  Angus N.~M. Weir, Katherine Williams, Sam~E. Williams, Thomas~E. Witte,
  Steffaney M.~Wood Rocca, Keith Yamada, Dong Yang, Dongsoo Yang, Jingwei Yu,
  Zhenyi Zhou, Nadine Ziemert, Lukas Zimmer, Alina Zimmermann, Christian
  Zimmermann, Justin J.~J. van~der Hooft, Roger~G. Linington, Tilmann Weber,
  and Marnix~H. Medema.
\newblock {MIBiG} 4.0: Advancing biosynthetic gene cluster curation through
  global collaboration.
\newblock \emph{Nucleic Acids Research}, 53\penalty0 (D1):\penalty0 D678--D690,
  2025.
\newblock \doi{10.1093/nar/gkae1115}.

\bibitem[Lai et~al.(2025)Lai, Yao, Zha, Zhang, Zhang, Ye, Zhang, Bai, and
  Ning]{bgcprophet}
Qilong Lai, Shuai Yao, Yuguo Zha, Haohong Zhang, Haobo Zhang, Ying Ye, Yonghui
  Zhang, Hong Bai, and Kang Ning.
\newblock Deciphering the biosynthetic potential of microbial genomes using a
  {BGC} language processing neural network model.
\newblock \emph{Nucleic Acids Research}, 53\penalty0 (7):\penalty0 gkaf305,
  2025.
\newblock \doi{10.1093/nar/gkaf305}.

\bibitem[Malygina and Kalinina(2024)]{Malygina2024NaturalPPLuM}
Tatiana Malygina and Olga Kalinina.
\newblock Exploring sequence landscape of biosynthetic gene clusters with
  protein language models.
\newblock In \emph{{ICML}'24 Workshop {ML} for Life and Material Science: From
  Theory to Industry Applications ({ML4LMS})}, 2024.
\newblock URL \url{https://openreview.net/forum?id=UguakxTKG5}.

\bibitem[Rios-Martinez et~al.(2023)Rios-Martinez, Bhattacharya, Amini,
  Crawford, and Yang]{bigcarp}
Carolina Rios-Martinez, Nicholas Bhattacharya, Ava~P. Amini, Lorin Crawford,
  and Kevin~K. Yang.
\newblock Deep self-supervised learning for biosynthetic gene cluster detection
  and product classification.
\newblock \emph{PLOS Computational Biology}, 19\penalty0 (5):\penalty0
  e1011162, 2023.
\newblock \doi{10.1371/journal.pcbi.1011162}.

\bibitem[Kawano et~al.(2026)Kawano, Shiraishi, Kuzuyama, and
  Umemura]{kawano2026roberta}
Tomoki Kawano, Taro Shiraishi, Tomohisa Kuzuyama, and Maiko Umemura.
\newblock A novel transformer-based platform for the prediction and design of
  biosynthetic gene clusters for (un)natural products.
\newblock \emph{PLOS Computational Biology}, 22\penalty0 (2):\penalty0
  e1013181, 2026.
\newblock \doi{10.1371/journal.pcbi.1013181}.

\bibitem[Hannigan et~al.(2019)Hannigan, Prihoda, Palicka, Soukup, Klempir,
  Rampula, Durcak, Wurst, Kotowski, Chang, Wang, Piizzi, Temesi, Hazuda, Woelk,
  and Bitton]{deepbgc}
Geoffrey~D. Hannigan, David Prihoda, Andrej Palicka, Jindrich Soukup, Ondrej
  Klempir, Lena Rampula, Jindrich Durcak, Michael Wurst, Jakub Kotowski, Dan
  Chang, Rurun Wang, Grazia Piizzi, Gergely Temesi, Daria~J. Hazuda,
  Christopher~H. Woelk, and Danny~A. Bitton.
\newblock A deep learning genome-mining strategy for biosynthetic gene cluster
  prediction.
\newblock \emph{Nucleic Acids Research}, 47\penalty0 (18):\penalty0 e110, 2019.
\newblock \doi{10.1093/nar/gkz654}.

\bibitem[Kim et~al.(2021)Kim, Wang, Leber, Nothias, Reher, Kang, {van der
  Hooft}, Dorrestein, Gerwick, and Cottrell]{npclassifier}
Hyun~Woo Kim, Mingxun Wang, Christopher~A. Leber, Louis-F{\'e}lix Nothias,
  Raphael Reher, Kyo~Bin Kang, Justin J.~J. {van der Hooft}, Pieter~C.
  Dorrestein, William~H. Gerwick, and Garrison~W. Cottrell.
\newblock {NPClassifier}: A deep neural network-based structural classification
  tool for natural products.
\newblock \emph{Journal of Natural Products}, 84\penalty0 (11):\penalty0
  2795--2807, 2021.
\newblock \doi{10.1021/acs.jnatprod.1c00399}.

\bibitem[Porokhin et~al.(2026)Porokhin, Mevers, van~der Hooft, and
  Hassoun]{porokhin2025bgcat}
Vladimir Porokhin, Emily Mevers, Justin J.~J. van~der Hooft, and Soha Hassoun.
\newblock Fine-grained structural classification of biosynthetic gene
  cluster-encoded products.
\newblock \emph{Bioinformatics}, 42\penalty0 (8):\penalty0 btag559, 2026.
\newblock \doi{10.1093/bioinformatics/btag559}.

\bibitem[Xiao et~al.(2025)Xiao, Breitling, Takano, and Alvarez]{xiao2025cross}
Hanlin Xiao, Rainer Breitling, Eriko Takano, and Mauricio~A. Alvarez.
\newblock Cross-granularity representations for biological sequences: Insights
  from {ESM} and {BiGCARP}.
\newblock In \emph{2025 {IEEE} International Conference on Bioinformatics and
  Biomedicine ({BIBM})}, pages 6936--6943. IEEE, 2025.
\newblock \doi{10.1109/bibm66473.2025.11356265}.

\bibitem[Hie et~al.(2020)Hie, Bryson, and Berger]{Hie2020Uncertainty}
Brian Hie, Bryan~D. Bryson, and Bonnie Berger.
\newblock Leveraging uncertainty in machine learning accelerates biological
  discovery and design.
\newblock \emph{Cell Systems}, 11\penalty0 (5):\penalty0 461--477.e9, 2020.
\newblock \doi{10.1016/j.cels.2020.09.007}.

\bibitem[Biswas et~al.(2021)Biswas, Khimulya, Alley, Esvelt, and
  Church]{Biswas2021LowN}
Surojit Biswas, Grigory Khimulya, Ethan~C. Alley, Kevin~M. Esvelt, and
  George~M. Church.
\newblock Low-{N} protein engineering with data-efficient deep learning.
\newblock \emph{Nature Methods}, 18\penalty0 (4):\penalty0 389--396, 2021.
\newblock \doi{10.1038/s41592-021-01100-y}.

\bibitem[Jiang et~al.(2025)Jiang, Yan, Di~Bernardo, Sgrizzi, Villiger,
  Kayabolen, Kim, Carscadden, Hiraizumi, Nishimasu, Gootenberg, and
  Abudayyeh]{Jiang2024EVOLVEpro}
Kaiyi Jiang, Zhaoqing Yan, Matteo Di~Bernardo, Samantha~R. Sgrizzi, Lukas
  Villiger, Alisan Kayabolen, B.~J. Kim, Josephine~K. Carscadden, Masahiro
  Hiraizumi, Hiroshi Nishimasu, Jonathan~S. Gootenberg, and Omar~O. Abudayyeh.
\newblock Rapid in silico directed evolution by a protein language model with
  {EVOLVEpro}.
\newblock \emph{Science}, 387\penalty0 (6732):\penalty0 eadr6006, 2025.
\newblock \doi{10.1126/science.adr6006}.

\bibitem[Cui et~al.(2025)Cui, Tejada-Lapuerta, Brbi{\'c}, Saez-Rodriguez,
  Cristea, Goodarzi, Lotfollahi, Theis, and
  Wang]{Cui2025MultimodalFoundationModels}
Haotian Cui, Alejandro Tejada-Lapuerta, Maria Brbi{\'c}, Julio Saez-Rodriguez,
  Simona Cristea, Hani Goodarzi, Mohammad Lotfollahi, Fabian~J. Theis, and
  Bo~Wang.
\newblock Towards multimodal foundation models in molecular cell biology.
\newblock \emph{Nature}, 640\penalty0 (8059):\penalty0 623--633, 2025.
\newblock \doi{10.1038/s41586-025-08710-y}.

\bibitem[Cimermancic et~al.(2014)Cimermancic, Medema, Claesen, Kurita,
  Wieland~Brown, Mavrommatis, Pati, Godfrey, Koehrsen, Clardy, Birren, Takano,
  Sali, Linington, and Fischbach]{cimermancic2014clusterfinder}
Peter Cimermancic, Marnix~H. Medema, Jan Claesen, Kenji Kurita, Laura~C.
  Wieland~Brown, Konstantinos Mavrommatis, Amrita Pati, Paul~A. Godfrey,
  Michael Koehrsen, Jon Clardy, Bruce~W. Birren, Eriko Takano, Andrej Sali,
  Roger~G. Linington, and Michael~A. Fischbach.
\newblock Insights into secondary metabolism from a global analysis of
  prokaryotic biosynthetic gene clusters.
\newblock \emph{Cell}, 158\penalty0 (2):\penalty0 412--421, 2014.
\newblock \doi{10.1016/j.cell.2014.06.034}.

\bibitem[Blin et~al.(2025)Blin, Shaw, Vader, Szenei, Reitz, Augustijn,
  Cediel-Becerra, de~Cr{\'e}cy-Lagard, Koetsier, Williams, Cruz-Morales,
  Wongwas, Segurado~Luchsinger, Biermann, Korenskaia, Zdouc, Meijer, Terlouw,
  van~der Hooft, Ziemert, Helfrich, Masschelein, Corre, Chevrette, van Wezel,
  Medema, and Weber]{antismash8}
Kai Blin, Simon Shaw, Lisa Vader, Judit Szenei, Zachary~L. Reitz, Hannah~E.
  Augustijn, Jos{\'e} D.~D. Cediel-Becerra, Val{\'e}rie de~Cr{\'e}cy-Lagard,
  Robert~A. Koetsier, Sam~E. Williams, Pablo Cruz-Morales, Sopida Wongwas,
  Alejandro~E. Segurado~Luchsinger, Friederike Biermann, Aleksandra Korenskaia,
  Mitja~M. Zdouc, David Meijer, Barbara~R. Terlouw, Justin J.~J. van~der Hooft,
  Nadine Ziemert, Eric J.~N. Helfrich, Joleen Masschelein, Christophe Corre,
  Marc~G. Chevrette, Gilles~P. van Wezel, Marnix~H. Medema, and Tilmann Weber.
\newblock {antiSMASH 8.0}: Extended gene cluster detection capabilities and
  analyses of chemistry, enzymology, and regulation.
\newblock \emph{Nucleic Acids Research}, 53\penalty0 (W1):\penalty0 W32--W38,
  2025.
\newblock \doi{10.1093/nar/gkaf334}.

\bibitem[Liu et~al.(2026)Liu, Li, Ong, Wong, Tay, Lim, Foo, and
  Koh]{liu2025unified}
Guimei Liu, Yiting Li, Gabriel Ong, Fong~Tian Wong, Dillon W.~P. Tay, Yee~Hwee
  Lim, Chuan~Sheng Foo, and Winston Koh.
\newblock Unified genomic and chemical representations enable bidirectional
  biosynthetic gene cluster and natural product retrieval.
\newblock \emph{Scientific Reports}, 16:\penalty0 21291, 2026.
\newblock \doi{10.1038/s41598-026-49955-5}.

\bibitem[Li et~al.(2025{\natexlab{a}})Li, Takano, Zhang, Draper, Yeo, and
  Breitling]{proscope}
Yiting Li, Eriko Takano, Jing Zhang, Ben Draper, Wayne Wei~Zhong Yeo, and
  Rainer Breitling.
\newblock {ProSCOPE}: Modeling biosynthetic diversity via probabilistic
  structure-aware co-embedding.
\newblock In \emph{2025 {IEEE} International Conference on Bioinformatics and
  Biomedicine ({BIBM})}, pages 4905--4913. IEEE, 2025{\natexlab{a}}.
\newblock \doi{10.1109/bibm66473.2025.11356651}.

\bibitem[Louwen et~al.(2023)Louwen, Medema, and van~der
  Hooft]{Louwen2023NPLinker}
Joris J.~R. Louwen, Marnix~H. Medema, and Justin J.~J. van~der Hooft.
\newblock Enhanced correlation-based linking of biosynthetic gene clusters to
  their metabolic products through chemical class matching.
\newblock \emph{Microbiome}, 11:\penalty0 13, 2023.
\newblock \doi{10.1186/s40168-022-01444-3}.

\bibitem[Rao et~al.(2019)Rao, Bhattacharya, Thomas, Duan, Chen, Canny, Abbeel,
  and Song]{rao2019tape}
Roshan Rao, Nicholas Bhattacharya, Neil Thomas, Yan Duan, Xi~Chen, John Canny,
  Pieter Abbeel, and Yun~S. Song.
\newblock Evaluating protein transfer learning with {TAPE}.
\newblock In \emph{Advances in Neural Information Processing Systems},
  volume~32, pages 9689--9701, 2019.
\newblock URL
  \url{https://proceedings.neurips.cc/paper_files/paper/2019/hash/37f65c068b7723cd7809ee2d31d7861c-Abstract.html}.

\bibitem[Xu et~al.(2022)Xu, Zhang, Lu, Zhu, Zhang, Ma, Liu, and
  Tang]{xu2022peer}
Minghao Xu, Zuobai Zhang, Jiarui Lu, Zhaocheng Zhu, Yangtian Zhang, Chang Ma,
  Runcheng Liu, and Jian Tang.
\newblock {PEER}: A comprehensive and multi-task benchmark for protein sequence
  understanding.
\newblock In \emph{Advances in Neural Information Processing Systems},
  volume~35, pages 35156--35173, 2022.
\newblock \doi{10.52202/068431-2548}.
\newblock URL
  \url{https://proceedings.neurips.cc/paper_files/paper/2022/hash/e467582d42d9c13fa9603df16f31de6d-Abstract-Datasets_and_Benchmarks.html}.

\bibitem[Dallago et~al.(2021)Dallago, Mou, Johnston, Wittmann, Bhattacharya,
  Goldman, Madani, and Yang]{dallago2021flip}
Christian Dallago, Jody Mou, Kadina~E. Johnston, Bruce~J. Wittmann, Nicholas
  Bhattacharya, Samuel Goldman, Ali Madani, and Kevin~K. Yang.
\newblock {FLIP}: Benchmark tasks in fitness landscape inference for proteins.
\newblock In \emph{Proceedings of the Neural Information Processing Systems
  Track on Datasets and Benchmarks}, volume~1, 2021.
\newblock URL
  \url{https://datasets-benchmarks-proceedings.neurips.cc/paper_files/paper/2021/hash/2b44928ae11fb9384c4cf38708677c48-Abstract-round2.html}.

\bibitem[Notin et~al.(2023)Notin, Kollasch, Ritter, van Niekerk, Paul, Spinner,
  Rollins, Shaw, Orenbuch, Weitzman, Frazer, Dias, Franceschi, Gal, and
  Marks]{notin2023proteingym}
Pascal Notin, Aaron~W. Kollasch, Daniel Ritter, Lood van Niekerk, Steffanie
  Paul, Han Spinner, Nathan Rollins, Ada Shaw, Rose Orenbuch, Ruben Weitzman,
  Jonathan Frazer, Mafalda Dias, Dinko Franceschi, Yarin Gal, and Debora~S.
  Marks.
\newblock {ProteinGym}: Large-scale benchmarks for protein fitness prediction
  and design.
\newblock In \emph{Advances in Neural Information Processing Systems},
  volume~36, pages 64331--64379, 2023.
\newblock \doi{10.52202/075280-2810}.
\newblock URL
  \url{https://proceedings.neurips.cc/paper_files/paper/2023/hash/cac723e5ff29f65e3fcbb0739ae91bee-Abstract-Datasets_and_Benchmarks.html}.

\bibitem[Wu et~al.(2018)Wu, Ramsundar, Feinberg, Gomes, Geniesse, Pappu,
  Leswing, and Pande]{wu2018moleculenet}
Zhenqin Wu, Bharath Ramsundar, Evan~N. Feinberg, Joseph Gomes, Caleb Geniesse,
  Aneesh~S. Pappu, Karl Leswing, and Vijay Pande.
\newblock {MoleculeNet}: A benchmark for molecular machine learning.
\newblock \emph{Chemical Science}, 9\penalty0 (2):\penalty0 513--530, 2018.
\newblock \doi{10.1039/C7SC02664A}.

\bibitem[Steshin(2023)]{steshin2023lohi}
Simon Steshin.
\newblock {Lo-Hi}: Practical {ML} drug discovery benchmark.
\newblock In \emph{Advances in Neural Information Processing Systems},
  volume~36, pages 64526--64554, 2023.
\newblock \doi{10.52202/075280-2816}.
\newblock URL
  \url{https://proceedings.neurips.cc/paper_files/paper/2023/hash/cb82f1f97ad0ca1d92df852a44a3bd73-Abstract-Datasets_and_Benchmarks.html}.

\bibitem[Bushuiev et~al.(2024)Bushuiev, Bushuiev, de~Jonge, Young, Kretschmer,
  Samusevich, Heirman, Wang, Zhang, D{\"u}hrkop, Ludwig, Haupt, Kalia, Brungs,
  Schmid, Greiner, Wang, Wishart, Liu, Rousu, Bittremieux, Rost, Mak, Hassoun,
  Huber, van~der Hooft, Stravs, B{\"o}cker, Sivic, and
  Pluskal]{bushuiev2024massspecgym}
Roman Bushuiev, Anton Bushuiev, Niek~F. de~Jonge, Adamo Young, Fleming
  Kretschmer, Raman Samusevich, Janne Heirman, Fei Wang, Luke Zhang, Kai
  D{\"u}hrkop, Marcus Ludwig, Nils~A. Haupt, Apurva Kalia, Corinna Brungs,
  Robin Schmid, Russell Greiner, Bo~Wang, David~S. Wishart, Li-Ping Liu, Juho
  Rousu, Wout Bittremieux, Hannes Rost, Tytus~D. Mak, Soha Hassoun, Florian
  Huber, Justin J.~J. van~der Hooft, Michael~A. Stravs, Sebastian B{\"o}cker,
  Josef Sivic, and Tom{\'a}{\v{s}} Pluskal.
\newblock {MassSpecGym}: A benchmark for the discovery and identification of
  molecules.
\newblock In \emph{Advances in Neural Information Processing Systems},
  volume~37, pages 110010--110027, 2024.
\newblock \doi{10.52202/079017-3491}.
\newblock URL
  \url{https://proceedings.neurips.cc/paper_files/paper/2024/hash/c6c31413d5c53b7d1c343c1498734b0f-Abstract-Datasets_and_Benchmarks_Track.html}.

\bibitem[Feng et~al.(2025)Feng, Wu, Zhao, Huff, Zhang, Wu, Lin, Wei, and
  Wu]{feng2025dnafm}
Haonan Feng, Lang Wu, Bingxin Zhao, Chad Huff, Jianjun Zhang, Jia Wu, Lifeng
  Lin, Peng Wei, and Chong Wu.
\newblock Benchmarking {DNA} foundation models for genomic and genetic tasks.
\newblock \emph{Nature Communications}, 16:\penalty0 10780, 2025.
\newblock \doi{10.1038/s41467-025-65823-8}.

\bibitem[Blin et~al.(2019)Blin, Pascal~Andreu, de~los Santos, Del~Carratore,
  Lee, Medema, and Weber]{blin2019antismashdb2}
Kai Blin, Vict{\`o}ria Pascal~Andreu, Emmanuel L.~C. de~los Santos, Francesco
  Del~Carratore, Sang~Yup Lee, Marnix~H. Medema, and Tilmann Weber.
\newblock The {antiSMASH} database version 2: A comprehensive resource on
  secondary metabolite biosynthetic gene clusters.
\newblock \emph{Nucleic Acids Research}, 47\penalty0 (D1):\penalty0 D625--D630,
  2019.
\newblock \doi{10.1093/nar/gky1060}.

\bibitem[Lu et~al.(2026)Lu, Li, Geng, Xu, Chen, Fu, Luesch, Ding, and
  Xie]{lu2026bgcmac}
Kechen Lu, Mengting Li, Hao Geng, Wenqiang Xu, Manyun Chen, Tian-Min Fu,
  Hendrik Luesch, Yousong Ding, and Wen~Jun Xie.
\newblock {BGC-MAC} and {BGC-MAP}: Attention-based models for biosynthetic gene
  cluster classification and product matching.
\newblock \emph{Journal of Chemical Information and Modeling}, 66\penalty0
  (1):\penalty0 138--151, 2026.
\newblock \doi{10.1021/acs.jcim.5c02082}.

\bibitem[Lin et~al.(2023)Lin, Akin, Rao, Hie, Zhu, Lu, Smetanin, Verkuil,
  Kabeli, Shmueli, dos Santos~Costa, Fazel-Zarandi, Sercu, Candido, and
  Rives]{lin2023esm2}
Zeming Lin, Halil Akin, Roshan Rao, Brian Hie, Zhongkai Zhu, Wenting Lu, Nikita
  Smetanin, Robert Verkuil, Ori Kabeli, Yaniv Shmueli, Allan dos Santos~Costa,
  Maryam Fazel-Zarandi, Tom Sercu, Salvatore Candido, and Alexander Rives.
\newblock Evolutionary-scale prediction of atomic-level protein structure with
  a language model.
\newblock \emph{Science}, 379\penalty0 (6637):\penalty0 1123--1130, 2023.
\newblock \doi{10.1126/science.ade2574}.

\bibitem[Candido et~al.(2026)Candido, Hayes, Derry, Rao, Lin, Verkuil, Wu, Lee,
  Bruguera, Keval, Kopylov, Pak, Wu, Thomas, Mataraso, Hsu, Trotman-Grant,
  Fatras, dos Santos~Costa, Badkundri, Ak{\i}n, Oktay, Deaton, Montabana,
  Sitwala, Yu, Wiggert, Carlin, Goering, Blazejewski, Sandora, Hla, Jia,
  Kloker, Sofroniew, Uehara, Pannu, Bachas, Liu, Sercu, and
  Rives]{candido2026esmc}
Salvatore Candido, Thomas Hayes, Alexander Derry, Roshan Rao, Zeming Lin,
  Robert Verkuil, Bryan~Z. Wu, Jin~Sub Lee, Elise~S. Bruguera, Jehan~A. Keval,
  Mykhailo Kopylov, John~E. Pak, Wesley Wu, Neil Thomas, Samson Mataraso, Alvin
  Hsu, Ashton~C. Trotman-Grant, Kilian Fatras, Allan dos Santos~Costa, Rohil
  Badkundri, Halil Ak{\i}n, Deniz Oktay, Jonathan Deaton, Elizabeth Montabana,
  Hrishita Sitwala, Yue Yu, Marius Wiggert, Dylan~Alexander Carlin, Anthony~W.
  Goering, Tomasz Blazejewski, McCullen Sandora, Michael Hla, Tina~Z. Jia,
  Leon~H. Kloker, Nicholas~J. Sofroniew, Masatoshi Uehara, Jassi Pannu, Sharrol
  Bachas, Daniel~S. Liu, Tom Sercu, and Alexander Rives.
\newblock Language modeling materializes a world model of protein biology.
\newblock bioRxiv preprint, 2026.
\newblock URL
  \url{https://www.biorxiv.org/content/10.64898/2026.06.03.729735v1}.

\bibitem[Wang et~al.(2025)Wang, Wang, Sadeh, Zancato, Achille, Karypis, and
  Rangwala]{wang2025lcplm}
Yingheng Wang, Zichen Wang, Gil Sadeh, Luca Zancato, Alessandro Achille, George
  Karypis, and Huzefa Rangwala.
\newblock {LC-PLM}: Long-context protein language modeling using bidirectional
  {Mamba} with shared projection layers.
\newblock \emph{Transactions on Machine Learning Research}, 2025.
\newblock URL \url{https://openreview.net/forum?id=dWvztQzfy4}.

\bibitem[Navarro-Mu{\~n}oz et~al.(2020)Navarro-Mu{\~n}oz, Selem-Mojica,
  Mullowney, Kautsar, Tryon, Parkinson, {De Los Santos}, Yeong, Cruz-Morales,
  Abubucker, Roeters, Lokhorst, Fernandez-Guerra, Cappelini, Goering, Thomson,
  Metcalf, Kelleher, Barona-Gomez, and Medema]{navarro2020bigscape}
Jorge~C. Navarro-Mu{\~n}oz, Nelly Selem-Mojica, Michael~W. Mullowney, Satria~A.
  Kautsar, James~H. Tryon, Elizabeth~I. Parkinson, Emmanuel L.~C. {De Los
  Santos}, Marley Yeong, Pablo Cruz-Morales, Sahar Abubucker, Arne Roeters,
  Wouter Lokhorst, Antonio Fernandez-Guerra, Luciana Teresa~Dias Cappelini,
  Anthony~W. Goering, Regan~J. Thomson, William~W. Metcalf, Neil~L. Kelleher,
  Francisco Barona-Gomez, and Marnix~H. Medema.
\newblock A computational framework to explore large-scale biosynthetic
  diversity.
\newblock \emph{Nature Chemical Biology}, 16\penalty0 (1):\penalty0 60--68,
  2020.
\newblock \doi{10.1038/s41589-019-0400-9}.

\bibitem[Draisma et~al.(2026)Draisma, Loureiro, Louwen, Kautsar,
  Navarro-Mu{\~n}oz, Doering, Mouncey, and Medema]{draisma2026bigscape2}
Arjan Draisma, Catarina Loureiro, Nico L.~L. Louwen, Satria~A. Kautsar,
  Jorge~C. Navarro-Mu{\~n}oz, Drew~T. Doering, Nigel~J. Mouncey, and Marnix~H.
  Medema.
\newblock {BiG-SCAPE} 2.0 and {BiG-SLiCE} 2.0: Scalable, accurate and
  interactive sequence clustering of metabolic gene clusters.
\newblock \emph{Nature Communications}, 17:\penalty0 2000, 2026.
\newblock \doi{10.1038/s41467-026-68733-5}.

\bibitem[Bemis and Murcko(1996)]{bemis1996murcko}
Guy~W. Bemis and Mark~A. Murcko.
\newblock The properties of known drugs. 1. {M}olecular frameworks.
\newblock \emph{Journal of Medicinal Chemistry}, 39\penalty0 (15):\penalty0
  2887--2893, 1996.
\newblock \doi{10.1021/jm9602928}.

\bibitem[Landrum et~al.(2026)]{rdkit}
Greg Landrum et~al.
\newblock {RDKit}: Open-source cheminformatics.
\newblock \url{https://www.rdkit.org}, 2026.
\newblock Version 2026.03.1. \doi{10.5281/zenodo.19250388}.

\bibitem[{Djoumbou Feunang} et~al.(2016){Djoumbou Feunang}, Eisner, Knox,
  Chepelev, Hastings, Owen, Fahy, Steinbeck, Subramanian, Bolton, Greiner, and
  Wishart]{classyfire}
Yannick {Djoumbou Feunang}, Roman Eisner, Craig Knox, Leonid Chepelev, Janna
  Hastings, Gareth Owen, Eoin Fahy, Christoph Steinbeck, Shankar Subramanian,
  Evan Bolton, Russell Greiner, and David~S. Wishart.
\newblock {ClassyFire}: Automated chemical classification with a comprehensive,
  computable taxonomy.
\newblock \emph{Journal of Cheminformatics}, 8:\penalty0 61, 2016.
\newblock \doi{10.1186/s13321-016-0174-y}.

\bibitem[Weininger(1988)]{weininger1988smiles}
David Weininger.
\newblock {SMILES}, a chemical language and information system. 1.
  {I}ntroduction to methodology and encoding rules.
\newblock \emph{Journal of Chemical Information and Computer Sciences},
  28\penalty0 (1):\penalty0 31--36, 1988.
\newblock \doi{10.1021/ci00057a005}.

\bibitem[Lipinski et~al.(1997)Lipinski, Lombardo, Dominy, and
  Feeney]{lipinski1997}
Christopher~A. Lipinski, Franco Lombardo, Beryl~W. Dominy, and Paul~J. Feeney.
\newblock Experimental and computational approaches to estimate solubility and
  permeability in drug discovery and development settings.
\newblock \emph{Advanced Drug Delivery Reviews}, 23\penalty0 (1--3):\penalty0
  3--25, 1997.
\newblock \doi{10.1016/S0169-409X(96)00423-1}.

\bibitem[Veber et~al.(2002)Veber, Johnson, Cheng, Smith, Ward, and
  Kopple]{veber2002}
Daniel~F. Veber, Stephen~R. Johnson, Hung-Yuan Cheng, Brian~R. Smith, Keith~W.
  Ward, and Kenneth~D. Kopple.
\newblock Molecular properties that influence the oral bioavailability of drug
  candidates.
\newblock \emph{Journal of Medicinal Chemistry}, 45\penalty0 (12):\penalty0
  2615--2623, 2002.
\newblock \doi{10.1021/jm020017n}.

\bibitem[Bickerton et~al.(2012)Bickerton, Paolini, Besnard, Muresan, and
  Hopkins]{bickerton2012qed}
G.~Richard Bickerton, Gaia~V. Paolini, J{\'e}r{\'e}my Besnard, Sorel Muresan,
  and Andrew~L. Hopkins.
\newblock Quantifying the chemical beauty of drugs.
\newblock \emph{Nature Chemistry}, 4\penalty0 (2):\penalty0 90--98, 2012.
\newblock \doi{10.1038/nchem.1243}.

\bibitem[Brown et~al.(2019)Brown, Fiscato, Segler, and
  Vaucher]{brown2019guacamol}
Nathan Brown, Marco Fiscato, Marwin H.~S. Segler, and Alain~C. Vaucher.
\newblock {GuacaMol}: Benchmarking models for de novo molecular design.
\newblock \emph{Journal of Chemical Information and Modeling}, 59\penalty0
  (3):\penalty0 1096--1108, 2019.
\newblock \doi{10.1021/acs.jcim.8b00839}.

\bibitem[Feher and Schmidt(2003)]{feher2003property}
Miklos Feher and Jonathan~M. Schmidt.
\newblock Property distributions: Differences between drugs, natural products,
  and molecules from combinatorial chemistry.
\newblock \emph{Journal of Chemical Information and Computer Sciences},
  43\penalty0 (1):\penalty0 218--227, 2003.
\newblock \doi{10.1021/ci0200467}.

\bibitem[Rogers and Hahn(2010)]{rogers2010ecfp}
David Rogers and Mathew Hahn.
\newblock Extended-connectivity fingerprints.
\newblock \emph{Journal of Chemical Information and Modeling}, 50\penalty0
  (5):\penalty0 742--754, 2010.
\newblock \doi{10.1021/ci100050t}.

\bibitem[Ross et~al.(2022)Ross, Belgodere, Chenthamarakshan, Padhi, Mroueh, and
  Das]{ross2022molformer}
Jerret Ross, Brian Belgodere, Vijil Chenthamarakshan, Inkit Padhi, Youssef
  Mroueh, and Payel Das.
\newblock Large-scale chemical language representations capture molecular
  structure and properties.
\newblock \emph{Nature Machine Intelligence}, 4\penalty0 (12):\penalty0
  1256--1264, 2022.
\newblock \doi{10.1038/s42256-022-00580-7}.

\bibitem[Radford et~al.(2021)Radford, Kim, Hallacy, Ramesh, Goh, Agarwal,
  Sastry, Askell, Mishkin, Clark, Krueger, and Sutskever]{radford2021clip}
Alec Radford, Jong~Wook Kim, Chris Hallacy, Aditya Ramesh, Gabriel Goh,
  Sandhini Agarwal, Girish Sastry, Amanda Askell, Pamela Mishkin, Jack Clark,
  Gretchen Krueger, and Ilya Sutskever.
\newblock Learning transferable visual models from natural language
  supervision.
\newblock In \emph{Proceedings of the 38th International Conference on Machine
  Learning}, volume 139 of \emph{Proceedings of Machine Learning Research},
  pages 8748--8763. PMLR, 2021.
\newblock URL \url{https://proceedings.mlr.press/v139/radford21a.html}.

\bibitem[Hacohen et~al.(2022)Hacohen, Dekel, and
  Weinshall]{hacohen2022typiclust}
Guy Hacohen, Avihu Dekel, and Daphna Weinshall.
\newblock Active learning on a budget: Opposite strategies suit high and low
  budgets.
\newblock In \emph{Proceedings of the 39th International Conference on Machine
  Learning}, volume 162 of \emph{Proceedings of Machine Learning Research},
  pages 8175--8195. PMLR, 2022.
\newblock URL \url{https://proceedings.mlr.press/v162/hacohen22a.html}.

\bibitem[Ash et~al.(2020)Ash, Zhang, Krishnamurthy, Langford, and
  Agarwal]{ash2020badge}
Jordan~T. Ash, Chicheng Zhang, Akshay Krishnamurthy, John Langford, and Alekh
  Agarwal.
\newblock Deep batch active learning by diverse, uncertain gradient lower
  bounds.
\newblock In \emph{International Conference on Learning Representations}, 2020.
\newblock URL \url{https://openreview.net/forum?id=ryghZJBKPS}.

\bibitem[Beck et~al.(2021)Beck, Sivasubramanian, Dani, Ramakrishnan, and
  Iyer]{beck2021effective}
Nathan Beck, Durga Sivasubramanian, Apurva Dani, Ganesh Ramakrishnan, and
  Rishabh Iyer.
\newblock Effective evaluation of deep active learning on image classification
  tasks.
\newblock arXiv preprint arXiv:2106.15324, 2021.
\newblock URL \url{https://arxiv.org/abs/2106.15324}.

\bibitem[Li et~al.(2025{\natexlab{b}})Li, Chiang, Frick, Dunlap, Wu, Zhu,
  Gonzalez, and Stoica]{li2025arenahard}
Tianle Li, Wei-Lin Chiang, Evan Frick, Lisa Dunlap, Tianhao Wu, Banghua Zhu,
  Joseph~E. Gonzalez, and Ion Stoica.
\newblock From crowdsourced data to high-quality benchmarks: {Arena-Hard} and
  {BenchBuilder} pipeline.
\newblock In \emph{Proceedings of the 42nd International Conference on Machine
  Learning}, volume 267 of \emph{Proceedings of Machine Learning Research},
  pages 34209--34231. PMLR, 2025{\natexlab{b}}.
\newblock URL \url{https://proceedings.mlr.press/v267/li25h.html}.

\bibitem[Ahmad et~al.(2022)Ahmad, Simon, Chithrananda, Grand, and
  Ramsundar]{ahmad2022chemberta2}
Walid Ahmad, Elana Simon, Seyone Chithrananda, Gabriel Grand, and Bharath
  Ramsundar.
\newblock {ChemBERTa-2}: Towards chemical foundation models.
\newblock arXiv preprint arXiv:2209.01712, 2022.
\newblock URL \url{https://arxiv.org/abs/2209.01712}.

\bibitem[Zhou et~al.(2023)Zhou, Gao, Ding, Zheng, Xu, Wei, Zhang, and
  Ke]{zhou2023unimol}
Gengmo Zhou, Zhifeng Gao, Qiankun Ding, Hang Zheng, Hongteng Xu, Zhewei Wei,
  Linfeng Zhang, and Guolin Ke.
\newblock {Uni-Mol}: A universal {3D} molecular representation learning
  framework.
\newblock In \emph{International Conference on Learning Representations}, 2023.
\newblock URL \url{https://openreview.net/forum?id=6K2RM6wVqKu}.

\bibitem[McIntosh-Smith et~al.(2024)McIntosh-Smith, Alam, and
  Woods]{mcintoshsmith2024isambardai}
Simon McIntosh-Smith, Sadaf~R. Alam, and Christopher Woods.
\newblock {Isambard-AI}: a leadership class supercomputer optimised
  specifically for artificial intelligence.
\newblock arXiv preprint arXiv:2410.11199, 2024.
\newblock URL \url{https://arxiv.org/abs/2410.11199}.

\bibitem[Butina(1999)]{butina1999clustering}
Darko Butina.
\newblock Unsupervised data base clustering based on {Daylight's} fingerprint
  and {Tanimoto} similarity: A fast and automated way to cluster small and
  large data sets.
\newblock \emph{Journal of Chemical Information and Computer Sciences},
  39\penalty0 (4):\penalty0 747--750, 1999.
\newblock \doi{10.1021/ci9803381}.

\bibitem[Xue and Sherman(2000)]{xue2000pikromycin}
Yongquan Xue and David~H. Sherman.
\newblock Alternative modular polyketide synthase expression controls
  macrolactone structure.
\newblock \emph{Nature}, 403\penalty0 (6769):\penalty0 571--575, 2000.
\newblock \doi{10.1038/35000624}.

\bibitem[Funa et~al.(1999)Funa, Ohnishi, Fujii, Shibuya, Ebizuka, and
  Horinouchi]{funa1999rppa}
Nobutaka Funa, Yasuo Ohnishi, Isao Fujii, Masaaki Shibuya, Yutaka Ebizuka, and
  Sueharu Horinouchi.
\newblock A new pathway for polyketide synthesis in microorganisms.
\newblock \emph{Nature}, 400\penalty0 (6747):\penalty0 897--899, 1999.
\newblock \doi{10.1038/23748}.

\bibitem[Fujii et~al.(2000)Fujii, Mori, Watanabe, Kubo, Tsuji, and
  Ebizuka]{fujii2000pks1}
Isao Fujii, Yuichiro Mori, Akira Watanabe, Yasuyuki Kubo, Gento Tsuji, and
  Yutaka Ebizuka.
\newblock Enzymatic synthesis of 1,3,6,8-tetrahydroxynaphthalene solely from
  malonyl coenzyme {A} by a fungal iterative type {I} polyketide synthase
  {PKS1}.
\newblock \emph{Biochemistry}, 39\penalty0 (30):\penalty0 8853--8858, 2000.
\newblock \doi{10.1021/bi000644j}.

\bibitem[Lovering et~al.(2009)Lovering, Bikker, and
  Humblet]{lovering2009escape}
Frank Lovering, Jack Bikker, and Christine Humblet.
\newblock Escape from flatland: Increasing saturation as an approach to
  improving clinical success.
\newblock \emph{Journal of Medicinal Chemistry}, 52\penalty0 (21):\penalty0
  6752--6756, 2009.
\newblock \doi{10.1021/jm901241e}.

\bibitem[Kuhn(2008)]{kuhn2008caret}
Max Kuhn.
\newblock Building predictive models in {R} using the {caret} package.
\newblock \emph{Journal of Statistical Software}, 28\penalty0 (5):\penalty0
  1--26, 2008.
\newblock \doi{10.18637/jss.v028.i05}.

\bibitem[Carhart et~al.(1985)Carhart, Smith, and
  Venkataraghavan]{carhart1985atompair}
Raymond~E. Carhart, Dennis~H. Smith, and R.~Venkataraghavan.
\newblock Atom pairs as molecular features in structure-activity studies:
  Definition and applications.
\newblock \emph{Journal of Chemical Information and Computer Sciences},
  25\penalty0 (2):\penalty0 64--73, 1985.
\newblock \doi{10.1021/ci00046a002}.

\bibitem[Durant et~al.(2002)Durant, Leland, Henry, and Nourse]{durant2002maccs}
Joseph~L. Durant, Burton~A. Leland, Douglas~R. Henry, and James~G. Nourse.
\newblock Reoptimization of {MDL} keys for use in drug discovery.
\newblock \emph{Journal of Chemical Information and Computer Sciences},
  42\penalty0 (6):\penalty0 1273--1280, 2002.
\newblock \doi{10.1021/ci010132r}.

\bibitem[Stachelhaus et~al.(1999)Stachelhaus, Mootz, and
  Marahiel]{stachelhaus1999specificity}
Torsten Stachelhaus, Henning~D. Mootz, and Mohamed~A. Marahiel.
\newblock The specificity-conferring code of adenylation domains in
  nonribosomal peptide synthetases.
\newblock \emph{Chemistry \& Biology}, 6\penalty0 (8):\penalty0 493--505, 1999.
\newblock \doi{10.1016/S1074-5521(99)80082-9}.

\end{thebibliography}
}

\clearpage
\appendix

\FloatBarrier
\section{Dataset construction and evaluation splits}
\label{app:data}

This section describes the MIBiG corpus, the construction of each evaluation
split, and the analyses used to characterize the resulting distribution
shifts. Task definitions and evaluation procedures follow in
Appendices~\ref{app:tasks} and~\ref{app:eval}.

\FloatBarrier
\subsection{Corpus construction}
\label{app:curation}
We build the corpus from MIBiG~$4.0$~\citep{mibig}, a curated repository of
experimentally characterized BGCs. First, we read biosynthetic classes,
product names and SMILES, organisms, and A-domain substrates from the official
JSON release and retain its $2{,}437$ active entries. Every active entry has
biosynthetic-class and organism annotations, and $1{,}944$ include
at least one product SMILES. Protein and nucleotide sequences come from the
corresponding GenBank files.

We derive natural-product class labels by applying NPClassifier~\citep{npclassifier}
to product SMILES and assigning each BGC the union of its products' classes.
We restrict this task to BGCs with annotated Pfam domains, retain classes
represented by at least ten such BGCs, and include entries with at least one
retained class.

We canonicalize product SMILES with RDKit, preserving specified
stereochemistry, and compute non-chiral Bemis--Murcko scaffolds. We construct
gene cluster families with BiG-SCAPE~\citep{navarro2020bigscape,draisma2026bigscape2}
at $c = 0.7$ for GCF-holdout evaluation. Split construction and leakage checks are
described in Appendix~\ref{app:splits}. Pfam annotations are mapped to each
encoder's vocabulary before embedding extraction.

\FloatBarrier
\subsection{Evaluation splits}
\label{app:splits}
We use four split families. \emph{Random} uses an 80/20 BGC split for coarse
class, chemical properties, and retrieval, iterative multi-label stratification
for natural-product classes, and substrate-stratified folds for A domains,
keeping all domains of a BGC together. \emph{Scaffold holdout} and \emph{GCF holdout}
hold out $20\,\%$ of groups, defined by first-product Bemis--Murcko scaffolds
and BiG-SCAPE families at $c = 0.7$, respectively. Products without a ring
system form a single scaffold group. \emph{Class holdout}, used for retrieval,
holds out one MIBiG class or the Hybrid group at a time.

Train and test BGC accessions are disjoint in every split. OOD holdouts also
keep their defining groups disjoint, while other grouping axes may overlap.
Scaffold groups use the first product, so additional products can share
scaffolds across the partition.

\FloatBarrier
\subsection{Sensitivity to grouping definitions}
\label{app:split-sensitivity}
We evaluated the grouping definitions with the twelve encoders and task
definitions of Table~\ref{tab:t3-main}, using one retrieval training seed per
split.

\paragraph{GCF cutoff.} We compared BiG-SCAPE cutoffs from $c = 0.3$ to $0.9$
by rebuilding the coarse-class GCF holdout on the same BGC cohort with one
split seed per cutoff (Table~\ref{tab:gcf-cutoff}). Residual train--test relatedness decreases up to
$c = 0.7$, with no consistent further reduction at higher cutoffs. We
therefore use $c = 0.7$ for GCF-holdout splits.

\begin{table}[!htbp]
  \caption{\textbf{Relatedness across GCF cutoffs.} NN distance is the median BiG-SCAPE distance to
  the nearest training BGC. AUROC uses class-specific nearest-neighbor similarity; arrows
  indicate less relatedness.}
  \label{tab:gcf-cutoff}
  \centering
  \small
  \setlength{\tabcolsep}{4pt}
    \begin{tabular}{lrrrrrrrr}
      \toprule
      & $c = 0.3$ & $c = 0.4$ & $c = 0.5$ & $c = 0.6$ & $c = 0.7$ & $c = 0.8$ & $c = 0.9$ & Random \\
      \midrule
      NN distance $\uparrow$ & 0.514 & 0.529 & 0.560 & 0.580 & 0.590 & 0.570 & 0.576 & 0.460 \\
      AUROC, BiG-SCAPE $\downarrow$ & 0.804 & 0.774 & 0.764 & 0.764 & 0.729 & 0.724 & 0.730 & 0.840 \\
      AUROC, Pfam Jaccard $\downarrow$ & 0.831 & 0.787 & 0.777 & 0.752 & 0.742 & 0.765 & 0.745 & 0.862 \\
      \bottomrule
    \end{tabular}
\end{table}

\paragraph{Alternative grouping tools.} We compared alternative groupings of
the same BGCs within each axis, using five split seeds per grouping and
Kendall's $\tau$ between the twelve linear-probe encoder rankings. Genotype
groupings used Butina clustering~\citep{butina1999clustering} of Pfam-domain
Jaccard distances, while phenotype groupings used generic Murcko scaffolds,
ECFP4 Butina clusters~\citep{rogers2010ecfp}, or the first product's full NPClassifier class set.
Clustering thresholds approximated the benchmark group counts. For chemical property
prediction, rank agreement between each alternative and the benchmark is
comparable to agreement across split seeds of the same grouping
(Table~\ref{tab:split-tools}).

\begin{table}[!htbp]
  \caption{\textbf{Encoder rankings under alternative groupings.} Chemical property prediction, linear probes. ARI (adjusted Rand index)
  compares partitions against the benchmark; $\tau$ averages ranking correlations within groupings or against the benchmark.}
  \label{tab:split-tools}
  \centering
  \small
    \begin{tabular}{lrrrr}
      \toprule
      Grouping & Groups & ARI & $\tau$ within & $\tau$ vs.\ benchmark \\
      \midrule
      \multicolumn{5}{l}{\emph{Genotype ($2{,}433$ BGCs)}} \\
      BiG-SCAPE families, $c = 0.7$ (benchmark) & 416 & -- & 0.806 & -- \\
      Pfam-domain Jaccard + Butina & 414 & 0.027 & 0.576 & 0.710 \\
      \midrule
      \multicolumn{5}{l}{\emph{Phenotype ($1{,}597$ BGCs)}} \\
      Bemis--Murcko scaffold (benchmark) & 1{,}105 & -- & 0.830 & -- \\
      Generic Murcko scaffold & 972 & 0.620 & 0.824 & 0.824 \\
      ECFP4 + Butina & 1{,}105 & 0.366 & 0.721 & 0.779 \\
      NPClassifier class set & 320 & 0.070 & 0.727 & 0.799 \\
      \bottomrule
    \end{tabular}
\end{table}

\FloatBarrier
\subsection{Genotype and phenotype partitions}
\label{app:gp-relationship}
Genotype and phenotype shifts probe different aspects of discovery. A BGC in
a held-out GCF belongs to a family absent from training but may still produce
a familiar chemical scaffold. Conversely, a held-out scaffold may be produced
by a BGC related to training examples. Evaluating both axes therefore asks
whether representations transfer across biosynthetic families, product
architectures, or both.

This distinction has a biochemical basis. Alternative module use within the
pikromycin assembly line yields 12- and 14-membered
macrolactones~\citep{xue2000pikromycin}, whereas unrelated bacterial type~III
and fungal type~I polyketide synthases can produce the same
tetrahydroxynaphthalene product~\citep{funa1999rppa,fujii2000pks1}. Genetic
relatedness and product-scaffold similarity therefore need not coincide.

On BGCs with both annotations, BiG-SCAPE families at $c = 0.7$ and standard
Bemis--Murcko scaffolds have an adjusted Rand index (ARI) of $0.052$. ARI
measures chance-corrected agreement: $1$ denotes identical partitions and $0$
denotes the agreement expected by chance. The low agreement indicates that
genotype and phenotype groupings capture different relationships among the BGCs.

We examined model behavior using twelve encoders with linear readouts across
five splits per axis on coarse-class prediction, chemical property prediction,
and retrieval, retaining each axis's eligible BGC pool. Aggregate within-axis
and cross-axis rank correlations are similar, while individual comparisons can
differ. In the single-split chemical-property analysis with MLP probes
(Table~\ref{tab:probe-mlp}), Kawano
ranks above Random-Pfam on GCF holdout, whereas their order reverses on
scaffold holdout. The two shifts thus do not always rank encoders the same
way, even though their overall ranking patterns are similar.

\FloatBarrier
\section{Task definitions}
\label{app:tasks}

For each task in Section~\ref{sec:bm:tasks}, we specify the input, target,
cohort, split, metric, and readout trained on the frozen embeddings. Shared
readout settings are listed in Appendix~\ref{app:probe-hp}.
Table~\ref{tab:task-specs} summarizes the training and test set sizes and
evaluation targets for the tasks in Table~\ref{tab:t3-main}. Coarse-class and
chemical-property evaluations use the BGCs covered by each encoder.

\begin{table}[!htbp]
  \caption{\textbf{Task sizes and targets.} Counts refer to BGCs except for A-domain prediction.
  Targets count evaluated classes, descriptors, substrates, or retrieval candidates;
  ranges span folds.}
  \label{tab:task-specs}
  \centering
  \small
    \begin{tabular}{lrrr}
      \toprule
      Split & Train & Test & Targets \\
      \midrule
      \multicolumn{4}{@{}l}{\emph{Coarse classes}: BGCs; MIBiG classes, multi-label; macro-AUROC} \\
      Random & 1{,}949 & 488 & 6 \\
      Scaffold & 1{,}570 & 374 & 6 \\
      GCF ($c = 0.7$) & 2{,}057 & 376 & 6 \\
      \midrule
      \multicolumn{4}{@{}l}{\emph{Fine classes}: BGCs; NPClassifier classes, multi-label; weighted macro-AUROC} \\
      Random (fold 4) & 980 & 241 & 48 \\
      Scaffold & 954 & 267 & 44 \\
      GCF ($c = 0.7$) & 1{,}009 & 212 & 40 \\
      \midrule
      \multicolumn{4}{@{}l}{\emph{Chemical properties}: BGCs; RDKit descriptors; mean Spearman $\rho$} \\
      Random & 1{,}565 & 379 & 13 \\
      Scaffold & 1{,}445 & 438 & 13 \\
      GCF ($c = 0.7$) & 1{,}640 & 301 & 13 \\
      \midrule
      \multicolumn{4}{@{}l}{\emph{BGC--molecule retrieval}: BGC queries; candidate molecules; class-macro R@10} \\
      Random & 1{,}552 & 387 & 713 \\
      Scaffold & 1{,}568 & 371 & 585 \\
      GCF ($c = 0.7$) & 1{,}637 & 302 & 451 \\
      Class (7 folds) & 1{,}399--1{,}878 & 61--540 & 47--862 \\
      \midrule
      \multicolumn{4}{@{}l}{\emph{A-domain substrate}: A domains; substrates; top-1 accuracy} \\
      5-fold CV & 1{,}621--1{,}624 & 403--406 & 28 \\
      GCF ($c = 0.7$) & 1{,}735 & 292 & 28 \\
      \bottomrule
    \end{tabular}
\end{table}

\FloatBarrier
\subsection{MIBiG coarse class prediction}
\label{app:task-mibig}
Each BGC receives all biosynthetic classes assigned in MIBiG~4.0, giving six
binary targets. The random split uses the active corpus; scaffold and GCF
holdouts require product SMILES and family assignments, respectively
(Appendix~\ref{app:splits}; Table~\ref{tab:task-specs}). We fit one-vs-rest
logistic regression to frozen embeddings and report AUROC averaged across
classes. Appendix~\ref{app:t1-perclass} gives the random-split per-class results.

\FloatBarrier
\subsection{NPClassifier class prediction}
\label{app:task-npc}
Each BGC receives the union of its products' NPClassifier classes
(Appendix~\ref{app:curation}). We train one-vs-rest logistic regression on
the same cohort and fixed $48$-class label space across encoders and splits.
The random result uses fold~4 of the stratified five-fold partition; scaffold
and GCF holdouts use the corresponding group assignments
(Appendix~\ref{app:splits}; Table~\ref{tab:task-specs}). We report a weighted
mean of per-class AUROCs, with weights proportional to test-positive counts.
Only classes with both positive and negative test BGCs enter this average;
the training label space remains fixed.

\FloatBarrier
\subsection{Chemical property prediction}
\label{app:task-chem}
The targets are $13$ RDKit descriptors computed from each BGC's first product
(Appendix~\ref{app:tchem-descriptors}). Random and GCF splits follow
coarse-class prediction, restricted to BGCs with product SMILES. The scaffold
split is drawn separately after excluding the $61$ BGCs labeled only as
Saccharide; the coarse-class and retrieval scaffold splits retain these BGCs. We fit
Ridge regression to the unscaled descriptor values and report Spearman
correlation averaged over descriptors. Table~\ref{tab:task-specs} gives split
sizes, and Appendix~\ref{app:probe-hp} gives probe settings.

\FloatBarrier
\subsubsection{Chemical descriptor panels}
\label{app:tchem-descriptors}
The main panel contains $13$ literature-derived RDKit descriptors~\citep{rdkit}
(Table~\ref{tab:tchem-descriptors}). Nine come from
GuacaMol~\citep{brown2019guacamol} and include the properties used in the
Lipinski and Veber rules~\citep{lipinski1997,veber2002}. We add
QED~\citep{bickerton2012qed} and three natural-product structure features:
the fraction of sp\textsuperscript{3} carbons~\citep{lovering2009escape,feher2003property},
stereocenter count~\citep{feher2003property}, and halogen atom count. Values
are computed from each BGC's first product SMILES.

Because these established descriptors are correlated, we also construct a
decorrelated panel to assess sensitivity to descriptor selection.

\begin{table}[!htbp]
\centering
\small
\caption{\textbf{Main RDKit descriptor panel.} Groups distinguish ADMET/drug-likeness from
  natural-product structure features.}
\label{tab:tchem-descriptors}
\begin{tabular}{lll}
\toprule
Descriptor & Group & Source \\
\midrule
MolWt & ADMET & Lipinski; GuacaMol \\
MolLogP & ADMET & Lipinski; GuacaMol \\
NumHDonors & ADMET & Lipinski; GuacaMol \\
NumHAcceptors & ADMET & Lipinski; GuacaMol \\
TPSA & ADMET & Veber; GuacaMol \\
NumRotatableBonds & ADMET & Veber; GuacaMol \\
qed & ADMET & QED \\
BertzCT & NP structure & GuacaMol \\
NumAliphaticRings & NP structure & GuacaMol \\
NumAromaticRings & NP structure & GuacaMol \\
FractionCSP3 & NP structure & Lovering \\
NumStereocenters & NP structure & Feher \\
HalogenAtomCount & NP structure & halogenase tailoring \\
\bottomrule
\end{tabular}
\end{table}

\paragraph{Data-driven descriptor panel.} We construct the alternative panel
in three steps:
\begin{enumerate}
  \item \textbf{Compute and filter descriptors.} We calculate all $217$ RDKit
  descriptors from the product SMILES and remove descriptors that are constant
  or non-zero in fewer than $10\,\%$ of products.
  \item \textbf{Remove redundant descriptors.} Starting with the most correlated
  pair, we remove one descriptor whenever the absolute Spearman correlation
  satisfies $|\rho| \geq 0.7$. We drop the member with the higher mean absolute
  correlation to the other retained descriptors, following caret's
  \texttt{findCorrelation} rule~\citep{kuhn2008caret}. We repeat until all
  remaining pairs satisfy $|\rho| < 0.7$.
  \item \textbf{Finalize the evaluation set.} The filters retain $66$
  descriptors. Before scoring, we exclude $17$ products containing metals or
  arsenic, for which selected charge-dependent descriptors cannot be calculated.
\end{enumerate}

The two descriptor panels yield similar encoder rankings under the linear
probe, with the same leading encoders across splits
(Table~\ref{tab:chem-panels}). Scores are generally lower with the data-driven
panel.

\begin{table}[!htbp]
  \caption{\textbf{Sensitivity to descriptor selection.} Scores are mean Spearman $\rho$ from
  linear probes; Kendall's $\tau$ compares encoder rankings between panels.}
  \label{tab:chem-panels}
  \centering
  \small
    \begin{tabular}{lrrrrrr}
      \toprule
      & \multicolumn{3}{c}{Literature panel (13)} & \multicolumn{3}{c}{Data-driven panel (66)} \\
      \cmidrule(lr){2-4}\cmidrule(lr){5-7}
      Encoder & Rand. & Scaf. & GCF & Rand. & Scaf. & GCF \\
      \midrule
      \multicolumn{7}{l}{\emph{Pfam-domain encoders}} \\
      Bag-of-Pfam        & 0.354 & 0.281 & 0.283 & 0.277 & 0.198 & 0.155 \\
      Random-Pfam        & 0.523 & 0.467 & 0.417 & 0.406 & 0.347 & 0.303 \\
      BiGCARP            & 0.574 & 0.519 & \textbf{0.509} & 0.430 & 0.370 & \textbf{0.340} \\
      Kawano             & 0.455 & 0.345 & 0.337 & 0.309 & 0.227 & 0.217 \\
      DeepBGC            & \textbf{0.591} & \textbf{0.558} & 0.482 & \textbf{0.440} & \textbf{0.388} & 0.333 \\
      \midrule
      \multicolumn{7}{l}{\emph{Amino-acid encoders}} \\
      Amino-acid 3-mer   & 0.506 & 0.428 & 0.332 & 0.396 & 0.317 & 0.228 \\
      BGC-MAC            & 0.466 & 0.395 & 0.355 & 0.356 & 0.279 & 0.249 \\
      BGCat              & 0.423 & 0.399 & 0.366 & 0.333 & 0.285 & 0.246 \\
      BGC-Prophet        & 0.500 & 0.419 & 0.412 & 0.378 & 0.323 & 0.289 \\
      LC-PLM             & 0.407 & 0.304 & 0.265 & 0.309 & 0.231 & 0.193 \\
      \midrule
      \multicolumn{7}{l}{\emph{Nucleotide encoders}} \\
      Nucleotide 6-mer   & 0.153 & 0.150 & 0.033 & 0.113 & 0.090 & 0.046 \\
      Evo 2 7B           & 0.442 & 0.409 & 0.355 & 0.328 & 0.278 & 0.239 \\
      \midrule
      Kendall $\tau$ & \multicolumn{3}{c}{} & 0.909 & 0.879 & 0.879 \\
      \bottomrule
    \end{tabular}
\end{table}

\FloatBarrier
\subsection{BGC--molecule retrieval}
\label{app:task-retrieval}
Each test BGC serves as a query against all distinct product molecules in
the test set. We use frozen embeddings for both BGCs and molecules, encoding
each molecule's canonical SMILES with MoLFormer~\citep{ross2022molformer}.
Table~\ref{tab:t3-main} reports BGC-to-compound retrieval (B$\to$C), while
Appendix~\ref{app:t3-grid} includes results in both directions.
Figure~\ref{fig:t3-chem-matrix} compares MoLFormer with six alternative
chemical encoders: ECFP4~\citep{rogers2010ecfp},
AtomPair~\citep{carhart1985atompair}, and MACCS~\citep{durant2002maccs}
fingerprints, RDKit descriptors~\citep{rdkit},
ChemBERTa-2~\citep{ahmad2022chemberta2}, and Uni-Mol~\citep{zhou2023unimol}.

The shared cohort contains $1{,}939$ BGCs, $3{,}110$ distinct molecules, and
$3{,}426$ curated BGC--product pairs (Appendix~\ref{app:curation}). A BGC may
have several products, and different BGCs can share a product. Each distinct
test-set molecule appears once in the candidate set. During training, each
BGC contributes one positive pair per epoch, cycling through its products.
Relation-masked InfoNCE excludes other known positive BGC--product pairs from
the in-batch negatives.

At evaluation, a query scores a hit if any of its known products appears
among the ten highest-ranked candidates. Exact ties receive their expected
score over possible orderings. We first average query scores within each
primary class (the six MIBiG classes and Hybrid), then average these class
means with equal weight to obtain class-macro R@10. The class-holdout result
averages the seven folds; the GCF split uses families at $c = 0.7$
(Appendix~\ref{app:splits}). Table~\ref{tab:task-specs} lists the training and
test set sizes and numbers of candidate molecules.

\FloatBarrier
\subsubsection{Retrieval head training}
\label{app:t3-protocol}
Two trainable linear projections map the frozen BGC and molecular embeddings
into a shared space. We train them with relation-masked InfoNCE. Every encoder uses
the same BGC cohort and training settings (Table~\ref{tab:t3-training}). We
fit the head on the training BGCs of each split and report test scores
averaged over three seeds.

\begin{table}[!htbp]
  \caption{\textbf{Shared training settings for retrieval heads.}}
  \label{tab:t3-training}
  \centering
  \small
  \begin{tabular}{lp{0.65\textwidth}}
    \toprule
    Setting & Value \\
    \midrule
    Shared dimension & $256$ \\
    Optimizer & AdamW \\
    Learning rate & $10^{-4}$ \\
    Weight decay & $10^{-4}$ \\
    Batch size & $256$ BGC--product pairs \\
    Temperature & Learnable; initialized at $0.07$, minimum $0.01$ \\
    Maximum epochs & $50$ \\
    Early stopping & Training loss fails to improve by $10^{-4}$ for five epochs \\
    \bottomrule
  \end{tabular}
\end{table}

\FloatBarrier
\subsection{A-domain substrate prediction}
\label{app:adomain}
Each example is an NRPS adenylation (A) domain labeled by its substrate.
Because the MIBiG annotations lack usable sequence coordinates, we match
annotated A domains to PF00501 (AMP-binding) hits by their order within each
protein and retain genes with matching counts. This matching yields
$2{,}504$ domains from $529$ BGCs. We keep domains with a single
known substrate and substrates represented by at least ten domains, yielding
$2{,}027$ domains from $479$ BGCs across $28$ substrate classes.

Domain embeddings are extracted as described in Section~\ref{sec:bm:tasks}.
The $k$-mer baselines use the same domain residues or corresponding codons as
the protein and nucleotide
models. Bag-of-Pfam uses the parent BGC's Pfam composition. Random-Pfam
assigns the same PF00501 vector to every A domain, yielding the majority
prediction. We fit multinomial logistic-regression probes
and measure top-1 accuracy under BGC-grouped five-fold cross-validation and
GCF holdout (Appendix~\ref{app:splits}). The cross-validation folds are
stratified by broad substrate groups and keep all domains from each BGC
together. We pool out-of-fold predictions to compute cross-validation
accuracy. Table~\ref{tab:task-specs} gives the split sizes.

\paragraph{Stachelhaus-code reference.} We fit the same probe to one-hot
encodings of nine specificity residues from the PF00501 alignment, following
the Stachelhaus code~\citep{stachelhaus1999specificity}. The conserved lysine
of the canonical ten-residue code lies outside the Pfam hit and is omitted.
We evaluate this reference on the same domains and splits as the encoders.
It achieves the highest top-1 accuracy on both splits
(Table~\ref{tab:adomain-stachelhaus}).

\begin{table}[!htbp]
  \caption{\textbf{A-domain substrate prediction.} Scores are top-1 accuracies.}
  \label{tab:adomain-stachelhaus}
  \centering
  \small
    \begin{tabular}{lcc}
      \toprule
      Features & 5-fold (BGC-grouped) & GCF holdout \\
      \midrule
      Stachelhaus code (9 residues) & \textbf{0.830} [0.814, 0.846] & \textbf{0.825} [0.781, 0.866] \\
      Amino-acid 3-mer (best encoder in Table~\ref{tab:t3-main}) & 0.781 [0.764, 0.799] & 0.747 [0.695, 0.791] \\
      Majority substrate & 0.106 & 0.127 \\
      \bottomrule
    \end{tabular}
\end{table}

\FloatBarrier
\section{Encoder representations and evaluation protocol}
\label{app:eval}

This section describes encoder inputs and pooling rules, shared readout
hyperparameters, and the variance protocol used for confidence intervals.

\FloatBarrier
\subsection{Encoder summary and downstream embedding extraction}
\label{app:encoders}

Table~\ref{tab:encoders} lists each encoder's backbone, pretraining setup,
vocabulary, and pooling rule. For BGC-MAC, BGCat, and BGC-Prophet, we use
pooled features from their frozen public protein backbones. To compare pooling
strategies, appendix tables also report BiGCARP-CLS, which replaces BiGCARP's
mean pooling with the state of a prepended token. The released DeepBGC detector
was trained with MIBiG~1.4 BGCs as detection positives, and $1{,}645$ of the
$2{,}437$ active MIBiG~4.0 entries share an accession with that training set.
It saw no class or chemical labels, but it was trained on these sequences.

\begin{table}[!tbp]
  \caption{\textbf{Encoder pretraining and embedding extraction.}}
  \label{tab:encoders}
  \centering
  \scriptsize
  \setlength{\tabcolsep}{3pt}
  \renewcommand{\arraystretch}{1.0}
  \begin{tabularx}{\textwidth}{@{}
    >{\raggedright\arraybackslash\hsize=0.85\hsize}X
    >{\raggedright\arraybackslash\hsize=1.0\hsize}X
    >{\raggedright\arraybackslash\hsize=0.95\hsize}X
    >{\raggedright\arraybackslash\hsize=0.8\hsize}X
    >{\raggedright\arraybackslash\hsize=0.75\hsize}X
    >{\raggedright\arraybackslash\hsize=1.65\hsize}X@{}}
    \toprule
    \textbf{Encoder} & \textbf{Backbone} & \textbf{Pretraining corpus} & \textbf{Vocabulary} & \textbf{Objective} & \textbf{Downstream pooling} \\
    \midrule
    Bag-of-Pfam                & none (frequency-based baseline) & no pretraining; raw Pfam-domain bag-of-counts representation & $3{,}932$ Pfam IDs                  & none                                & per-BGC Pfam-domain count vector, $L_2$-normalized ($3{,}932$-dim) \\
    Random-Pfam                & none (frequency-based baseline) & no pretraining; one fixed random vector per Pfam ID & $3{,}932$ Pfam IDs & none & i.i.d.\ $\mathcal{N}(0,1)$ vector per Pfam ID (seed 42), averaged over the cluster's Pfam tokens, $L_2$-normalized ($256$-dim) \\
    BiGCARP~\citep{bigcarp}    & ByteNet, $32$ layers, $d_{\text{model}}{=}256$, ${\sim}34$\,M params & antiSMASH-DB~\citep{blin2019antismashdb2} BGC Pfam-domain sequences ($127{,}294$ BGCs before train/validation/test split) & $19{,}509$ tokens & masked-token prediction (MLM)       & mean of last-layer token states across the Pfam sequence ($256$-dim) \\
    BiGCARP-CLS~\citep{bigcarp} (ablation) & same as BiGCARP                                                  & same as BiGCARP                                          & same as BiGCARP                & same as BiGCARP                     & last-layer state of a mask token prepended at position~0 ($256$-dim) \\
    Kawano~\citep{kawano2026roberta} & RoBERTa, $8$ layers, $16$ heads, hidden $1024$ ($\sim$110\,M params) & antiSMASH-DB bacterial BGC Pfam-domain sequences ($239{,}021$ BGCs; Model~I of four released models) & $19{,}722$-token vocabulary & masked-token prediction (MLM)       & mean of last-layer hidden states across the Pfam sequence, $\texttt{max\_length}{=}512$ ($1{,}024$-dim) \\
    DeepBGC~\citep{deepbgc}    & Pfam2vec ($100$-dim) {+} BiLSTM ($128$ units, bidirectional) & Pfam-domain sequences of $3{,}376$ bacterial genomes (Pfam2vec); MIBiG~1.4 BGCs ($1{,}811$ accessions) and $10{,}128$ synthetic negatives (BiLSTM; released v0.1.0 detector) & $13{,}311$ Pfam IDs & supervised binary BGC detection     & cluster boundary flags as input; mean last-layer BiLSTM states ($256$-dim) \\
    \addlinespace
    Amino-acid 3-mer           & none (frequency-based baseline) & no pretraining & $8{,}000$ 3-mers of the $20$ standard residues & none & 3-mers counted within each CDS, summed over the cluster, as frequencies ($8{,}000$-dim) \\
    BGC-MAC~\citep{lu2026bgcmac} & ESM-2~\citep{lin2023esm2} ($650$\,M, $33$ layers, hidden $1280$) & UniRef50 & amino-acid residues ($33$ tokens, ESM-2 alphabet) & masked-token prediction (MLM) & last-layer residue states averaged per CDS (first $1{,}022$ residues), then averaged over CDSs ($1{,}280$-dim) \\
    BGCat~\citep{porokhin2025bgcat} & ESM~C~\citep{candido2026esmc} ($600$\,M, hidden $1152$) & UniRef, MGnify, and JGI protein sequences & amino-acid residues (ESM~C alphabet, $33$ tokens) & masked-token prediction (MLM) & as BGC-MAC ($1{,}152$-dim) \\
    BGC-Prophet~\citep{bgcprophet} & ESM-2 ($8$\,M, $6$ layers, hidden $320$) & UniRef50 & amino-acid residues ($33$ tokens, ESM-2 alphabet) & masked-token prediction (MLM) & as BGC-MAC ($320$-dim) \\
    LC-PLM~\citep{wang2025lcplm} & BiMamba-S, $48$ bidirectional Mamba layers with shared projections, hidden $1{,}536$ (${\sim}750$\,M params) & UniRef50 & amino-acid residues (ESM-2 tokenizer) & masked-token prediction (MLM) & last-layer residue states averaged per full-length CDS (no $1{,}022$ cut), then over CDSs ($1{,}536$-dim) \\
    \addlinespace
    Nucleotide 6-mer           & none (frequency-based baseline) & no pretraining & $2{,}080$ canonical DNA 6-mers & none & strand-collapsed 6-mer frequencies over the cluster DNA, skipping 6-mers with non-ACGT bases ($2{,}080$-dim) \\
    Evo~2 7B~\citep{brixi2026evo2} & StripedHyena~2, $32$ blocks, hidden $4{,}096$ ($7$\,B params) & OpenGenome2 ($8.8$\,T nucleotides, all domains of life) & single nucleotides & next-nucleotide prediction & mean block-28 states in windows of up to $48$\,kb; length-weighted mean across windows ($4{,}096$-dim) \\
    \bottomrule
  \end{tabularx}
\end{table}

\subsection{Probe settings}
\label{app:probe-hp}
Classification and regression probes use frozen embeddings standardized with
statistics from the training fold. Table~\ref{tab:probe-settings} lists their
settings; MLP comparisons are reported in Appendix~\ref{app:probe-mlp}.

\begin{table}[!htbp]
  \caption{\textbf{Classification and regression probe settings.}}
  \label{tab:probe-settings}
  \centering
  \small
  \begin{tabular}{@{}p{0.32\textwidth}p{0.64\textwidth}@{}}
    \toprule
    Task or setting & Configuration \\
    \midrule
    \multicolumn{2}{@{}l}{\emph{Linear probes}} \\
    Multi-label classification & One-vs-rest logistic regression \\
    A-domain substrates & Multinomial logistic regression \\
    Logistic regression settings & L$_2$ penalty; $C = 1.0$; maximum $5{,}000$ iterations \\
    Chemical properties & Ridge regression; $\alpha = 1.0$ \\
    \midrule
    \multicolumn{2}{@{}l}{\emph{MLP classification and regression}} \\
    Hidden layers & $512$ and $256$ units; batch normalization, ReLU and dropout $0.3$ after each layer \\
    Optimization & Adam; learning rate $10^{-3}$; cosine schedule \\
    Batch size & $64$ \\
    Maximum epochs & $200$ \\
    Validation set & $15\,\%$ of the training fold \\
    Early stopping & Validation loss; patience $10$ epochs; restore best weights \\
    \bottomrule
  \end{tabular}
\end{table}

For regression, Ridge fits targets in their original units. The MLP
standardizes each target using its internal training subset, minimizes mean
squared error in that space, and converts predictions back to the original
units.

Retrieval does not standardize the input embeddings and follows the training
protocol in Appendix~\ref{app:t3-protocol}. Active-learning runs refit the
task readout at each acquisition round (Appendix~\ref{app:al-protocol}).

\subsection{Variance protocol and bootstrap confidence intervals}
\label{app:variance}
We compute $95\,\%$ percentile confidence intervals for the primary metrics
in Table~\ref{tab:t3-main} using $500$ bootstrap resamples of the held-out
examples, with a fixed resampling seed. For coarse-class, natural-product
class, and chemical-property prediction, we resample test BGCs. For substrate
prediction, we resample individual test A domains without grouping them by
parent BGC; for retrieval, we resample queries within biosynthetic classes.

For retrieval, we train the head with three random seeds and bootstrap each
run separately. We report the envelope of the three intervals, defined by the
lowest lower bound and highest upper bound. For natural-product class
prediction, the random-split result uses fixed fold~4 of the five-fold
partition, so its intervals do not include variation across folds.
Active-learning comparisons use seeds as the unit of resampling and
statistical testing (Appendix~\ref{app:al-protocol}).

For each column of Table~\ref{tab:t3-main}, the Separability row gives the
fraction of encoder pairs with non-overlapping $95\,\%$ intervals, using
the original interval bounds (Table~\ref{tab:main-ci}).

\FloatBarrier
\subsection{Compute resources}
\label{app:compute}
Experiments ran on CPU workers and NVIDIA A100, L40S, H200, and GH200 GPUs at
the University of Manchester Computational Shared Facility and on Isambard-AI.
With the release scripts, embedding extraction and scoring for all twelve
encoders take about $10$ hours on one NVIDIA A100 GPU with $80$~GB of memory,
with per-encoder embedding times ranging from seconds on a CPU for the
frequency-based baselines to $5.2$ hours for Evo~2 7B on an NVIDIA L40S GPU, and
the $3{,}240$ active-learning curves take about $30$ CPU-hours.

\FloatBarrier
\section{Task-level results}
\label{app:results}

This section provides confidence intervals and task-level breakdowns for
Table~\ref{tab:t3-main}. Additional A-domain results appear in
Appendix~\ref{app:adomain}; sensitivity analyses follow in
Appendix~\ref{app:probe}.

\FloatBarrier
\subsection{Main-result confidence intervals}
\label{app:main-ci}

Table~\ref{tab:main-ci} reports the full confidence intervals for the main
results. Task definitions and resampling procedures are given in
Appendices~\ref{app:tasks} and~\ref{app:variance}.

\begin{table}[!htbp]
  \caption{\textbf{Confidence intervals for Table~\ref{tab:t3-main}.} Superscripts and subscripts
  show distances to the upper and lower $95\,\%$ bounds.}
  \label{tab:main-ci}
  \centering
  \scriptsize
  \setlength{\tabcolsep}{1.2pt}
  \renewcommand{\arraystretch}{1.2}
  \begin{tabular}{l*{9}{c}}
  \toprule
  \multicolumn{10}{l}{\textbf{(a) Classification and chemical properties}} \\
  \midrule
  Model & \multicolumn{3}{c}{Coarse classes} & \multicolumn{3}{c}{Fine classes} & \multicolumn{3}{c}{Chem.\ properties} \\
   & \multicolumn{3}{c}{\tiny macro-AUROC} & \multicolumn{3}{c}{\tiny weighted macro-AUROC} & \multicolumn{3}{c}{\tiny mean Spearman $\rho$} \\
  \cmidrule(lr){2-4}\cmidrule(lr){5-7}\cmidrule(lr){8-10}
   & Rand. & Scaf. & GCF & Rand. & Scaf. & GCF & Rand. & Scaf. & GCF \\
  \midrule
  \multicolumn{10}{l}{\emph{Pfam-domain encoders}} \\
  Bag-of-Pfam & \ci{0.929}{0.020}{0.020} & \ci{0.877}{0.036}{0.042} & \ci{0.884}{0.027}{0.029} & \ci{0.880}{0.023}{0.031} & \ci{0.825}{0.028}{0.028} & \ci{0.753}{0.041}{0.039} & \ci{0.354}{0.052}{0.057} & \ci{0.281}{0.048}{0.057} & \ci{0.283}{0.045}{0.054} \\
  Random-Pfam & \ci{0.936}{0.017}{0.014} & \ci{0.909}{0.023}{0.026} & \ci{0.896}{0.020}{0.022} & \ci{0.880}{0.023}{0.027} & \ci{0.851}{0.025}{0.026} & \ci{0.786}{0.033}{0.033} & \ci{0.523}{0.043}{0.049} & \ci{0.467}{0.042}{0.053} & \ci{0.417}{0.051}{0.061} \\
  BiGCARP & \ci{0.952}{0.015}{0.014} & \ci{0.935}{0.019}{0.022} & \ci{0.909}{0.024}{0.023} & \ci{0.880}{0.025}{0.027} & \ci{0.872}{0.022}{0.023} & \ci{0.834}{0.031}{0.028} & \ci{0.574}{0.041}{0.045} & \ci{0.519}{0.038}{0.051} & \ci{0.509}{0.053}{0.056} \\
  Kawano & \ci{0.934}{0.017}{0.016} & \ci{0.926}{0.024}{0.028} & \ci{0.911}{0.019}{0.020} & \ci{0.873}{0.025}{0.029} & \ci{0.844}{0.024}{0.023} & \ci{0.810}{0.030}{0.031} & \ci{0.455}{0.044}{0.051} & \ci{0.345}{0.045}{0.054} & \ci{0.337}{0.059}{0.054} \\
  DeepBGC & \ci{0.953}{0.015}{0.015} & \ci{0.945}{0.019}{0.023} & \ci{0.922}{0.019}{0.020} & \ci{0.893}{0.021}{0.026} & \ci{0.863}{0.021}{0.022} & \ci{0.830}{0.027}{0.029} & \ci{0.591}{0.038}{0.043} & \ci{0.558}{0.037}{0.045} & \ci{0.482}{0.047}{0.054} \\
  \midrule
  \multicolumn{10}{l}{\emph{Amino-acid encoders}} \\
  3-mer (aa) & \ci{0.920}{0.021}{0.020} & \ci{0.913}{0.027}{0.029} & \ci{0.863}{0.031}{0.036} & \ci{0.887}{0.023}{0.027} & \ci{0.850}{0.027}{0.029} & \ci{0.822}{0.036}{0.037} & \ci{0.506}{0.052}{0.055} & \ci{0.428}{0.048}{0.050} & \ci{0.332}{0.056}{0.062} \\
  BGC-MAC & \ci{0.948}{0.013}{0.014} & \ci{0.920}{0.020}{0.021} & \ci{0.917}{0.018}{0.019} & \ci{0.853}{0.032}{0.039} & \ci{0.812}{0.030}{0.029} & \ci{0.763}{0.033}{0.039} & \ci{0.466}{0.049}{0.057} & \ci{0.395}{0.051}{0.051} & \ci{0.355}{0.057}{0.062} \\
  BGCat & \ci{0.930}{0.018}{0.019} & \ci{0.913}{0.018}{0.020} & \ci{0.891}{0.024}{0.027} & \ci{0.847}{0.035}{0.039} & \ci{0.789}{0.034}{0.033} & \ci{0.750}{0.034}{0.039} & \ci{0.423}{0.049}{0.055} & \ci{0.399}{0.046}{0.052} & \ci{0.366}{0.056}{0.058} \\
  BGC-Prophet & \ci{0.939}{0.014}{0.016} & \ci{0.913}{0.018}{0.020} & \ci{0.902}{0.020}{0.023} & \ci{0.854}{0.028}{0.031} & \ci{0.833}{0.024}{0.027} & \ci{0.763}{0.033}{0.037} & \ci{0.500}{0.044}{0.046} & \ci{0.419}{0.050}{0.055} & \ci{0.412}{0.055}{0.056} \\
  LC-PLM & \ci{0.940}{0.015}{0.015} & \ci{0.921}{0.017}{0.021} & \ci{0.905}{0.020}{0.025} & \ci{0.843}{0.032}{0.036} & \ci{0.810}{0.026}{0.026} & \ci{0.701}{0.038}{0.048} & \ci{0.407}{0.048}{0.051} & \ci{0.304}{0.052}{0.053} & \ci{0.265}{0.058}{0.057} \\
  \midrule
  \multicolumn{10}{l}{\emph{Nucleotide encoders}} \\
  6-mer (nt) & \ci{0.808}{0.030}{0.030} & \ci{0.773}{0.034}{0.033} & \ci{0.701}{0.037}{0.038} & \ci{0.834}{0.027}{0.029} & \ci{0.778}{0.030}{0.029} & \ci{0.740}{0.032}{0.036} & \ci{0.153}{0.057}{0.053} & \ci{0.150}{0.043}{0.046} & \ci{0.033}{0.065}{0.065} \\
  Evo 2 7B & \ci{0.896}{0.017}{0.022} & \ci{0.857}{0.025}{0.029} & \ci{0.858}{0.029}{0.033} & \ci{0.801}{0.030}{0.034} & \ci{0.750}{0.033}{0.033} & \ci{0.751}{0.035}{0.037} & \ci{0.442}{0.049}{0.050} & \ci{0.409}{0.042}{0.050} & \ci{0.355}{0.056}{0.057} \\
  \bottomrule
  \end{tabular}

  \vspace{6pt}

  \begin{tabular}{l*{6}{c}}
  \toprule
  \multicolumn{7}{l}{\textbf{(b) Retrieval and A-domain substrate}} \\
  \midrule
  Model & \multicolumn{4}{c}{BGC--molecule retrieval} & \multicolumn{2}{c}{A-domain substrate} \\
   & \multicolumn{4}{c}{\tiny class-macro R@10} & \multicolumn{2}{c}{\tiny top-1 accuracy} \\
  \cmidrule(lr){2-5}\cmidrule(lr){6-7}
   & Rand. & Scaf. & GCF & Class & 5-fold & GCF \\
  \midrule
  \multicolumn{7}{l}{\emph{Pfam-domain encoders}} \\
  Bag-of-Pfam & \ci{0.624}{0.060}{0.062} & \ci{0.546}{0.107}{0.109} & \ci{0.495}{0.075}{0.081} & \ci{0.275}{0.039}{0.033} & \ci{0.196}{0.018}{0.018} & \ci{0.151}{0.043}{0.043} \\
  Random-Pfam & \ci{0.542}{0.065}{0.068} & \ci{0.504}{0.092}{0.119} & \ci{0.448}{0.097}{0.105} & \ci{0.209}{0.031}{0.029} & -- & -- \\
  BiGCARP & \ci{0.559}{0.065}{0.073} & \ci{0.466}{0.121}{0.118} & \ci{0.435}{0.087}{0.083} & \ci{0.222}{0.031}{0.031} & \ci{0.233}{0.019}{0.021} & \ci{0.106}{0.038}{0.034} \\
  Kawano & \ci{0.532}{0.072}{0.075} & \ci{0.382}{0.107}{0.087} & \ci{0.387}{0.092}{0.081} & \ci{0.187}{0.029}{0.032} & \ci{0.234}{0.019}{0.020} & \ci{0.137}{0.038}{0.038} \\
  DeepBGC & \ci{0.440}{0.072}{0.082} & \ci{0.397}{0.083}{0.102} & \ci{0.407}{0.070}{0.072} & \ci{0.163}{0.026}{0.026} & -- & -- \\
  \midrule
  \multicolumn{7}{l}{\emph{Amino-acid encoders}} \\
  3-mer (aa) & \ci{0.549}{0.061}{0.085} & \ci{0.418}{0.115}{0.091} & \ci{0.318}{0.090}{0.078} & \ci{0.198}{0.034}{0.036} & \ci{0.781}{0.018}{0.018} & \ci{0.747}{0.045}{0.051} \\
  BGC-MAC & \ci{0.507}{0.074}{0.078} & \ci{0.507}{0.072}{0.108} & \ci{0.421}{0.075}{0.079} & \ci{0.220}{0.033}{0.031} & \ci{0.759}{0.018}{0.019} & \ci{0.736}{0.045}{0.055} \\
  BGCat & \ci{0.469}{0.069}{0.077} & \ci{0.485}{0.052}{0.056} & \ci{0.373}{0.074}{0.070} & \ci{0.201}{0.037}{0.033} & \ci{0.758}{0.018}{0.020} & \ci{0.733}{0.048}{0.051} \\
  BGC-Prophet & \ci{0.461}{0.071}{0.078} & \ci{0.431}{0.104}{0.124} & \ci{0.406}{0.090}{0.100} & \ci{0.189}{0.026}{0.026} & \ci{0.656}{0.020}{0.020} & \ci{0.610}{0.057}{0.057} \\
  LC-PLM & \ci{0.250}{0.080}{0.083} & \ci{0.250}{0.100}{0.088} & \ci{0.184}{0.067}{0.057} & \ci{0.137}{0.029}{0.034} & \ci{0.745}{0.017}{0.018} & \ci{0.740}{0.048}{0.055} \\
  \midrule
  \multicolumn{7}{l}{\emph{Nucleotide encoders}} \\
  6-mer (nt) & \ci{0.299}{0.058}{0.066} & \ci{0.163}{0.124}{0.053} & \ci{0.150}{0.059}{0.057} & \ci{0.151}{0.033}{0.034} & \ci{0.597}{0.021}{0.022} & \ci{0.466}{0.053}{0.058} \\
  Evo 2 7B & \ci{0.183}{0.059}{0.063} & \ci{0.128}{0.044}{0.037} & \ci{0.161}{0.070}{0.063} & \ci{0.116}{0.033}{0.029} & \ci{0.562}{0.022}{0.023} & \ci{0.527}{0.051}{0.058} \\
  \bottomrule
  \end{tabular}
\end{table}

\clearpage
\subsection{Per-class breakdown of biosynthetic class prediction}
\label{app:t1-perclass}

Table~\ref{tab:t1-perclass} separates the coarse-class result into its six
biosynthetic classes. Among non-nucleotide encoders, most differences arise
from Saccharide and Other; performance on PKS, NRPS, RiPP, and Terpene is
uniformly high.

\begin{table}[!htbp]
  \caption{\textbf{Per-class biosynthetic prediction.} Scores are random-split AUROCs from linear
  probes.}
  \label{tab:t1-perclass}
  \centering
  \small
    \begin{tabular}{lrrrrrr}
      \toprule
      Encoder & PKS & NRPS & Ribosomal (RiPP) & Terpene & Saccharide & Other \\
      \midrule
      \multicolumn{7}{l}{\emph{Pfam-domain encoders}} \\
      Bag-of-Pfam      & 0.950 & 0.941 & \textbf{0.997} & 0.942 & 0.868 & 0.873 \\
      Random-Pfam      & 0.962 & 0.952 & 0.992 & 0.963 & 0.892 & 0.856 \\
      BiGCARP          & 0.973 & 0.961 & 0.980 & 0.969 & 0.935 & 0.896 \\
      BiGCARP-CLS      & 0.954 & 0.933 & 0.993 & 0.937 & 0.912 & 0.842 \\
      Kawano           & 0.966 & 0.965 & 0.966 & 0.957 & 0.921 & 0.833 \\
      DeepBGC          & \textbf{0.975} & \textbf{0.969} & 0.991 & \textbf{0.971} & 0.915 & \textbf{0.899} \\
      \midrule
      \multicolumn{7}{l}{\emph{Amino-acid encoders}} \\
      Amino-acid 3-mer & 0.957 & 0.960 & 0.946 & 0.935 & 0.859 & 0.864 \\
      BGC-MAC          & 0.961 & 0.944 & 0.995 & 0.969 & 0.945 & 0.876 \\
      BGCat            & 0.944 & 0.938 & 0.985 & 0.944 & 0.908 & 0.859 \\
      BGC-Prophet      & 0.957 & 0.940 & 0.979 & 0.963 & \textbf{0.952} & 0.846 \\
      LC-PLM           & 0.952 & 0.929 & 0.991 & 0.957 & 0.933 & 0.876 \\
      \midrule
      \multicolumn{7}{l}{\emph{Nucleotide encoders}} \\
      Nucleotide 6-mer & 0.825 & 0.847 & 0.883 & 0.789 & 0.811 & 0.693 \\
      Evo 2 7B         & 0.925 & 0.902 & 0.976 & 0.874 & 0.871 & 0.830 \\
      \bottomrule
    \end{tabular}
\end{table}

\FloatBarrier
\subsection{Chemical property results by encoder, split, and probe}
\label{app:chem}

\begin{figure}[!htbp]
  \centering
  \includegraphics[width=0.8\textwidth]{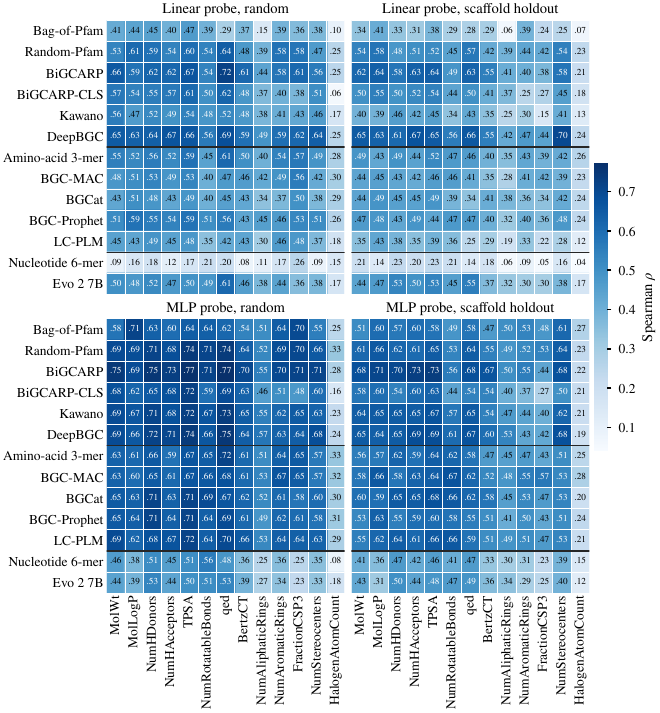}
  \caption{\textbf{Chemical-property prediction by descriptor.} Panel columns compare random and
  scaffold splits; panel rows compare Ridge and MLP probes on a shared Spearman $\rho$
  scale.}
  \label{fig:chem-heatmap}
\end{figure}

Table~\ref{tab:chem-spearman} summarizes linear-probe predictions over $13$
descriptors and four splits: random, scaffold, and GCF at $c = 0.7$ and $0.5$.
The averaged confidence bounds summarize split-specific uncertainty; they are
not a confidence interval for the overall mean.
Figure~\ref{fig:chem-heatmap} compares linear and MLP predictions for
individual descriptors.

\begin{table}[!htbp]
  \caption{\textbf{Chemical-property prediction: mean Spearman $\rho$.} Bounds average split-specific
  CI endpoints.}
  \label{tab:chem-spearman}
  \centering
  \small
    \begin{tabular}{lrr}
      \toprule
      Encoder              & $\bar{\rho}$ & Mean split-CI bounds \\
      \midrule
      DeepBGC                & \textbf{0.546} & [0.499, 0.587] \\
      BiGCARP                & 0.535 & [0.485, 0.580] \\
      Random-Pfam            & 0.469 & [0.415, 0.515] \\
      BGC-Prophet            & 0.452 & [0.401, 0.499] \\
      BiGCARP-CLS            & 0.442 & [0.392, 0.484] \\
      Amino-acid 3-mer       & 0.431 & [0.375, 0.482] \\
      Evo 2 7B               & 0.413 & [0.362, 0.461] \\
      BGCat                  & 0.403 & [0.348, 0.452] \\
      BGC-MAC                & 0.398 & [0.342, 0.448] \\
      Kawano                 & 0.393 & [0.342, 0.441] \\
      LC-PLM                 & 0.309 & [0.255, 0.362] \\
      Bag-of-Pfam            & 0.295 & [0.238, 0.345] \\
      Nucleotide 6-mer       & 0.110 & [0.054, 0.165] \\
      \bottomrule
    \end{tabular}
\end{table}

DeepBGC and BiGCARP lead the linear-probe comparison.

\FloatBarrier

Size- and polarity-related descriptors are generally predicted more
accurately than halogen counts across both probes and splits.

\clearpage
\subsection{BGC--molecule retrieval across encoders and splits}
\label{app:t3-grid}

Table~\ref{tab:t3-grid} extends the main retrieval results to both directions.
Bag-of-Pfam leads or ties in most settings, and every encoder performs worse
on class holdout than on the random split.

\begin{table}[!htbp]
  \caption{\textbf{Retrieval in both directions.} Class-macro R@10 for BGC-to-compound (left) and
  compound-to-BGC (right); mean $\pm$ standard deviation over three seeds.}
  \label{tab:t3-grid}
  \centering
  \footnotesize
  \setlength{\tabcolsep}{1.5pt}
    \resizebox{\textwidth}{!}{%
    \begin{tabular}{lrrrrrrrr}
      \toprule
      & \multicolumn{4}{c}{B$\to$C} & \multicolumn{4}{c}{C$\to$B} \\
      \cmidrule(lr){2-5}\cmidrule(lr){6-9}
      Encoder & Rand. & Scaf. & GCF & Class & Rand. & Scaf. & GCF & Class \\
      \midrule
      \multicolumn{9}{l}{\emph{Pfam-domain encoders}} \\
      Bag-of-Pfam      & \textbf{0.624}\,{\tiny $\pm$0.006} & \textbf{0.546}\,{\tiny $\pm$0.021} & \textbf{0.495}\,{\tiny $\pm$0.019} & \textbf{0.275}\,{\tiny $\pm$0.012} & \textbf{0.633}\,{\tiny $\pm$0.009} & \textbf{0.607}\,{\tiny $\pm$0.029} & \textbf{0.493}\,{\tiny $\pm$0.013} & 0.286\,{\tiny $\pm$0.007} \\
      Random-Pfam      & 0.542\,{\tiny $\pm$0.005} & 0.504\,{\tiny $\pm$0.041} & 0.448\,{\tiny $\pm$0.032} & 0.209\,{\tiny $\pm$0.006} & 0.553\,{\tiny $\pm$0.003} & 0.544\,{\tiny $\pm$0.043} & 0.466\,{\tiny $\pm$0.022} & \textbf{0.288}\,{\tiny $\pm$0.015} \\
      BiGCARP          & 0.559\,{\tiny $\pm$0.004} & 0.466\,{\tiny $\pm$0.037} & 0.435\,{\tiny $\pm$0.020} & 0.222\,{\tiny $\pm$0.007} & 0.556\,{\tiny $\pm$0.005} & 0.503\,{\tiny $\pm$0.011} & 0.470\,{\tiny $\pm$0.021} & 0.256\,{\tiny $\pm$0.002} \\
      BiGCARP-CLS      & 0.372\,{\tiny $\pm$0.024} & 0.291\,{\tiny $\pm$0.024} & 0.248\,{\tiny $\pm$0.020} & 0.119\,{\tiny $\pm$0.005} & 0.373\,{\tiny $\pm$0.031} & 0.429\,{\tiny $\pm$0.021} & 0.263\,{\tiny $\pm$0.024} & 0.140\,{\tiny $\pm$0.004} \\
      Kawano           & 0.532\,{\tiny $\pm$0.014} & 0.382\,{\tiny $\pm$0.009} & 0.387\,{\tiny $\pm$0.019} & 0.187\,{\tiny $\pm$0.010} & 0.541\,{\tiny $\pm$0.025} & 0.459\,{\tiny $\pm$0.007} & 0.394\,{\tiny $\pm$0.014} & 0.219\,{\tiny $\pm$0.016} \\
      DeepBGC          & 0.440\,{\tiny $\pm$0.014} & 0.397\,{\tiny $\pm$0.005} & 0.407\,{\tiny $\pm$0.003} & 0.163\,{\tiny $\pm$0.005} & 0.465\,{\tiny $\pm$0.004} & 0.434\,{\tiny $\pm$0.033} & 0.437\,{\tiny $\pm$0.008} & 0.221\,{\tiny $\pm$0.006} \\
      \midrule
      \multicolumn{9}{l}{\emph{Amino-acid encoders}} \\
      Amino-acid 3-mer & 0.549\,{\tiny $\pm$0.019} & 0.418\,{\tiny $\pm$0.012} & 0.318\,{\tiny $\pm$0.023} & 0.198\,{\tiny $\pm$0.011} & 0.519\,{\tiny $\pm$0.022} & 0.456\,{\tiny $\pm$0.019} & 0.360\,{\tiny $\pm$0.016} & 0.235\,{\tiny $\pm$0.018} \\
      BGC-MAC          & 0.507\,{\tiny $\pm$0.007} & 0.507\,{\tiny $\pm$0.022} & 0.421\,{\tiny $\pm$0.010} & 0.220\,{\tiny $\pm$0.008} & 0.506\,{\tiny $\pm$0.021} & \textbf{0.607}\,{\tiny $\pm$0.022} & 0.479\,{\tiny $\pm$0.013} & 0.271\,{\tiny $\pm$0.006} \\
      BGCat            & 0.469\,{\tiny $\pm$0.008} & 0.485\,{\tiny $\pm$0.011} & 0.373\,{\tiny $\pm$0.010} & 0.201\,{\tiny $\pm$0.011} & 0.478\,{\tiny $\pm$0.010} & 0.548\,{\tiny $\pm$0.013} & 0.400\,{\tiny $\pm$0.015} & 0.254\,{\tiny $\pm$0.010} \\
      BGC-Prophet      & 0.461\,{\tiny $\pm$0.009} & 0.431\,{\tiny $\pm$0.027} & 0.406\,{\tiny $\pm$0.029} & 0.189\,{\tiny $\pm$0.004} & 0.445\,{\tiny $\pm$0.013} & 0.517\,{\tiny $\pm$0.046} & 0.442\,{\tiny $\pm$0.015} & 0.258\,{\tiny $\pm$0.007} \\
      LC-PLM           & 0.250\,{\tiny $\pm$0.020} & 0.250\,{\tiny $\pm$0.024} & 0.184\,{\tiny $\pm$0.008} & 0.137\,{\tiny $\pm$0.011} & 0.249\,{\tiny $\pm$0.013} & 0.250\,{\tiny $\pm$0.021} & 0.244\,{\tiny $\pm$0.014} & 0.175\,{\tiny $\pm$0.010} \\
      \midrule
      \multicolumn{9}{l}{\emph{Nucleotide encoders}} \\
      Nucleotide 6-mer & 0.299\,{\tiny $\pm$0.005} & 0.163\,{\tiny $\pm$0.029} & 0.150\,{\tiny $\pm$0.006} & 0.151\,{\tiny $\pm$0.012} & 0.294\,{\tiny $\pm$0.035} & 0.243\,{\tiny $\pm$0.013} & 0.187\,{\tiny $\pm$0.023} & 0.171\,{\tiny $\pm$0.009} \\
      Evo 2 7B         & 0.183\,{\tiny $\pm$0.002} & 0.128\,{\tiny $\pm$0.008} & 0.161\,{\tiny $\pm$0.014} & 0.116\,{\tiny $\pm$0.011} & 0.209\,{\tiny $\pm$0.013} & 0.207\,{\tiny $\pm$0.023} & 0.210\,{\tiny $\pm$0.009} & 0.134\,{\tiny $\pm$0.002} \\
      \bottomrule
    \end{tabular}%
    }
\end{table}

\begin{sidewaysfigure}
  \centering
  \includegraphics{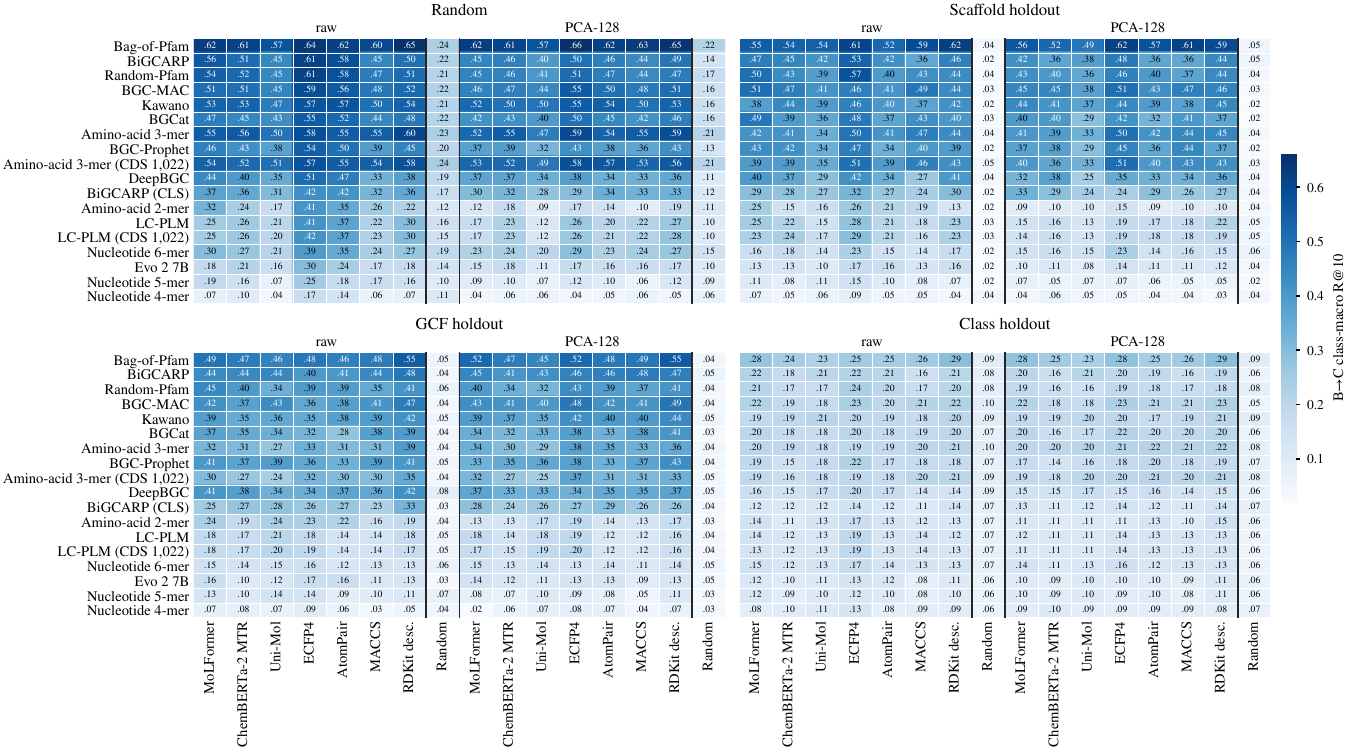}
  \caption{\textbf{BGC and chemical encoder combinations.} BGC-to-compound class-macro R@10,
  averaged over three seeds, with native chemical embeddings and PCA-$128$ projections.}
  \label{fig:t3-chem-matrix-full}
\end{sidewaysfigure}

\FloatBarrier
\paragraph{BGC and chemical encoder combinations.}
Figure~\ref{fig:t3-chem-matrix-full} extends the chemical-encoder comparison
to all splits and additional BGC representations. Chemical PCA is fitted on
all product molecules. The CDS controls keep the first $1{,}022$ residues of
each protein, and the random control assigns one vector per product. BGC encoder choice
accounts for most of the variation in retrieval performance. Bag-of-Pfam and
descriptor-based chemical encoders remain strong when the chemical
embeddings are reduced to a common width.

\clearpage
\FloatBarrier

\section{Sensitivity analyses and ablations}
\label{app:probe}

In this section, we examine how alternative evaluation settings affect
predictive performance and encoder rankings. We vary probe capacity,
embedding dimensionality, continued pretraining, and the source of chemical
labels.

\FloatBarrier
\subsection{Probe-capacity sensitivity}
\label{app:probe-mlp}

We replace the linear probes with MLPs on the same frozen embeddings and
splits (Appendix~\ref{app:probe-hp}). For regression, this comparison changes
target scaling as well as probe capacity. Table~\ref{tab:probe-mlp} reports
the BGC-level classification and regression results.

\begin{table}[!htbp]
  \caption{\textbf{MLP probes for BGC-level prediction.} Metrics and splits follow
  Table~\ref{tab:t3-main}.}
  \label{tab:probe-mlp}
  \centering
  \small
  \setlength{\tabcolsep}{4pt}
  \renewcommand{\arraystretch}{1.15}
    \resizebox{\textwidth}{!}{%
    \begin{tabular}{l ccc ccc ccc}
      \toprule
      & \multicolumn{3}{c}{Coarse classes (macro-AUROC)}
      & \multicolumn{3}{c}{Fine classes (weighted macro-AUROC)}
      & \multicolumn{3}{c}{Chem. prop. (mean Spearman $\rho$)} \\
      \cmidrule(lr){2-4}\cmidrule(lr){5-7}\cmidrule(lr){8-10}
      Encoder & Random & Scaffold & GCF & Random & Scaffold & GCF & Random & Scaffold & GCF \\
      \midrule
      \multicolumn{10}{l}{\emph{Pfam-domain encoders}} \\
      Bag-of-Pfam      & 0.950 & 0.896 & 0.891 & 0.886 & 0.859 & 0.778 & 0.585 & 0.523 & 0.469 \\
      Random-Pfam      & 0.957 & 0.957 & 0.929 & 0.905 & \textbf{0.899} & 0.828 & 0.655 & 0.560 & 0.529 \\
      BiGCARP          & \textbf{0.970} & 0.950 & 0.943 & 0.912 & 0.893 & 0.847 & \textbf{0.678} & \textbf{0.604} & 0.568 \\
      BiGCARP-CLS      & 0.951 & 0.933 & 0.931 & 0.876 & 0.840 & 0.832 & 0.574 & 0.479 & 0.523 \\
      Kawano           & 0.958 & 0.956 & 0.939 & 0.910 & 0.868 & 0.846 & 0.630 & 0.552 & 0.550 \\
      DeepBGC          & 0.959 & \textbf{0.962} & 0.936 & \textbf{0.919} & 0.877 & \textbf{0.858} & 0.641 & 0.573 & \textbf{0.577} \\
      \midrule
      \multicolumn{10}{l}{\emph{Amino-acid encoders}} \\
      Amino-acid 3-mer & 0.950 & 0.931 & 0.906 & 0.899 & 0.875 & 0.846 & 0.602 & 0.519 & 0.433 \\
      BGC-MAC          & 0.956 & 0.947 & \textbf{0.948} & 0.899 & 0.837 & 0.823 & 0.604 & 0.563 & 0.536 \\
      BGCat            & 0.948 & 0.931 & 0.928 & 0.869 & 0.831 & 0.786 & 0.610 & 0.553 & 0.537 \\
      BGC-Prophet      & 0.944 & 0.952 & 0.930 & 0.873 & 0.857 & 0.805 & 0.608 & 0.512 & 0.481 \\
      LC-PLM           & 0.954 & 0.951 & 0.937 & 0.871 & 0.831 & 0.770 & 0.622 & 0.542 & 0.533 \\
      \midrule
      \multicolumn{10}{l}{\emph{Nucleotide encoders}} \\
      Nucleotide 6-mer & 0.845 & 0.823 & 0.781 & 0.847 & 0.824 & 0.776 & 0.385 & 0.361 & 0.169 \\
      Evo 2 7B         & 0.877 & 0.824 & 0.854 & 0.824 & 0.803 & 0.735 & 0.392 & 0.376 & 0.196 \\
      \bottomrule
    \end{tabular}%
    }
\end{table}

The overall pattern is consistent across probes: Pfam-domain encoders lead
most BGC-level task--split settings. The probe changes some individual leaders,
especially in chemical-property prediction.

\FloatBarrier
\paragraph{Retrieval and A-domain prediction.}
For retrieval, each linear projection is replaced by a $512$-unit hidden
layer with GELU and dropout $0.1$, followed by the $256$-dimensional output.
The loss, optimization settings, cohort, seeds, and training budget are
unchanged (Appendix~\ref{app:t3-protocol}). Retrieval scores are averaged
over three head-training seeds. A-domain prediction uses the
classification MLP of Appendix~\ref{app:probe-hp}.

\begin{table}[!htbp]
  \caption{\textbf{MLP readouts for retrieval and A-domain prediction.}}
  \label{tab:probe-mlp-t3t5}
  \centering
  \small
  \setlength{\tabcolsep}{4pt}
  \renewcommand{\arraystretch}{1.15}
    \begin{tabular}{l cccc cc}
      \toprule
      & \multicolumn{4}{c}{Retrieval (class-macro R@10)}
      & \multicolumn{2}{c}{A-domain substrate (top-1)} \\
      \cmidrule(lr){2-5}\cmidrule(lr){6-7}
      Encoder & Random & Scaffold & GCF & Class & 5-fold & GCF \\
      \midrule
      \multicolumn{7}{l}{\emph{Pfam-domain encoders}} \\
      Bag-of-Pfam      & \textbf{0.631} & \textbf{0.582} & \textbf{0.502} & \textbf{0.249} & 0.198 & 0.137 \\
      Random-Pfam      & 0.580 & 0.540 & 0.442 & 0.219 & -- & -- \\
      BiGCARP          & 0.595 & 0.529 & 0.460 & 0.232 & 0.277$^{\uparrow}$ & 0.168 \\
      BiGCARP-CLS      & 0.385 & 0.328 & 0.252 & 0.135 & -- & -- \\
      Kawano           & 0.517 & 0.467 & 0.397 & 0.194 & 0.245 & 0.161 \\
      DeepBGC          & 0.470 & 0.376 & 0.417 & 0.168 & -- & -- \\
      \midrule
      \multicolumn{7}{l}{\emph{Amino-acid encoders}} \\
      Amino-acid 3-mer & 0.303$^{\downarrow}$ & 0.242 & 0.241 & 0.127$^{\downarrow}$ & 0.764 & 0.743 \\
      BGC-MAC          & 0.505 & 0.491 & 0.422 & 0.225 & 0.745 & 0.750 \\
      BGCat            & 0.414 & 0.343 & 0.305 & 0.204 & \textbf{0.768} & 0.757 \\
      BGC-Prophet      & 0.460 & 0.436 & 0.405 & 0.199 & 0.663 & 0.610 \\
      LC-PLM           & 0.281 & 0.260 & 0.182 & 0.148 & 0.735 & \textbf{0.774} \\
      \midrule
      \multicolumn{7}{l}{\emph{Nucleotide encoders}} \\
      Nucleotide 6-mer & 0.162$^{\downarrow}$ & 0.108 & 0.131 & 0.108 & 0.596 & 0.486 \\
      Evo 2 7B         & 0.215 & 0.169 & 0.169 & 0.124 & 0.523 & 0.452 \\
      \midrule
      Kendall's $\tau$     & 0.67 & 0.64 & 0.97 & 0.82 & 0.87 & 0.63 \\
      \bottomrule
    \end{tabular}
  \par\smallskip
  {\footnotesize $^{\uparrow}$/$^{\downarrow}$: MLP $95\,\%$ CI entirely
  above/below the matched linear CI; $\tau$: ranking agreement for the twelve
  main encoders (ten for A domains).\par}
\end{table}

Bag-of-Pfam remains the leading retrieval encoder on every split. Amino-acid
encoders remain the leading group for A-domain prediction, although the best
individual encoder changes with the readout.

\FloatBarrier
\subsection{Embedding dimensionality}
\label{app:pca}

We project the nine embeddings wider than $256$ dimensions onto their first
$256$ principal components and rerun the main evaluation with the same
probes and retrieval heads. PCA is fitted on training rows only, separately
for each cross-validation fold, without whitening or $L_2$ renormalization.
Classification and regression probes still standardize the projected
features; retrieval heads use the unscaled projections. BiGCARP, DeepBGC,
and Random-Pfam are already $256$-dimensional and remain unchanged.

\par\medskip\noindent
\begin{minipage}{\linewidth}
\makeatletter
\def\@captype{table}
\makeatother
  \setlength{\abovecaptionskip}{0pt}\setlength{\belowcaptionskip}{7pt}
  \caption{\textbf{Main results after PCA to 256 dimensions.} Only the nine projected encoders
  are shown; bold marks the best among them. Arrows mark $95\,\%$ intervals
  entirely above or below the same encoder's native interval.}
  \label{tab:pca256}
  \centering
  \footnotesize
  \setlength{\tabcolsep}{1.5pt}
  \begin{tabular}{l*{15}{c}}
  \toprule
  Model & \multicolumn{3}{c}{Coarse classes} & \multicolumn{3}{c}{Fine classes} & \multicolumn{3}{c}{Chem.\ properties} & \multicolumn{4}{c}{BGC--molecule retrieval} & \multicolumn{2}{c}{A-domain substrate} \\
   & \multicolumn{3}{c}{\tiny macro-AUROC} & \multicolumn{3}{c}{\tiny weighted macro-AUROC} & \multicolumn{3}{c}{\tiny mean Spearman $\rho$} & \multicolumn{4}{c}{\tiny class-macro R@10} & \multicolumn{2}{c}{\tiny top-1 accuracy} \\
  \cmidrule(lr){2-4}\cmidrule(lr){5-7}\cmidrule(lr){8-10}\cmidrule(lr){11-14}\cmidrule(lr){15-16}
   & R & S & G & R & S & G & R & S & G & R & S & G & C & CV & G \\
  \midrule
  \multicolumn{16}{l}{\emph{Pfam-domain encoders}} \\
  Bag-of-Pfam & \textbf{.954} & \textbf{.938} & \textbf{.920} & \textbf{.886} & \textbf{.846} & .789 & \textbf{.622}$^{\uparrow}$ & \textbf{.595}$^{\uparrow}$ & \textbf{.525}$^{\uparrow}$ & \textbf{.596} & \textbf{.551} & \textbf{.532} & \textbf{.247} & .194 & .106 \\
  Kawano & .929 & .915 & .885 & .837 & .804 & .787 & .565$^{\uparrow}$ & .501$^{\uparrow}$ & .474$^{\uparrow}$ & .512 & .457 & .413 & .177 & .200 & .106 \\
  \midrule
  \multicolumn{16}{l}{\emph{Amino-acid encoders}} \\
  Amino-acid 3-mer & .912 & .881 & .858 & .854 & .836 & \textbf{.794} & .464 & .469 & .338 & .103$^{\downarrow}$ & .135$^{\downarrow}$ & .113$^{\downarrow}$ & .099$^{\downarrow}$ & .727$^{\downarrow}$ & .692 \\
  BGC-MAC & .942 & .908 & .908 & .845 & .791 & .739 & .533 & .539$^{\uparrow}$ & .511$^{\uparrow}$ & .472 & .438 & .360 & .216 & \textbf{.760} & .729 \\
  BGCat & .930 & .907 & .882 & .820 & .748 & .736 & .505 & .523$^{\uparrow}$ & .509$^{\uparrow}$ & .370 & .266$^{\downarrow}$ & .250 & .172 & .732 & \textbf{.760} \\
  BGC-Prophet & .920 & .879 & .876 & .793 & .772$^{\downarrow}$ & .721 & .503 & .436 & .423 & .427 & .418 & .401 & .181 & .643 & .596 \\
  LC-PLM & .920 & .899 & .883 & .798 & .750$^{\downarrow}$ & .674 & .527$^{\uparrow}$ & .473$^{\uparrow}$ & .441$^{\uparrow}$ & .253 & .214 & .156 & .123 & .690$^{\downarrow}$ & .664 \\
  \midrule
  \multicolumn{16}{l}{\emph{Nucleotide encoders}} \\
  Nucleotide 6-mer & .822 & .783 & .767 & .798 & .720 & .727 & .380$^{\uparrow}$ & .356$^{\uparrow}$ & .236$^{\uparrow}$ & .052$^{\downarrow}$ & .076 & .060 & .105 & .494$^{\downarrow}$ & .360 \\
  Evo 2 7B & .900 & .858 & .836 & .756 & .709 & .732 & .418 & .395 & .326 & .203 & .142 & .149 & .120 & .605 & .534 \\
  \bottomrule
  \end{tabular}
\end{minipage}
\par\medskip

PCA leaves biosynthetic class prediction largely unchanged and improves
chemical-property prediction for most encoders. Bag-of-Pfam leads all three
chemical-property splits after projection. The retrieval losses of the
$k$-mer baselines depend on feature scale: a separate diagnostic that
$L_2$-normalizes the projected vectors yields intervals overlapping their
native counterparts. The results therefore reflect both dimensionality
reduction and its interaction with the readout.

\FloatBarrier
\subsection{Continued pretraining}
\label{app:adaptation}

We continue masked-token pretraining of BiGCARP and Kawano on the Pfam
sequences of $2{,}626$ MIBiG BGCs, including unlabeled sequences from the
evaluation holdouts. Training uses three seeds, a learning rate of $10^{-4}$,
and up to $50$ epochs, retaining the checkpoint with the lowest loss on a
$10\,\%$ validation subset. The adapted encoders are then evaluated frozen
with the same downstream protocols as the released checkpoints.

\begin{table}[!htbp]
  \caption{\textbf{Effect of continued pretraining.} Mean score change across three seeds (adapted
  minus released).}
  \label{tab:ssl-value}
  \centering
  \small
    \resizebox{\textwidth}{!}{%
    \begin{tabular}{ll rrr rrr}
      \toprule
      & & \multicolumn{3}{c}{BiGCARP} & \multicolumn{3}{c}{Kawano} \\
      \cmidrule(lr){3-5}\cmidrule(lr){6-8}
      Task & Probe & Rand. & Scaf. & GCF & Rand. & Scaf. & GCF \\
      \midrule
      Coarse classes & linear & $-0.001$ & $-0.007$ & $+0.010$ & $+0.014$ & $+0.008$ & $0.000$ \\
       & MLP & $-0.003$ & $+0.006$ & $+0.001$ & $+0.003$ & $+0.007$ & $+0.001$ \\
      \addlinespace
      Fine classes & linear & $0.000$ & $-0.002$ & $-0.014$ & $+0.029$ & $+0.027$ & $+0.014$ \\
       & MLP & $-0.001$ & $-0.003$ & $-0.011$ & $+0.001$ & $+0.036$ & $+0.004$ \\
      \addlinespace
      Chem.\ properties & linear & $-0.008$ & $+0.013$ & $-0.032$ & $-0.004$ & $+0.022$ & $+0.010$ \\
      \addlinespace
      Retrieval &  & $-0.008$ & $-0.027$ & $+0.019$ & $+0.060$ & $+0.169$ & $+0.058$ \\
      \bottomrule
    \end{tabular}%
    }
\end{table}

Kawano gains most on retrieval, particularly on scaffold holdout, while
BiGCARP shows no consistent improvement across tasks.

\FloatBarrier
\begin{samepage}
\subsection{Label-source sensitivity}
\label{app:label-provenance}

ClassyFire~\citep{classyfire} assigns molecules to the hierarchical ChemOnt
taxonomy using structural rules, providing a general chemical classification
alongside NPClassifier's natural-product taxonomy. Using linear probes, we compare NPClassifier
labels (A) with ClassyFire labels on the same BGCs and splits (B), or with a
ClassyFire-defined cohort and splits (C). Classes require at least ten BGCs;
B keeps BGCs without a qualifying ClassyFire class as all-negative examples.
\par
\end{samepage}

\begin{sidewaystable}
  \setlength{\abovecaptionskip}{0pt}\setlength{\belowcaptionskip}{7pt}
  \caption{\textbf{Sensitivity to chemical label source.} Weighted macro-AUROC with $95\,\%$
  intervals. $\tau$ compares rankings with NPClassifier; its interval resamples encoders.}
  \label{tab:label-provenance}
  \centering
  \small
  \begin{tabular}{l ccc ccc ccc}
  \toprule
   & \multicolumn{3}{c}{NPClassifier (arm A)} & \multicolumn{3}{c}{ClassyFire, assignment-matched (arm B)} & \multicolumn{3}{c}{ClassyFire, native (arm C)} \\
  \cmidrule(lr){2-4}\cmidrule(lr){5-7}\cmidrule(lr){8-10}
  Encoder & Random & Scaffold & GCF & Random & Scaffold & GCF & Random & Scaffold & GCF \\
  \midrule
  \multicolumn{10}{l}{\emph{Pfam-domain encoders}} \\
  Bag-of-Pfam & \ci{0.880}{0.023}{0.031} & \ci{0.825}{0.028}{0.028} & \ci{0.753}{0.041}{0.039} & \ci{0.826}{0.032}{0.036} & \ci{0.740}{0.034}{0.034} & \ci{0.726}{0.041}{0.039} & \ci{0.810}{0.029}{0.028} & \ci{0.795}{0.028}{0.027} & \ci{0.730}{0.033}{0.035} \\
  Random-Pfam & \ci{0.880}{0.023}{0.027} & \ci{0.851}{0.025}{0.026} & \ci{0.786}{0.033}{0.033} & \ci{0.819}{0.035}{0.036} & \ci{0.769}{0.030}{0.037} & \ci{0.727}{0.040}{0.047} & \ci{0.826}{0.027}{0.025} & \ci{0.811}{0.024}{0.025} & \ci{0.744}{0.029}{0.029} \\
  BiGCARP & \ci{0.880}{0.025}{0.027} & \ciB{0.872}{0.022}{0.023} & \ciB{0.834}{0.031}{0.028} & \ci{0.824}{0.033}{0.034} & \ci{0.805}{0.031}{0.032} & \ci{0.761}{0.040}{0.042} & \ci{0.831}{0.027}{0.028} & \ci{0.826}{0.025}{0.023} & \ci{0.780}{0.030}{0.028} \\
  BiGCARP-CLS & \ci{0.831}{0.028}{0.028} & \ci{0.806}{0.028}{0.030} & \ci{0.783}{0.032}{0.034} & \ci{0.744}{0.032}{0.034} & \ci{0.735}{0.034}{0.033} & \ci{0.704}{0.038}{0.036} & \ci{0.752}{0.027}{0.030} & \ci{0.773}{0.028}{0.025} & \ci{0.737}{0.029}{0.032} \\
  Kawano & \ci{0.873}{0.025}{0.029} & \ci{0.844}{0.024}{0.023} & \ci{0.810}{0.030}{0.031} & \ci{0.823}{0.036}{0.033} & \ci{0.803}{0.029}{0.032} & \ci{0.780}{0.040}{0.040} & \ci{0.799}{0.029}{0.029} & \ci{0.786}{0.025}{0.027} & \ci{0.752}{0.031}{0.029} \\
  DeepBGC & \ciB{0.893}{0.021}{0.026} & \ci{0.863}{0.021}{0.022} & \ci{0.830}{0.027}{0.029} & \ciB{0.842}{0.027}{0.032} & \ciB{0.823}{0.028}{0.029} & \ciB{0.801}{0.042}{0.040} & \ciB{0.844}{0.024}{0.027} & \ciB{0.850}{0.027}{0.023} & \ciB{0.801}{0.027}{0.027} \\
  \midrule
  \multicolumn{10}{l}{\emph{Amino-acid encoders}} \\
  Amino-acid 3-mer & \ci{0.887}{0.023}{0.027} & \ci{0.850}{0.027}{0.029} & \ci{0.822}{0.036}{0.037} & \ci{0.817}{0.036}{0.041} & \ci{0.768}{0.027}{0.031} & \ci{0.716}{0.047}{0.050} & \ci{0.826}{0.030}{0.027} & \ci{0.805}{0.029}{0.030} & \ci{0.716}{0.039}{0.040} \\
  BGC-MAC & \ci{0.853}{0.032}{0.039} & \ci{0.812}{0.030}{0.029} & \ci{0.763}{0.033}{0.039} & \ci{0.814}{0.034}{0.032} & \ci{0.757}{0.032}{0.035} & \ci{0.697}{0.039}{0.045} & \ci{0.807}{0.029}{0.029} & \ci{0.795}{0.029}{0.025} & \ci{0.734}{0.030}{0.033} \\
  BGCat & \ci{0.847}{0.035}{0.039} & \ci{0.789}{0.034}{0.033} & \ci{0.750}{0.034}{0.039} & \ci{0.802}{0.035}{0.037} & \ci{0.738}{0.035}{0.037} & \ci{0.705}{0.048}{0.046} & \ci{0.812}{0.026}{0.026} & \ci{0.787}{0.031}{0.030} & \ci{0.711}{0.034}{0.031} \\
  BGC-Prophet & \ci{0.854}{0.028}{0.031} & \ci{0.833}{0.024}{0.027} & \ci{0.763}{0.033}{0.037} & \ci{0.818}{0.034}{0.035} & \ci{0.782}{0.029}{0.029} & \ci{0.740}{0.037}{0.041} & \ci{0.784}{0.028}{0.030} & \ci{0.783}{0.027}{0.028} & \ci{0.727}{0.035}{0.033} \\
  LC-PLM & \ci{0.843}{0.032}{0.036} & \ci{0.810}{0.026}{0.026} & \ci{0.701}{0.038}{0.048} & \ci{0.840}{0.029}{0.033} & \ci{0.775}{0.030}{0.036} & \ci{0.683}{0.044}{0.041} & \ci{0.806}{0.026}{0.026} & \ci{0.784}{0.030}{0.025} & \ci{0.731}{0.032}{0.031} \\
  \midrule
  \multicolumn{10}{l}{\emph{Nucleotide encoders}} \\
  Nucleotide 6-mer & \ci{0.834}{0.027}{0.029} & \ci{0.778}{0.030}{0.029} & \ci{0.740}{0.032}{0.036} & \ci{0.750}{0.038}{0.048} & \ci{0.699}{0.037}{0.040} & \ci{0.661}{0.040}{0.043} & \ci{0.700}{0.033}{0.029} & \ci{0.710}{0.031}{0.036} & \ci{0.636}{0.037}{0.034} \\
  Evo 2 7B & \ci{0.801}{0.030}{0.034} & \ci{0.750}{0.033}{0.033} & \ci{0.751}{0.035}{0.037} & \ci{0.735}{0.035}{0.042} & \ci{0.716}{0.037}{0.039} & \ci{0.671}{0.044}{0.042} & \ci{0.731}{0.033}{0.033} & \ci{0.709}{0.031}{0.029} & \ci{0.670}{0.034}{0.035} \\
  \midrule
  Classes / test BGCs & 48 / 241 & 44 / 267 & 40 / 212 & 24 / 241 & 23 / 267 & 20 / 212 & 33 / 343 & 31 / 306 & 25 / 279 \\
  $\tau$ vs.\ arm A & -- & -- & -- & 0.615 & 0.692 & 0.615 & 0.641 & 0.744 & 0.590 \\
  \quad 95\,\% interval & -- & -- & -- & $[0.08, 1.00]$ & $[0.41, 0.89]$ & $[0.36, 0.83]$ & $[0.32, 0.88]$ & $[0.40, 0.97]$ & $[0.16, 0.92]$ \\
  \bottomrule
  \end{tabular}
\end{sidewaystable}

The label source changes individual encoder rankings, while Pfam-domain
encoders retain the leading scores. DeepBGC leads every ClassyFire comparison.

\FloatBarrier

\section{Active-learning analysis}
\label{app:al}

In this section, we describe the active-learning protocol and present the
results supporting Figure~\ref{fig:al-main}. We compare acquisition
strategies and quantify encoder label efficiency.

\FloatBarrier
\subsection{Acquisition protocol and tests}
\label{app:al-protocol}
\paragraph{Data and acquisition rounds.}
We use each task's random split, taking fold~4 as the test set for fine-class
and A-domain prediction. We reserve $20\,\%$ of the training split for
validation and use the remaining examples as the acquisition pool. A-domain
partitions keep all domains from a BGC together. Each run starts with $16$
labeled examples. The label budget grows by a factor of about $\sqrt{2}$ per
round, following $n_k = \mathrm{round}(16 \cdot 2^{k/2})$, until the whole
pool is labeled. Full-pool scores exclude the validation subset;
Table~\ref{tab:t3-main} uses the full training split and all five A-domain
folds.

At each round, we refit the task's readout from scratch on the labeled
examples while keeping the encoder frozen (Appendix~\ref{app:probe-hp}).
Retrieval refits its relation-masked contrastive head with batches of
$\min(256,n)$; the epoch count is chosen to match the main-table budget of
$350$ optimizer steps. We then acquire the next batch of labels using one
of the following strategies.

\paragraph{Acquisition strategies.}
Random acquisition follows a fixed random ordering of the pool.
TypiClust~\citep{hacohen2022typiclust} standardizes the embeddings and
reduces them to $50$ principal components. It prioritizes clusters with few
labeled examples and selects typical unlabeled members, with the number
of clusters capped at one tenth of the pool size.
BADGE~\citep{ash2020badge} applies $k$-means++ seeding to gradient embeddings
formed from the readout's predicted labels. For retrieval, its gradient
embeddings use products of currently labeled BGCs as candidates.
Classification and retrieval compare all three strategies; chemical-property
regression compares TypiClust with random acquisition.

\paragraph{Learning-curve summary and strategy selection.}
We summarize performance between $16$ and $256$ labels using
$\mathrm{AULC}_{\mathrm{low}}$. This is the trapezoidal area under the score
curve on a $\log_2 n$ axis, divided by the axis range. It gives each doubling
of the label budget equal weight and remains on the scale of the task metric.

We run $20$ seeds, each defining starting labels shared by all encoders and
strategies and the ordering used by random acquisition. For each encoder, we use seeds~1--10 to select the strategy with the highest
validation $\mathrm{AULC}_{\mathrm{low}}$. Figure~\ref{fig:al-main} reports
test results on seeds~11--20, with bands from $2{,}000$ bootstrap resamples
of those seeds. The appendix tables use all $20$ seeds.

\paragraph{Statistical comparisons.}
We compare each strategy with random acquisition through seed-paired
$\mathrm{AULC}_{\mathrm{low}}$ differences. Per-encoder comparisons use
two-sided seed-bootstrap tests with $10{,}000$ resamples and
Benjamini--Hochberg correction across $104$ comparisons. For task-level
comparisons, we first average the differences over encoders within each seed,
then apply an exact two-sided sign-flip test over the $20$ seeds and Holm
correction across nine strategy--task comparisons. We use a significance
level of $0.05$ after correction. The data partitions stay fixed, so these
tests capture variation across acquisition seeds.

The framework can also be extended to refit pooling heads or fine-tune the
encoder on the labeled examples at each round.

\FloatBarrier

\subsection{Acquisition gains over random acquisition}
\label{app:al-gain}

Table~\ref{tab:al-gain-encoder} compares acquisition strategies for each
encoder. TypiClust significantly improves the task-level summary on four of the
five tasks, while BADGE significantly improves only coarse class prediction.

\begin{table}[!htbp]
  \caption{\textbf{Acquisition gains by encoder.} Mean $\mathrm{AULC}_{\mathrm{low}}$ gains over
  random acquisition, in metric points ($\times100$). Bold marks significant differences
  under the corrected tests.}
  \label{tab:al-gain-encoder}
  \centering
  \footnotesize
  \setlength{\tabcolsep}{4pt}
  \begin{tabular}{l*{9}{r}}
  \toprule
   & \multicolumn{5}{c}{TypiClust $-$ random} & \multicolumn{4}{c}{BADGE $-$ random} \\
  \cmidrule(lr){2-6}\cmidrule(lr){7-10}
  Encoder & Coarse & Fine & Chem. & Retr. & A-dom. & Coarse & Fine & Retr. & A-dom. \\
  \midrule
  Bag-of-Pfam & +0.5 & \textbf{+0.8} & $-$1.6 & +0.3 & \textbf{+0.8} & \textbf{+2.6} & +0.1 & \textbf{+2.3} & 0.0 \\
  Random-Pfam & \textbf{+3.0} & +0.7 & \textbf{+4.0} & \textbf{+5.5} & -- & +0.9 & $-$0.2 & +1.4 & -- \\
  BiGCARP & \textbf{+1.1} & \textbf{+1.5} & \textbf{+6.8} & \textbf{+5.3} & \textbf{+1.5} & \textbf{+1.6} & $-$0.4 & +1.3 & 0.0 \\
  Kawano & \textbf{+3.0} & \textbf{+1.5} & \textbf{+3.7} & \textbf{+5.9} & \textbf{+1.2} & \textbf{+1.3} & $-$0.5 & +0.8 & +0.1 \\
  DeepBGC & \textbf{+1.7} & \textbf{+1.5} & \textbf{+1.5} & \textbf{+3.2} & -- & +1.1 & \textbf{+0.8} & +0.5 & -- \\
  \addlinespace[2pt]
  Amino-acid 3-mer & +0.7 & $-$0.6 & $-$0.9 & \textbf{+5.7} & \textbf{+3.3} & \textbf{$-$1.2} & \textbf{$-$2.9} & +0.4 & +1.5 \\
  BGC-MAC & \textbf{+2.1} & +0.1 & \textbf{+2.4} & \textbf{+5.0} & \textbf{+5.1} & \textbf{+1.9} & $-$0.6 & 0.0 & +0.8 \\
  BGCat & \textbf{+1.7} & +0.6 & +0.1 & \textbf{+3.0} & \textbf{+5.0} & \textbf{+1.6} & $-$0.4 & $-$0.6 & \textbf{+1.7} \\
  BGC-Prophet & \textbf{+1.6} & $-$0.1 & $-$0.4 & \textbf{+3.2} & \textbf{+3.7} & \textbf{+1.8} & \textbf{$-$1.1} & +0.2 & $-$0.2 \\
  LC-PLM & \textbf{+1.9} & $-$0.3 & \textbf{+3.6} & \textbf{+3.1} & \textbf{+4.4} & +0.9 & +0.2 & $-$0.3 & $-$0.1 \\
  \addlinespace[2pt]
  Nucleotide 6-mer & 0.0 & +0.4 & +0.3 & \textbf{+4.7} & +0.8 & +0.8 & $-$0.5 & $-$0.9 & \textbf{$-$1.2} \\
  Evo 2 7B & \textbf{+1.7} & $-$0.2 & +1.5 & \textbf{+3.6} & \textbf{+1.1} & +0.8 & +0.1 & $-$0.1 & \textbf{$-$0.8} \\
  \midrule
  Encoder mean & \textbf{+1.6} & +0.5 & \textbf{+1.8} & \textbf{+4.0} & \textbf{+2.7} & \textbf{+1.2} & $-$0.5 & +0.4 & +0.2 \\
  Holm $p$ & $6{\times}10^{-4}$ & 0.079 & $4{\times}10^{-4}$ & $3{\times}10^{-5}$ & $2{\times}10^{-5}$ & 0.005 & 0.103 & 0.719 & 0.719 \\
  \bottomrule
  \end{tabular}
\end{table}

\subsection{Label efficiency}
\label{app:al-savings}
We measure label efficiency by how many labeled examples an encoder needs
under random acquisition to reach the same test score as a reference model. In
Table~\ref{tab:al-passive}, the reference is Bag-of-Pfam under random
acquisition. Its $20$-seed mean scores at $32$, $64$, and $128$ labels define
three target scores.

For each target score, we find where the encoder's $20$-seed mean curve
first reaches it, interpolating linearly against $\log_2 n$. We divide
Bag-of-Pfam's label requirement by the encoder's requirement. At a target
score, a ratio of two means that the encoder needs half as many labels. We
report the geometric mean of the three ratios.

If an encoder reaches a target by the initial $16$ labels, its saving is a
lower bound because performance at smaller budgets is unobserved. If the
target remains unreached when the pool is exhausted, its saving is an upper
bound. This table always uses Bag-of-Pfam as the reference;
Figure~\ref{fig:al-main} instead selects the frequency-based baseline by
validation performance.

\begin{table}[!htbp]
  \caption{\textbf{Label savings of each encoder under random acquisition, relative to Bag-of-Pfam under random acquisition.}}
  \label{tab:al-passive}
  \centering
  \small
  \setlength{\tabcolsep}{6pt}
  \begin{tabular}{l*{5}{r}}
  \toprule
  Encoder & Coarse & Fine & Chem. & Retr. & A-dom. \\
  \midrule
  Bag-of-Pfam & 1.00 & 1.00 & 1.00 & 1.00 & 1.00 \\
  Random-Pfam & 1.57 & 1.13 & 0.28 & 0.83 & -- \\
  BiGCARP & 2.22 & 1.12 & $\geq$1.00 & 1.18 & 0.91 \\
  Kawano & 1.72 & 1.01 & 0.60 & 0.78 & 1.19 \\
  DeepBGC & 1.49 & 0.86 & 1.26 & 0.58 & -- \\
  \addlinespace[2pt]
  Amino-acid 3-mer & 0.27 & 0.81 & 0.63 & 0.53 & $\geq$4.00 \\
  BGC-MAC & 0.52 & 0.42 & 0.28 & 0.53 & $\geq$4.00 \\
  BGCat & 0.28 & 0.31 & 0.20 & 0.50 & $\geq$4.00 \\
  BGC-Prophet & 0.48 & 0.45 & 0.16 & 0.48 & $\geq$3.79 \\
  LC-PLM & 0.41 & 0.31 & 0.43 & 0.23 & $\geq$3.90 \\
  \addlinespace[2pt]
  Nucleotide 6-mer & $\leq$0.04 & 0.26 & $\leq$0.05 & 0.27 & $\geq$3.42 \\
  Evo 2 7B & 0.08 & $\leq$0.15 & 0.13 & $\leq$0.17 & 2.75 \\
  \bottomrule
  \end{tabular}
\end{table}

\clearpage

\section*{NeurIPS Paper Checklist}

\begin{enumerate}

\item {\bf Claims}
    \item[] Question: Do the main claims made in the abstract and introduction accurately reflect the paper's contributions and scope?
    \item[] Answer: \answerYes{}.
    \item[] Justification: The abstract and introduction (Section~\ref{sec:intro}) state the benchmark scope, the task and split design, and the empirical findings. The results and limitations sections (Sections~\ref{sec:results} and~\ref{sec:discussion}) qualify these claims by task, split, and probe.

\item {\bf Limitations}
    \item[] Question: Does the paper discuss the limitations of the work performed by the authors?
    \item[] Answer: \answerYes{}.
    \item[] Justification: Limitations are discussed in the dedicated Limitations section, including corpus size and bias, tool-derived labels and splits, encoder coverage, and task scope.

\item {\bf Theory assumptions and proofs}
    \item[] Question: For each theoretical result, does the paper provide the full set of assumptions and a complete (and correct) proof?
    \item[] Answer: \answerNA{}.
    \item[] Justification: The paper is an empirical benchmark and dataset contribution; it does not introduce theoretical results, theorems, or proofs.

    \item {\bf Experimental result reproducibility}
    \item[] Question: Does the paper fully disclose all the information needed to reproduce the main experimental results of the paper to the extent that it affects the main claims and/or conclusions of the paper (regardless of whether the code and data are provided or not)?
    \item[] Answer: \answerYes{}.
    \item[] Justification: Section~\ref{sec:bm} and the appendices specify dataset construction and splits (Appendix~\ref{app:data}), the tasks (Appendix~\ref{app:tasks}), the encoders, probes and variance estimation (Appendix~\ref{app:eval}), the sensitivity analyses and ablations (Appendix~\ref{app:probe}), and the active-learning protocol (Appendix~\ref{app:al}); the released code and dataset contain the curated MIBiG-derived inputs, labels, split files, molecular features, main-benchmark and active-learning evaluation code, and a pinned \texttt{uv} environment. The release scripts reproduce Table~\ref{tab:t3-main} and the active-learning results.

\item {\bf Open access to data and code}
    \item[] Question: Does the paper provide open access to the data and code, with sufficient instructions to faithfully reproduce the main experimental results, as described in supplemental material?
    \item[] Answer: \answerYes{}.
    \item[] Justification: The code (\url{https://github.com/Nugkta/BGC-Bench}) and the dataset (\url{https://huggingface.co/datasets/Nugkta/bgc-bench}), linked from the abstract and described in Appendices~\ref{app:data} and~\ref{app:tasks}, are publicly available without authentication. They include reproduction commands, a pinned environment, raw model inputs, labels, split files, a dataset card, Croissant metadata, and a file manifest.

\item {\bf Experimental setting/details}
    \item[] Question: Does the paper specify all the training and test details (e.g., data splits, hyperparameters, how they were chosen, type of optimizer) necessary to understand the results?
    \item[] Answer: \answerYes{}.
    \item[] Justification: The benchmark construction, task splits, frozen-encoder protocol, probe architectures, retrieval head, active-learning strategies, hyperparameters, and seed protocol are specified in Section~\ref{sec:bm} and Appendices~\ref{app:probe-hp}, \ref{app:t3-protocol}, and \ref{app:al-protocol}.

\item {\bf Experiment statistical significance}
    \item[] Question: Does the paper report error bars suitably and correctly defined or other appropriate information about the statistical significance of the experiments?
    \item[] Answer: \answerYes{}.
    \item[] Justification: The main table reports each bootstrap 95\% confidence interval as $\pm$ the larger distance to either bound (exact intervals in Table~\ref{tab:main-ci}) and, per column, the fraction of encoder pairs whose intervals do not overlap; Appendix~\ref{app:variance} defines the resampling protocol of each task, and Appendix~\ref{app:al} gives the seed-level tests used for active learning.

\item {\bf Experiments compute resources}
    \item[] Question: For each experiment, does the paper provide sufficient information on the computer resources (type of compute workers, memory, time of execution) needed to reproduce the experiments?
    \item[] Answer: \answerYes{}.
    \item[] Justification: Appendix~\ref{app:compute} gives the hardware and the runtimes of the main benchmark and active-learning experiments, which used CPU workers and NVIDIA A100, L40S, H200, and GH200 GPUs across the University of Manchester Computational Shared Facility (CSF) and Isambard-AI. The appendix sensitivity analyses are not timed separately, and a total for the full research project is not provided.

\item {\bf Code of ethics}
    \item[] Question: Does the research conducted in the paper conform, in every respect, with the NeurIPS Code of Ethics \url{https://neurips.cc/public/EthicsGuidelines}?
    \item[] Answer: \answerYes{}.
    \item[] Justification: The benchmark uses public MIBiG-derived sequences and metadata, derived embeddings, and computed labels; it involves no human subjects, no scraped personal data, and no high-risk generative model.

\item {\bf Broader impacts}
    \item[] Question: Does the paper discuss both potential positive societal impacts and negative societal impacts of the work performed?
    \item[] Answer: \answerYes{}.
    \item[] Justification: As motivated in the introduction (Section~\ref{sec:intro}), standardized, reproducible evaluation of BGC models supports natural-product discovery, including commercial products such as drugs and crop-protection agents. The main dual-use risk is that better gene-cluster-to-chemistry mapping could in principle help predict bioactive small molecules from genomic data. The benchmark redistributes public MIBiG-derived data and computed annotations, and introduces no generative model or laboratory synthesis protocol.

\item {\bf Safeguards}
    \item[] Question: Does the paper describe safeguards that have been put in place for responsible release of data or models that have a high risk for misuse (e.g., pre-trained language models, image generators, or scraped datasets)?
    \item[] Answer: \answerNA{}.
    \item[] Justification: The release is an evaluation benchmark over public MIBiG-derived sequences, metadata, and computed annotations, not a high-risk generative model or scraped personal dataset.

\item {\bf Licenses for existing assets}
    \item[] Question: Are the creators or original owners of assets (e.g., code, data, models), used in the paper, properly credited and are the license and terms of use explicitly mentioned and properly respected?
    \item[] Answer: \answerYes{}.
    \item[] Justification: Section~\ref{sec:bm}, Table~\ref{tab:model-inventory}, and Appendices~\ref{app:curation} and~\ref{app:encoders} cite the data sources, annotation tools, and encoders. The dataset package specifies CC BY 4.0 and the benchmark code specifies MIT, with attribution to MIBiG, NPClassifier, RDKit, and the encoder authors. Third-party data keep their source terms: MIBiG~4.0 is CC BY 4.0, Pfam is CC0, RDKit is BSD-3-Clause, and the NPClassifier labels come from its public web service, which declares no license; ClassyFire, used only in an appendix analysis, is cited, and its labels are not redistributed. Checkpoint downloads are pinned; some use project-maintained mirrors. Upstream terms remain applicable: BiGCARP, DeepBGC, ESM-2, and Uni-Mol use MIT; Evo~2 and MoLFormer use Apache 2.0; BiG-SCAPE uses AGPL-3.0; Kawano and LC-PLM use CC BY-NC 4.0; ESM~C uses its noncommercial model license; ChemBERTa-2 declares no license, so CC BY 4.0 covers its redistributed vectors only as far as the authors hold rights in them. The benchmark code license does not replace these checkpoint terms.

\item {\bf New assets}
    \item[] Question: Are new assets introduced in the paper well documented and is the documentation provided alongside the assets?
    \item[] Answer: \answerYes{}.
    \item[] Justification: Appendices~\ref{app:data} and~\ref{app:tasks} document the dataset construction, splits, and tasks. The released code and dataset contain \texttt{README} files, a dataset card describing contents and limitations, Croissant metadata, a file manifest, license information, reproduction commands, and a pinned \texttt{uv} environment. The dataset card is not a separate Datasheet for Datasets, and the runtime examples are not a complete compute accounting.

\item {\bf Crowdsourcing and research with human subjects}
    \item[] Question: For crowdsourcing experiments and research with human subjects, does the paper include the full text of instructions given to participants and screenshots, if applicable, as well as details about compensation (if any)?
    \item[] Answer: \answerNA{}.
    \item[] Justification: The benchmark uses public microbial gene-cluster and natural-product metadata and does not involve crowdsourcing, user studies, or human subjects research.

\item {\bf Institutional review board (IRB) approvals or equivalent for research with human subjects}
    \item[] Question: Does the paper describe potential risks incurred by study participants, whether such risks were disclosed to the subjects, and whether Institutional Review Board (IRB) approvals (or an equivalent approval/review based on the requirements of your country or institution) were obtained?
    \item[] Answer: \answerNA{}.
    \item[] Justification: No human subjects, participant data, or crowdsourced labor are used, so IRB or equivalent review is not applicable.

\item {\bf Declaration of LLM usage}
    \item[] Question: Does the paper describe the usage of LLMs if it is an important, original, or non-standard component of the core methods in this research? Note that if the LLM is used only for writing, editing, or formatting purposes and does \emph{not} impact the core methodology, scientific rigor, or originality of the research, declaration is not required.
    \item[] Answer: \answerNA{}.
    \item[] Justification: LLMs are not part of the benchmark construction, model evaluation, scientific analysis, or reported methodology; writing or formatting assistance does not require declaration under this question.

\end{enumerate}

\end{document}